\documentclass[12pt]{article}

\def\modifymargins#1#2{
\newdimen\addtoh
\newdimen\addtow
\addtoh=#1
\addtow=#2

\advance\topmargin by -\addtoh
\multiply\addtoh by 2
\advance\textheight by \addtoh

\advance\oddsidemargin by -\addtow
\advance\evensidemargin by -\addtow
\multiply\addtow by 2
\advance\textwidth by \addtow
}

\expandafter\ifx\csname ttytty\endcsname\relax
\else\modifymargins{60pt}{0pt}\def\_{\tt\char 95}
\raggedright\parindent 40pt

\fi

\def\draft{\iftrue\begingroup\par\em=============[draft]=============\par}
\def\enddraft{\par=============[/draft]=============\par\endgroup\fi}
\def\draft{\iffalse}
\let\enddraft=\fi

\def\nojournal{\iftrue\begingroup\par\em=============[nojournal]=============\par}
\def\endnojournal{\par=============[/nojournal]=============\par\endgroup\fi}
\def\nojournal{\iftrue}
\let\endnojournal=\fi

\def\possnewtheorem#1#2{
\expandafter\ifx\csname #1\endcsname\relax
\newtheorem{#1}{#2}
\fi
}

\possnewtheorem{theorem}{Theorem}
\possnewtheorem{corollary}{Corollary}
\possnewtheorem{lemma}{Lemma}
\possnewtheorem{definition}{Definition}
\possnewtheorem{algorithm}{Algorithm}
\newtheorem{condition}{Condition}

\long\def\nop#1{}
\def\np{{\rm NP}}
\def\true{{\sf true}}
\def\false{{\sf false}}
\def\proof{\noindent {\sl Proof.\ \ }}
\def\qed{\hfill{\boxit{}}
  \ifdim\lastskip<\medskipamount \removelastskip\penalty55\medskip\fi}
\long\def\boxit#1{\vbox{\hrule\hbox{\vrule\kern3pt
  \vbox{\kern3pt#1\kern3pt}\kern3pt\vrule}\hrule}}
\let\plural=\relax
\def\color[#1]#2{}

\def\equiset{\mbox{\sc equiset}}
\def\equiall{\mbox{\sc equiall}}

\title{One head is better than two:
a polynomial restriction for propositional definite Horn forgetting}
\author{Paolo Liberatore\\
DIAG - Sapienza University of Rome,\\
Rome, Italy.\\
Email: {\tt liberato@diag.uniroma1.it}\\
ORCID: 0000-0001-5355-3766}

\date{}

\begin{document}

\maketitle

\draft
\section{Synopsis}

\begin{enumerate}

\item introduction (introduction.tex)

\item preliminaries (preliminaries.tex)

\item the insidious single-head equivalence problem (insidious.tex, other.tex);
\begin{itemize}

\item dato che con single-head si puo' fare forget e common equivalence in
tempo polinomiale, ha senso chiedersi se una formula ne ha una equivalente
single-head

\item una condizione necessaria sembra a prima vista sufficiente ma non lo e'

\item una serie di esempi mostra che la cosa non e' semplice come sembra

\item si dimostra anche una seconda condizione necessaria, e sempre non e'
sufficiente

\item ma tutti questi esempi sono ciclici, quindi nella prossima sezione viene
studiato il caso di formule acicliche

\item alcune formule sono single-head equivalenti ma nessun sottoinsieme e'
single-head equivalente

\item il lemma delle precondizioni equivalenti: le clausole $P' \rightarrow x$
hanno $F \models P \equiv P'$ dove $P \rightarrow x$ e' la singola clausola con
$x$

\end{itemize}

\item cyclicity (cyclic.tex)

dato che i problemi vengono fuori sulle formule cicliche,
ha senso studiare la ciclicita'

\begin{itemize}

\item esiste una definizione sintattica e una semantica di ciclicita'

\item si dimostra che corrispondono: una formula e' semanticamente aciclica se
e solo se ha una forma equivalente che e' sintatticamente aciclica

\item sulle formula semanticamente acicliche la prima condizione necessaria e'
anche sufficiente

\item la ciciclita' sembra a prima vista corrispondere alla equivalenza fra
insiemi di variabili che non sono contenuti l'uno nell'altro; si definisce
quindi l'assenza di questi come inequivalenza; pero' non corrisponde alla
aciclicita': inequivalenza implica ciclicita' ma non il contrario; pero' la
condizione di inequivalenza si vedra' poi essere comunque utile

\end{itemize}

\item the order and the algorithm (order.tex)

se una formula e' single-head, allora ogni insieme di variabili $B$ che implica
$x$ deve ``passare'' per l'unica clausola $A \rightarrow x \in F$ che ha $x$
come testa, ossia $F \models B \rightarrow A$; dato che quest'ultima e' una
condizione semantica vale per tutte le formule equivalenti; questo suggerisce
di definire un ordine fra insiemi di variabili e poi prendere il minimo

\begin{itemize}

\item il lemma della convergenza: se $A$ e $B$ implicano $x$, allora esiste un
sottoinsieme di $A \cup B$ che e' implicato sia da $A$ che da $B$ e che implica
$x$

\item la formula $MIN(F)$ dei corpi minimi: contiene tutte le clausole $A
\rightarrow x$ per le quali non esiste $B$ che sia minore di $A$ (per $<_F$ o
per $\subset$) e che implica $x$

\item se questa formula e' single-head ed equivalente a $F$ allora $F$ e'
single-head equivalente; questo e' ovvio, ma poi vale anche il contrario se $F$
e' inequivalente (definito nella sezione precedente)

\item calcolare $MIN(F)$ non e' facile, e potrebbe essere esponenzialmente
grande nel caso non-singlehead; quindi invece si calcola $SHMIN(F)$, ottenuta
prendendo una sola clausola di $MIN(F)$ per ogni testa; e' uguale a $MIN(F)$ se
questa e' single-head, ma si calcola in tempo polinomiale

\item una $SHMIN(F)$ si ottiene partendo da $A \rightarrow x \in F$ e
minimizzando $A$ secondo l'ordine $\leq_F$ e poi secondo l'ordine $\subset$

\item un modo possibile di minimizzare secondo $<_F$ e' sottrarre una variabile
per volta dalle sue conseguenze; si definisce $BCN(A,F)$ come l'insieme delle
variabili implicate da $A$ usando $F$; se $BCN(A,F) \backslash \{a,x\}$ implica
$x$ allora $BCN(A,F) \backslash \{a,x\} <_F A$ e quindi $A$ non e' minimo
perche' $BCN(A,F) \backslash \{a,x\}$ e' strettamente minore; questo funziona
se $F$ e' inequivalente

\item poi si deve minimizzare $A$ secondo il contenimento, ma questo e' facile
verificarlo togliendo una variabile per volta da $A$ e vedendo se il risultato
ancora implica $x$

\item entrambe le verifiche si possono fare usando $BCN(A,F)$, l'insieme di
tutte le variabili implicate da $A$ secondo $F$; ma poi si vedra' che e' anche
utile l'insieme delle conseguenze reali $RCN(A,F)$, che sono le variabili che
si derivano da $A$ con almeno un passo di derivazione; la differenza e' che le
variabili di $A$ stesso ci sono solo se sono esse stesse conseguenze; quindi
$BCN(A,F) = A \cup RCN(A,F)$

\item viene mostrato un algoritmo di calcolo di $RCN(A,F)$; l'algoritmo produce
anche
{} $\{B' \rightarrow x \in F \mid B' \subseteq BCN(B,F)\}$,
le clausole attive con $A$; si dimostra che ci si puo' limitare a questo
insieme quando si cercano le variabili implicate da $A$; questi due insiemi
servono anche nelle sezioni successive

\item algoritmo di calcolo di $SHMIN(F)$, quello poi usato nel programma; usa
questi ultimi risultati

\item viene considerato se esiste una condizione sufficiente per
non-single-head in modo da interromperlo prima; ma non basta vedere se un corpo
genera due sostituzioni minori inconfrontabili; si puo' fare solo cercando
dall'inizio una clausola minima e poi confrontando ogni volta quella con quelle
ottenute; probabilmente e' lo stesso che cercare un minimo dall'inizio

\item dato che il problema e' su corpi equivalenti la condizione di single-head
viene estesa accettando casi del genere; ma forget e' esponenziale in questo
caso

\item descrizione del programma python

il programma di verifica di single-head equivalenza (singlehead.py) la funzione
shmin(), corretta ma incompleta; implementa il sistema descritto in order.tex;
non presenta particolari problemi di implementazione

\end{itemize}

\end{enumerate}

\enddraft

\begin{abstract}

Logical forgetting is \np-complete even in the simple case of propositional
Horn formulae, and may exponentially increase their size. A way to forget is to
replace each variable to forget with the body of each clause whose head is the
variable. It takes polynomial time in the single-head case: each variable is at
most the head of a clause. Some formulae are not single-head but can be made so
to simplify forgetting. They are single-head equivalent. The first contribution
of this article is the study of a semantical characterization of single-head
equivalence. Two necessary conditions are given. They are sufficient when the
formula is inequivalent: it makes two sets of variables equivalent only if they
are also equivalent to their intersection. All acyclic formulae are
inequivalent. The second contribution of this article is an incomplete
algorithm for turning a formula single-head. In case of success, forgetting
becomes possible in polynomial time and produces a polynomial-size formula,
none of which is otherwise guaranteed. The algorithm is complete on
inequivalent formulae.

\end{abstract}

{\bf Keywords:}
Logical forgetting,
knowledge representation,
logic minimizazion,
computational complexity.

\section{Introduction}

Logical forgetting is restricting a formula to a part of the
language~\cite{eite-kern-19}. It is done for reducing the size needed to store
the formula~\cite{eite-kern-19}, to increase the efficiency of reasoning on
it~\cite{erde-ferr-07,wang-etal-05}, to clarify its semantics and the relations
between the conditions it predicates about~\cite{delg-17}, to formalize
reasoning by agents of limited memory~\cite{fagi-etal-95,raja-etal-14}, for
implementing privacy~\cite{gonc-etal-17} and for tailoring the formula to a
specific application~\cite{eite-06}.

Logical forgetting is defined for many logics such as propositional
logic~\cite{bool-54,moin-07,delg-17}, answer set
programming~\cite{wang-etal-14,gonc-etal-16}, description
logics~\cite{kone-etal-09,eite-06}, first-order
logic~\cite{lin-reit-94,zhou-zhan-11}, modal logics~\cite{vand-etal-09}, logics
about actions~\cite{erde-ferr-07,raja-etal-14} defeasible
logic~\cite{anto-etal-12} and belief revision~\cite{naya-chen-lin-07}.

For propositional logic, forgetting is removing some variables from a formula
while leaving the same consequences on the others. It is \np-complete even in
one of the simplest cases: propositional definite Horn~\cite{libe-20-a}.

A way to forget variables from a definite Horn formula is to recursively
replace them~\cite{libe-20-a}. Forgetting from general Horn formulae can be
done by turning the formula definite Horn before forgetting and adding some
clauses afterwards~\cite{libe-20-a}. Therefore, while this article concentrates
on definite Horn formulae, the results apply to the general Horn case. In
particular, it shows how efficiency increases by modifying the input formula
before running the replacement algorithm. This enlarges the restriction that
makes the algorithm polynomial and decreases its running time in other cases.

The replacing algorithm~\cite{libe-20-a} forgets $a$ by replacing it with $bc$
if the formula contains $bc \rightarrow a$; this replacement turns $ad
\rightarrow e$ into $bcd \rightarrow e$.

If a definite Horn formula contains two clauses with head $a$, both their
bodies are possible replacements. A single clause $ad \rightarrow e$ becomes
two. If $d$ is also to be forgotten, it is replaced as well. Again, two clauses
with head $d$ make each clause two. The total is four.

Multiple clauses with the same head may produce a result that is exponentially
large. Their absence guarantees polynomiality.

This is a polynomial restriction: each variable is the head of at most one
clause. Otherwise? Forget is semantical: it produces a formula with the same
consequences on a given alphabet~\cite{delg-17,libe-20-a}. It is the same on
all formulae equivalent to the given one. Among them, one may be single-head
even if the given one was not.

For example, $F = \{a \rightarrow b, b \rightarrow c, a \rightarrow c\}$ is not
single-head because $c$ is the head of two clauses. Formulae with variables in
multiple heads may extort exponential time from the replacing algorithm. Yet,
this particular formula is equivalent to $F = \{a \rightarrow b, b \rightarrow
c\}$, which is single-head.

\

\noindent\hfill
\setlength{\unitlength}{5000sp}%
\begingroup\makeatletter\ifx\SetFigFont\undefined%
\gdef\SetFigFont#1#2#3#4#5{%
  \reset@font\fontsize{#1}{#2pt}%
  \fontfamily{#3}\fontseries{#4}\fontshape{#5}%
  \selectfont}%
\fi\endgroup%
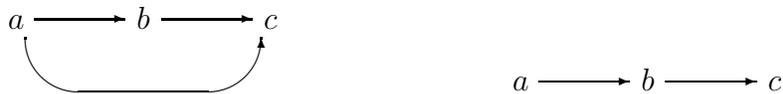
\begin{picture}(1290,474)(5656,-4120)
\thinlines
{\color[rgb]{0,0,0}\put(6301,-3841){\oval(1170,540)[bl]}
\put(6301,-3841){\oval(1170,540)[br]}
\put(6886,-3841){\vector( 0, 1){0}}
}%
{\color[rgb]{0,0,0}\put(5761,-3751){\vector( 1, 0){450}}
}%
{\color[rgb]{0,0,0}\put(6391,-3751){\vector( 1, 0){450}}
}%
\put(5671,-3796){\makebox(0,0)[b]{\smash{{\SetFigFont{12}{24.0}
{\rmdefault}{\mddefault}{\updefault}{\color[rgb]{0,0,0}$a$}%
}}}}
\put(6301,-3796){\makebox(0,0)[b]{\smash{{\SetFigFont{12}{24.0}
{\rmdefault}{\mddefault}{\updefault}{\color[rgb]{0,0,0}$b$}%
}}}}
\put(6931,-3796){\makebox(0,0)[b]{\smash{{\SetFigFont{12}{24.0}
{\rmdefault}{\mddefault}{\updefault}{\color[rgb]{0,0,0}$c$}%
}}}}
\end{picture}%
\nop{
a ---> b ---> c
 \            ^
   \        /
     +----+
}
\hfill
\setlength{\unitlength}{5000sp}%
\begingroup\makeatletter\ifx\SetFigFont\undefined%
\gdef\SetFigFont#1#2#3#4#5{%
  \reset@font\fontsize{#1}{#2pt}%
  \fontfamily{#3}\fontseries{#4}\fontshape{#5}%
  \selectfont}%
\fi\endgroup%
\begin{picture}(1290,165)(5656,-3811)
\thinlines
{\color[rgb]{0,0,0}\put(5761,-3751){\vector( 1, 0){450}}
}%
{\color[rgb]{0,0,0}\put(6391,-3751){\vector( 1, 0){450}}
}%
\put(5671,-3796){\makebox(0,0)[b]{\smash{{\SetFigFont{12}{24.0}
{\rmdefault}{\mddefault}{\updefault}{\color[rgb]{0,0,0}$a$}%
}}}}
\put(6301,-3796){\makebox(0,0)[b]{\smash{{\SetFigFont{12}{24.0}
{\rmdefault}{\mddefault}{\updefault}{\color[rgb]{0,0,0}$b$}%
}}}}
\put(6931,-3796){\makebox(0,0)[b]{\smash{{\SetFigFont{12}{24.0}
{\rmdefault}{\mddefault}{\updefault}{\color[rgb]{0,0,0}$c$}%
}}}}
\end{picture}%
\nop{
a ---> b ---> c
}
\hfill\break

\

Forgetting from two equivalent formulae is the same. Except that it only takes
polynomial time on single-head formulae like the second. An algorithm may first
convert the given formula into a single-head formula if possible, and then
perform forgetting. Equivalence guarantees that the result is the same. The
single-head form guarantees that the running time is polynomial. Correctness
and efficiency, what else? The cost of conversion: only if polynomial, the
overall method of converting and then forgetting is polynomial-time.

This article is about efficiently turning a formula into single-head form if
possible.

Such a translation is the first step of an improved algorithm for forgetting.
Let $B$ be a single-head formula equivalent to $A$ if any, otherwise $A$
itself. Either way, these formulae are equivalent. Since forgetting is
independent on the syntax, it is the same on $A$ and $B$. What changes is the
running time: the original algorithm may take exponential time on formulae that
are single-head equivalent but not single-head; the modified algorithm takes
polynomial time on them, if computing the single-head form takes polynomial
time.

If that takes polynomial time.

Whether it does is an open question.

This article attempts to solve it: given a formula, find an equivalent formula
that does not contain two clauses with the same head, if any.

Just following the formulation of this question and looping over the equivalent
formulae does not work because of their number. A reformulation would help,
something about the formula itself and its clauses. Something similar exists
for similar questions. For example, whether a formula is equivalent to a Horn
formula can be established by checking a certain condition over every pair of
its models~\cite{mcki-43}. This is the aim of Section~\ref{insidious}: find
conditions that express single-head equivalence but do not explicitly refer to
equivalent formulae. The problem seems easy at a first glance. It is not. Loops
of clauses complicates it. Two conditions are easily shown necessary to
single-head equivalence, but a cyclic formula disproves their sufficiency.

Cyclic formulae proved problematic from the very beginning. Even the very first
attempt at reformulating single-head equivalence fails only on formulae that
contain loops. This motivates a further analysis of cyclicity in Horn clauses.
The concept has been studied by Hammer and Kogan~\cite{hamm-koga-95} in the
context of formula minimization. Not surprisingly, this is another problem on
equivalent formulae: find an equivalent formula that contains a minimal number
of clauses or literal occurrences. Not surprisingly, it turns out easy on
acyclic formulae. Section~\ref{cyclic} investigates the effect of acyclicity on
single-head equivalence. First, acyclicity can be defined syntactically and
semantically. The first relates single-head equivalence with irredundancy. The
second makes single-head equivalence the same as the first condition that was
attempted at expressing it --- without success in general. Semantical
acyclicity is a subcase of inequivalence: a formula is inequivalent if every
two sets of variables it makes equivalent are also equivalent to their
intersection. Every inequivalent formula is also acyclic, but not the other way
around. Therefore, inequivalence extends acyclicity. It is a significant
extension because forgetting on inequivalent formulae can be done in polynomial
time.

If a formula is inequivalent, it can be turned into single-head form if
possible by a polynomial time algorithm. Running time is polynomial even in the
general case, but success is no longer guaranteed: a single-head form of a
formula may not be found even if one exists. The algorithm is based on ordering
the bodies of the clauses entailed by the formula. If the formula is
single-head, the only clause of a given head has a minimal body according to
both this ordering and set containment. This is not yet polynomial since
minimal bodies may be exponentially many. The solution is to only search for a
single minimal body for each head. When the formula is inequivalent, this can
be done efficiently. This is the polynomial algorithm described in
Section~\ref{order} along with its optimization and implementation, available
from
{} {\tt https://github.com/paololiberatore/singlehead}.
It always runs in polynomial time and is always correct, and is complete in the
inequivalent case. In general, it may fail to produce the single-head version
of a formula even if one exists. Another article~\cite{libe-20-c} presents an
algorithm that is always able to find the single-head form of a formula if any,
but does not always run in polynomial time.

The content of this article can be summarized in a single paragraph. Logical
forgetting may take exponential time even for Horn theories. Running time
decreases to polynomial if the formula is not only Horn but also single-head:
each variable is the head of at most one clause. Some formulae are not
single-head but can be made so; they are called single-head equivalent. Two
necessary but insufficient conditions to single-head equivalence are given;
they become sufficient when the formula meets a certain condition called
inequivalence, which extends the acyclicity of the clauses. An algorithm for
turning a formula single-head if possible is given; it is incomplete in general
but complete on inequivalent formulae.

The main contribution of this article are:

\begin{itemize}

\item the study of single-head equivalence, the condition of a formula being
equivalent to a single-head one, which are easy for forgetting;

\item the incomplete algorithm for translating a formula in single-head form if
possible, which makes it easy for forgetting.

\end{itemize}

All of this is shown in the following parts of the article.
Section~\ref{insidious} shows that converting a formula in single-head form if
possible is not as easy as it may look; rather the contrary: even just
establishing the existence of a single-head equivalent formula is complicated.
The culprits are the cycles of clauses, as the problem becomes easy when they
are not present; Section~\ref{cyclic} analyzes the case of acyclic formulae and
their generalization to inequivalent formulae. Section~\ref{order} introduces
an order between sets of variables and uses it in a sound but incomplete
algorithm for turning a formula in single-head form; the algorithm is
implemented in the Python language~\cite{vanr-drak-11}.

\section{Preliminaries}
\label{preliminaries}

All formulae considered in this article are propositional and definite Horn:
they are set of Horn clauses, each Horn clause being the disjunction of
literals where exactly one is positive, where a positive literal is a variable
and a negative literal is the negation of a variable. The positive literal is
mandatory: clauses made only of negative literals are forbidden. Because of the
definite Horn form, clauses can be written in a simplified form like $abc
\rightarrow d$, instead of $\neg a \vee \neg b \vee \neg c \vee d$ or $(a
\wedge b \wedge c) \rightarrow d$.

The implicates of a formula $F$ are the clauses it entails.
Its prime implicates $PI(F)$ are the implicates that are no longer entailed if
removed any literal.

Forgetting is removing variables while maintaining the implications of a
formula on the others. It is not unique because of equivalence. This is way it
is not defined as "forgetting is this" but rather "forgetting is expressed by
this".

\begin{definition}
\label{forget}

A formula $B$ expresses forgetting the variables $X$ from a formula $A$ if $B$
only mentions the variables of $A$ not in $X$ and every formula on these
variables that is entailed by $A$ is also entailed by $B$ and vice versa.

\end{definition}

A way to forget a variable $x$ from $A$ is to replace $x$ with $\true$ in $A$,
then with $\false$, and then disjoin the results~\cite{bool-54}. The resulting
formula is exponentially larger than the original when forgetting multiple
variables, and is not Horn even if $A$ is. It is not even a set of clauses any
longer.

Another way to forget variables is to resolve them
out~\cite{gott-87,delg-wass-13,wang-15,delg-17}: combine each clause where a
variable is positive with each where it is negative. This algorithm may
unnecessarily require exponential space~\cite{libe-20-a}.

This is avoided by instead replacing each variable with each of the bodies of
the clauses that have it as their head~\cite{libe-20-a}. This can be seen as
constraining the order of resolution of the clauses in a way which limits the
required working space. If each variable is the head of at most a clause, the
algorithm is guaranteed to take only polynomial time and produce a
polynomially-sized result~\cite{libe-20-a}. Turning a formula in this form if
possible makes forgetting efficient in both time and space.

\begin{definition}
\label{single-head}

A formula $F = \{C_1,\ldots,C_m\}$ is {\em single-head} if every variable $x$
is the head of at most one clause:

\[
\forall x ~.~
	\not\exists i,j ~.~
		i \neq j \mbox{ and }
		x \in C_i \mbox{ and }
		x \in C_j
\]

\end{definition}

\begin{definition}
\label{single-head-equivalent}

A formula $F$ is {\em single-head} equivalent if it is equivalent to a formula
that is single-head:

\[
\exists F' ~.~ F \equiv F' \mbox{ and } F \mbox{ is single-head}
\]

\end{definition}

A restriction where checking whether this is possible is that of inequivalence.
A formula $F$ is inequivalent if it makes equivalent only sets of variables
that are equivalent to their intersection.

\begin{definition}
\label{inequivalent}

A formula $F$ is {\em inequivalent}
if $F \models A \equiv B$ implies $F \models A \equiv (A \cap B)$
for every pair of sets of variables $A$ and $B$.

\end{definition}

An implicate of a formula is a clause that the formula entails. A prime
implicate is an implicate such that no subclause is also an implicate.

For every formula $F$ and variable $x$, the formula $F^x$ is the set of clauses
of $F$ that do not contain $x$ with either sign. For example, if
{} $F = \{a \vee b, b \vee c, \neg a \vee c\}$,
then
{} $F^a = \{b \vee c\}$.
Both $a \vee b$ and $\neg a \vee c$ are removed since they contain $a$.

Other concepts introduced in the following are summarized here for reference.

The ordering $A \leq_F B$ between two sets of variables $A$ and $B$ induced by
the formula $F$ is defined as $F \models B \rightarrow A$; its strict and
counterparts $A <_F B$ is $A \leq_F B$ and $B \not\leq_F A$; its induced
equivalence relation $A \equiv_F B$ is $A \leq_F B$ and $B \leq_F A$.

Given a formula $F$ set of variables $B$, the following sets of variables and
clauses are defined:

\begin{eqnarray*}
BCN(B,F) &=& \{ x \mid F \cup B \models x \}				\\
RCN(B,F) &=& \{ x \mid F \cup (BCN(B,F) \backslash \{x\} \models x\}
\end{eqnarray*}

A comparison with the other articles of this series is in order.
A first article analyzes the complexity of establishing whether forgetting some
variables from a formula can be expressed in a given size~\cite{libe-20}.
A second article shows that forgetting and a restricted form of equivalence
reduce to each other, presents the replacing algorithm for forgetting and
analyzes the problem of minimizing the formula that results from forgetting
when introducing new variables is allowed or not~\cite{libe-20-a}.
A final article presents an algorithm for turning a formula into single-head
form if possible~\cite{libe-20-c}; it is complete but not polynomial-time,
complementing the incomplete but polynomial-time algorithm in the current
article.

\section{The insidious single-head equivalence}
\label{insidious}

The example in the introduction makes converting a formula in single-head form
look easy. It is easy when the formula is simple like $F = \{a \rightarrow b, b
\rightarrow c, a \rightarrow c\}$. All it takes is to remove some clauses:
since $c$ is the head of two clauses, one must be deleted; $F \backslash \{b
\rightarrow c\}$ is not equivalent to $F$, but $F \backslash \{a \rightarrow
c\}$ is. Problem solved: the latter formula is single-head and equivalent to
$F$. This section shows that the translation is not always so easy.

As a matter of fact, even checking for the existence of a single-head
equivalent formula without producing it is complicated. In theory, it could be
done on the set of models. Indeed, two equivalent formulae either have a
single-head equivalent formula both, or they both do not. A similar concept is
that of Horn equivalence: the existence of a Horn formula equivalent to a given
one; it is the case if every two models of the formula intersect to another
model of the same formula~\cite{mcki-43}. Unfortunately, a similarly simple
formulation of single-head equivalence does not seem to exist. Whether it does
not exist, in addition to does not seem to exist, is left as an open problem by
this article. The rest of this section gives some hints suggesting it does not.
These are just hints. What is certain at this point is only that it is an
insidious problem.

\subsection{Where the traps are}

Single-head equivalence is insidious because it looks like it has a simple
semantical formulation, but the following list shows a number of traps in the
intuitive ways it appears to be expressible.

\begin{enumerate}

\item If a formula is single-head, it contains at most one clause $A
\rightarrow x$ for each variable $x$. If $F$ is equivalent to a single-head
formula $F'$, this formula $F'$ may only contain a single clause $A \rightarrow
x$ with $x$ in the head. All other clauses $B \rightarrow x$ of $F$ are
consequences of $A \rightarrow x$. Formally, if $B \rightarrow x \in F$ then $F
\models B \rightarrow A$.

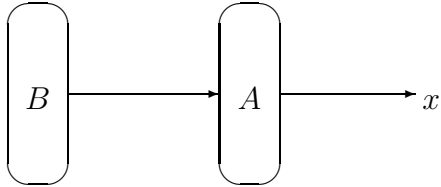
\setlength{\unitlength}{5000sp}%
\begingroup\makeatletter\ifx\SetFigFont\undefined%
\gdef\SetFigFont#1#2#3#4#5{%
  \reset@font\fontsize{#1}{#2pt}%
  \fontfamily{#3}\fontseries{#4}\fontshape{#5}%
  \selectfont}%
\fi\endgroup%
\begin{picture}(2127,924)(7189,-6223)
\thinlines
{\color[rgb]{0,0,0}\put(7306,-6106){\oval(210,210)[bl]}
\put(7306,-5416){\oval(210,210)[tl]}
\put(7396,-6106){\oval(210,210)[br]}
\put(7396,-5416){\oval(210,210)[tr]}
\put(7306,-6211){\line( 1, 0){ 90}}
\put(7306,-5311){\line( 1, 0){ 90}}
\put(7201,-6106){\line( 0, 1){690}}
\put(7501,-6106){\line( 0, 1){690}}
}%
{\color[rgb]{0,0,0}\put(8356,-6106){\oval(210,210)[bl]}
\put(8356,-5416){\oval(210,210)[tl]}
\put(8446,-6106){\oval(210,210)[br]}
\put(8446,-5416){\oval(210,210)[tr]}
\put(8356,-6211){\line( 1, 0){ 90}}
\put(8356,-5311){\line( 1, 0){ 90}}
\put(8251,-6106){\line( 0, 1){690}}
\put(8551,-6106){\line( 0, 1){690}}
}%
{\color[rgb]{0,0,0}\put(7501,-5761){\vector( 1, 0){750}}
}%
{\color[rgb]{0,0,0}\put(8551,-5761){\vector( 1, 0){675}}
}%
\put(7351,-5836){\makebox(0,0)[b]{\smash{{\SetFigFont{12}{24.0}
{\rmdefault}{\mddefault}{\updefault}{\color[rgb]{0,0,0}$B$}%
}}}}
\put(9301,-5836){\makebox(0,0)[b]{\smash{{\SetFigFont{12}{24.0}
{\rmdefault}{\mddefault}{\updefault}{\color[rgb]{0,0,0}$x$}%
}}}}
\put(8401,-5836){\makebox(0,0)[b]{\smash{{\SetFigFont{12}{24.0}
{\rmdefault}{\mddefault}{\updefault}{\color[rgb]{0,0,0}$A$}%
}}}}
\end{picture}%
\nop{
B --> A --> x
}

The first attempt at formalizing single-head equivalence is that every variable
$x$ has a set of variables $A$ such that $F \models B \rightarrow x$ implies $F
\models B \rightarrow A$. This is the case for every single-head formula by
setting $A$ to the body of the only clause $A \rightarrow x$ with $x$ in the
head. It is also the case for every single-head equivalent formula because the
condition is semantical: it does not depend on the syntax of the formula.

\begin{condition}
\label{first-condition}

For each variable $x$,
	there exists a set $A$ of variables such that
		$x \not\in A$ and such that
		for every set $B$ of variables such that $x \not\in B$,
			$F \models B \rightarrow x$ implies
				$F \models B \rightarrow A$ and
				$F \models A \rightarrow x$.

\end{condition}

Why is $F \models A \rightarrow x$ only required when $F \models B \rightarrow
x$? Otherwise, every variable $x$ would be forced to be the head of a clause $A
\rightarrow x$ entailed by $F$, which may not be the case.

This condition is necessary for $F$ being single-head equivalent, as formally
proved in Lemma~\ref{first-necessary}, below.

\item Every formula that is equivalent to a single-head formula satisfies
Condition~\ref{first-condition}, but not the other way around.
Condition~\ref{first-condition} is not sufficient to ensure equivalence to a
single-head formula. The following is a counterexample, also in the {\tt
inloop.py} test file of the {\tt singlehead.py} program.


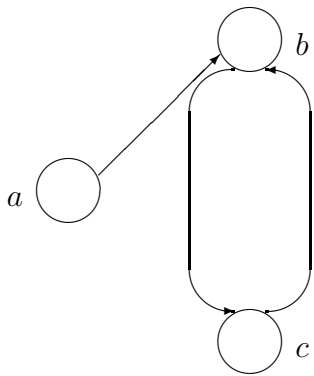
\begin{equation}
\label{inloop}
F = \{a \rightarrow b, b \rightarrow c, c \rightarrow b\}
\end{equation}

\setlength{\unitlength}{5000sp}%
\begingroup\makeatletter\ifx\SetFigFont\undefined%
\gdef\SetFigFont#1#2#3#4#5{%
  \reset@font\fontsize{#1}{#2pt}%
  \fontfamily{#3}\fontseries{#4}\fontshape{#5}%
  \selectfont}%
\fi\endgroup%
\begin{picture}(1449,1850)(6811,-6536)
\thinlines
{\color[rgb]{0,0,0}\put(7876,-5611){\oval(450,1200)[tl]}
\put(7876,-5611){\oval(450,1200)[bl]}
\put(7876,-6211){\vector( 1, 0){0}}
}%
{\color[rgb]{0,0,0}\put(8026,-5611){\oval(450,1200)[br]}
\put(8026,-5611){\oval(450,1200)[tr]}
\put(8026,-5011){\vector(-1, 0){0}}
}%
{\color[rgb]{0,0,0}\put(7051,-5611){\circle{336}}
}%
{\color[rgb]{0,0,0}\put(7951,-4861){\circle{336}}
}%
{\color[rgb]{0,0,0}\put(7951,-6361){\circle{336}}
}%
{\color[rgb]{0,0,0}\put(7201,-5536){\vector( 1, 1){600}}
}%
\put(6826,-5686){\makebox(0,0)[rb]{\smash{{\SetFigFont{12}{24.0}
{\rmdefault}{\mddefault}{\updefault}{\color[rgb]{0,0,0}$a$}%
}}}}
\put(8176,-4936){\makebox(0,0)[lb]{\smash{{\SetFigFont{12}{24.0}
{\rmdefault}{\mddefault}{\updefault}{\color[rgb]{0,0,0}$b$}%
}}}}
\put(8176,-6436){\makebox(0,0)[lb]{\smash{{\SetFigFont{12}{24.0}
{\rmdefault}{\mddefault}{\updefault}{\color[rgb]{0,0,0}$c$}%
}}}}
\end{picture}%
\nop{
a -> b <-> c
}

This formula is not equivalent to any single-head formula. The proof is by
contradiction: a single-head definite Horn formula $F'$ is assumed equivalent
to $F$. Since $F' \models b \rightarrow c$, Lemma~\ref{set-implies-set} tells
that $F'$ contains a clause $A \rightarrow c$ such that $F' \models \{b\}
\rightarrow A$. The only variables $b$ implies are itself and $c$. Therefore,
either $A$ is $\{b\}$ or is $\{c\}$. The second is ruled out as $A \rightarrow
c$ is a tautology. As a result, $F'$ contains $b \rightarrow c$. For the same
reason, it also contains $c \rightarrow b$. Since $F'$ is by assumption
single-head, it does not contain any other clause with $b$ or $c$ in the head.
The only other clauses it may contain have head $a$. These are $b \rightarrow
a$, $c \rightarrow a$ and $bc \rightarrow a$. None of them is entailed by $F$.
As a result, $F'$ is $\{b \rightarrow c, c \rightarrow b\}$, which is not
equivalent to $F$.

In spite of $F$ being equivalent to no single-head formula, it satisfies
Condition~\ref{first-condition} with the set $\{c\}$ for $b$, the set $\{b\}$
for $c$ and the set $\emptyset$ for $a$. The first set $\{c\}$ is valid for $b$
because the non-tautological entailed clauses with $b$ as their heads are $a
\rightarrow b$ and $c \rightarrow b$; regarding the first, $F$ entails both $a
\rightarrow \{c\}$ and $\{c\} \rightarrow b$; regarding the second, $F$ entails
both $c \rightarrow \{c\}$ and again $\{c\} \rightarrow b$. The same argument
proves that $\{b\}$ is valid for $c$ by symmetry. For $a$ the condition is
trivially satisfied because $a$ is the head of no non-tautological clause
entailed by $F$.

Condition~\ref{first-condition} suggests that $a \rightarrow b$ is redundant in
$F$ thanks to $A \rightarrow x$, which is $a \rightarrow \{c\}$ in this case.
Yet, $a \rightarrow \{c\}$ only holds because of $a \rightarrow b \in F$.

\item The previous point proves that Condition~\ref{first-condition} is not
sufficient to single-head equivalence. It correctly states that $B \rightarrow
x$ is a consequence of $B \rightarrow A$ and $A \rightarrow x$, but neglects
the case where $B \rightarrow A$ only holds as a consequence of $B \rightarrow
x$. Lemma~\ref{set-implies-set} allows cutting this loop: it proves that
{} $F \models B \rightarrow x$
is the same as
{} $F^x \models B \rightarrow A'$
where $A' \rightarrow x$ is a clause of $F$ and $F^x$ is the set of clauses of
$F$ that do not contain $x$. Because of the single-heads, $A'$ is the same as
$A$.

\[
\forall x \exists A ~.~
F \models B \rightarrow x
~~~ \Rightarrow ~~~
F^x \models B \rightarrow A ,~ F \models A \rightarrow x
\]

This condition ensures that $B \rightarrow A$ is not itself a consequence of $A
\rightarrow x$, since $F^x$ does not contain this clause.

Unfortunately, $F^x$ is a syntactic construction: it is a subset of $F$. The
condition holds for all single-head formulae but not all their equivalent
formulae. The following is a counterexample.

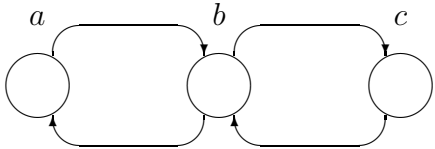
\[
F = \{a \rightarrow b, b \rightarrow a, b \rightarrow c, c \rightarrow b\}
\]

\setlength{\unitlength}{5000sp}%
\begingroup\makeatletter\ifx\SetFigFont\undefined%
\gdef\SetFigFont#1#2#3#4#5{%
  \reset@font\fontsize{#1}{#2pt}%
  \fontfamily{#3}\fontseries{#4}\fontshape{#5}%
  \selectfont}%
\fi\endgroup%
\begin{picture}(2152,774)(6575,-5920)
\thinlines
{\color[rgb]{0,0,0}\put(7576,-5461){\vector( 0,-1){0}}
\put(7201,-5461){\oval(750,300)[tr]}
\put(7201,-5461){\oval(750,300)[tl]}
}%
{\color[rgb]{0,0,0}\put(6826,-5761){\vector( 0, 1){0}}
\put(7201,-5761){\oval(750,300)[bl]}
\put(7201,-5761){\oval(750,300)[br]}
}%
{\color[rgb]{0,0,0}\put(8476,-5461){\vector( 0,-1){0}}
\put(8101,-5461){\oval(750,300)[tr]}
\put(8101,-5461){\oval(750,300)[tl]}
}%
{\color[rgb]{0,0,0}\put(7726,-5761){\vector( 0, 1){0}}
\put(8101,-5761){\oval(750,300)[bl]}
\put(8101,-5761){\oval(750,300)[br]}
}%
{\color[rgb]{0,0,0}\put(6751,-5611){\circle{336}}
}%
{\color[rgb]{0,0,0}\put(7651,-5611){\circle{336}}
}%
{\color[rgb]{0,0,0}\put(8551,-5611){\circle{336}}
}%
\put(6751,-5311){\makebox(0,0)[b]{\smash{{\SetFigFont{12}{24.0}
{\rmdefault}{\mddefault}{\updefault}{\color[rgb]{0,0,0}$a$}%
}}}}
\put(7651,-5311){\makebox(0,0)[b]{\smash{{\SetFigFont{12}{24.0}
{\rmdefault}{\mddefault}{\updefault}{\color[rgb]{0,0,0}$b$}%
}}}}
\put(8551,-5311){\makebox(0,0)[b]{\smash{{\SetFigFont{12}{24.0}
{\rmdefault}{\mddefault}{\updefault}{\color[rgb]{0,0,0}$c$}%
}}}}
\end{picture}%
\nop{
a <--> b <--> c
}

Removing all clauses containing $b$ results in an empty set $F^b = \emptyset$,
which entails neither $a \rightarrow \{c\}$ nor $c \rightarrow \{a\}$;
therefore, neither $\{c\}$ nor $\{a\}$ are valid sets for $b$. Removing only
the clauses with $b$ in the head gives the same outcome:
{} $\{b \rightarrow a, b \rightarrow c\}$
entails neither $a \rightarrow \{c\}$ nor $c \rightarrow \{a\}$.

Yet, an equivalent single-head formula exists:

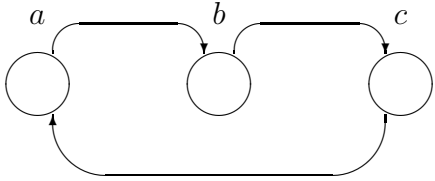
\[
F' = \{a \rightarrow b, b \rightarrow c, c \rightarrow a\}
\]

\setlength{\unitlength}{5000sp}%
\begingroup\makeatletter\ifx\SetFigFont\undefined%
\gdef\SetFigFont#1#2#3#4#5{%
  \reset@font\fontsize{#1}{#2pt}%
  \fontfamily{#3}\fontseries{#4}\fontshape{#5}%
  \selectfont}%
\fi\endgroup%
\begin{picture}(2152,924)(6575,-6070)
\thinlines
{\color[rgb]{0,0,0}\put(7576,-5461){\vector( 0,-1){0}}
\put(7201,-5461){\oval(750,300)[tr]}
\put(7201,-5461){\oval(750,300)[tl]}
}%
{\color[rgb]{0,0,0}\put(8476,-5461){\vector( 0,-1){0}}
\put(8101,-5461){\oval(750,300)[tr]}
\put(8101,-5461){\oval(750,300)[tl]}
}%
{\color[rgb]{0,0,0}\put(6826,-5761){\vector( 0, 1){0}}
\put(7651,-5761){\oval(1650,600)[bl]}
\put(7651,-5761){\oval(1650,600)[br]}
}%
{\color[rgb]{0,0,0}\put(6751,-5611){\circle{336}}
}%
{\color[rgb]{0,0,0}\put(7651,-5611){\circle{336}}
}%
{\color[rgb]{0,0,0}\put(8551,-5611){\circle{336}}
}%
\put(6751,-5311){\makebox(0,0)[b]{\smash{{\SetFigFont{12}{24.0}
{\rmdefault}{\mddefault}{\updefault}{\color[rgb]{0,0,0}$a$}%
}}}}
\put(7651,-5311){\makebox(0,0)[b]{\smash{{\SetFigFont{12}{24.0}
{\rmdefault}{\mddefault}{\updefault}{\color[rgb]{0,0,0}$b$}%
}}}}
\put(8551,-5311){\makebox(0,0)[b]{\smash{{\SetFigFont{12}{24.0}
{\rmdefault}{\mddefault}{\updefault}{\color[rgb]{0,0,0}$c$}%
}}}}
\end{picture}%
\nop{
a --> b --> c
^           |
+-----------+
}

This shows that syntactic nature of $F^x$ has practical effects, as its usage
to recognize single-head equivalence may be incorrect.

\item That the counterexample involves equivalence is not an incident.
Equivalence is the root of all problems of Condition~\ref{first-condition}. If
$B \rightarrow x$ is a consequence of $B \rightarrow A$ and $A \rightarrow x$,
it is redundant and can be removed when no equivalences are present.

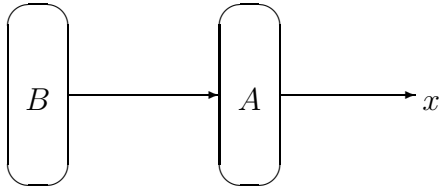
\setlength{\unitlength}{5000sp}%
\begingroup\makeatletter\ifx\SetFigFont\undefined%
\gdef\SetFigFont#1#2#3#4#5{%
  \reset@font\fontsize{#1}{#2pt}%
  \fontfamily{#3}\fontseries{#4}\fontshape{#5}%
  \selectfont}%
\fi\endgroup%
\begin{picture}(2127,924)(7189,-6223)
\thinlines
{\color[rgb]{0,0,0}\put(7306,-6106){\oval(210,210)[bl]}
\put(7306,-5416){\oval(210,210)[tl]}
\put(7396,-6106){\oval(210,210)[br]}
\put(7396,-5416){\oval(210,210)[tr]}
\put(7306,-6211){\line( 1, 0){ 90}}
\put(7306,-5311){\line( 1, 0){ 90}}
\put(7201,-6106){\line( 0, 1){690}}
\put(7501,-6106){\line( 0, 1){690}}
}%
{\color[rgb]{0,0,0}\put(8356,-6106){\oval(210,210)[bl]}
\put(8356,-5416){\oval(210,210)[tl]}
\put(8446,-6106){\oval(210,210)[br]}
\put(8446,-5416){\oval(210,210)[tr]}
\put(8356,-6211){\line( 1, 0){ 90}}
\put(8356,-5311){\line( 1, 0){ 90}}
\put(8251,-6106){\line( 0, 1){690}}
\put(8551,-6106){\line( 0, 1){690}}
}%
{\color[rgb]{0,0,0}\put(7501,-5761){\vector( 1, 0){750}}
}%
{\color[rgb]{0,0,0}\put(8551,-5761){\vector( 1, 0){675}}
}%
\put(7351,-5836){\makebox(0,0)[b]{\smash{{\SetFigFont{12}{24.0}
{\rmdefault}{\mddefault}{\updefault}{\color[rgb]{0,0,0}$B$}%
}}}}
\put(9301,-5836){\makebox(0,0)[b]{\smash{{\SetFigFont{12}{24.0}
{\rmdefault}{\mddefault}{\updefault}{\color[rgb]{0,0,0}$x$}%
}}}}
\put(8401,-5836){\makebox(0,0)[b]{\smash{{\SetFigFont{12}{24.0}
{\rmdefault}{\mddefault}{\updefault}{\color[rgb]{0,0,0}$A$}%
}}}}
\end{picture}%
\nop{
B --> A --> x
}

Redundancy disappears only when $B \rightarrow A$ is itself a consequence of $B
\rightarrow x$, but this implies an equivalence between sets of variables.

This is the case when $B$ implies $x$ and some other variables $C$ that
together imply $A$. Formally,
{} $F \models B \rightarrow C \cup \{x\}$
and
{} $F \models C \cup \{x\} \rightarrow A$.
Graphically, a two-tail arrow goes from $C$ and $x$ to $A$.

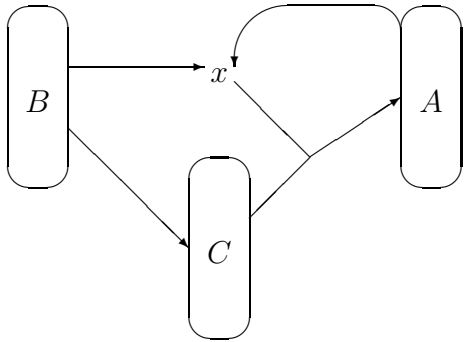
\setlength{\unitlength}{5000sp}%
\begingroup\makeatletter\ifx\SetFigFont\undefined%
\gdef\SetFigFont#1#2#3#4#5{%
  \reset@font\fontsize{#1}{#2pt}%
  \fontfamily{#3}\fontseries{#4}\fontshape{#5}%
  \selectfont}%
\fi\endgroup%
\begin{picture}(2274,1679)(7189,-6973)
\thinlines
{\color[rgb]{0,0,0}\put(8787,-5461){\oval(728,314)[tr]}
\put(8787,-5611){\oval(922,614)[tl]}
\put(8326,-5611){\vector( 0,-1){0}}
}%
{\color[rgb]{0,0,0}\put(7306,-6106){\oval(210,210)[bl]}
\put(7306,-5416){\oval(210,210)[tl]}
\put(7396,-6106){\oval(210,210)[br]}
\put(7396,-5416){\oval(210,210)[tr]}
\put(7306,-6211){\line( 1, 0){ 90}}
\put(7306,-5311){\line( 1, 0){ 90}}
\put(7201,-6106){\line( 0, 1){690}}
\put(7501,-6106){\line( 0, 1){690}}
}%
{\color[rgb]{0,0,0}\put(9256,-6106){\oval(210,210)[bl]}
\put(9256,-5416){\oval(210,210)[tl]}
\put(9346,-6106){\oval(210,210)[br]}
\put(9346,-5416){\oval(210,210)[tr]}
\put(9256,-6211){\line( 1, 0){ 90}}
\put(9256,-5311){\line( 1, 0){ 90}}
\put(9151,-6106){\line( 0, 1){690}}
\put(9451,-6106){\line( 0, 1){690}}
}%
{\color[rgb]{0,0,0}\put(7501,-5611){\vector( 1, 0){675}}
}%
{\color[rgb]{0,0,0}\put(8206,-6856){\oval(210,210)[bl]}
\put(8206,-6166){\oval(210,210)[tl]}
\put(8296,-6856){\oval(210,210)[br]}
\put(8296,-6166){\oval(210,210)[tr]}
\put(8206,-6961){\line( 1, 0){ 90}}
\put(8206,-6061){\line( 1, 0){ 90}}
\put(8101,-6856){\line( 0, 1){690}}
\put(8401,-6856){\line( 0, 1){690}}
}%
{\color[rgb]{0,0,0}\put(8326,-5686){\line( 1,-1){375}}
\put(8701,-6061){\line(-1,-1){300}}
}%
{\color[rgb]{0,0,0}\put(8701,-6061){\vector( 3, 2){450}}
}%
{\color[rgb]{0,0,0}\put(7501,-5911){\vector( 1,-1){600}}
}%
\put(7351,-5836){\makebox(0,0)[b]{\smash{{\SetFigFont{12}{24.0}
{\rmdefault}{\mddefault}{\updefault}{\color[rgb]{0,0,0}$B$}%
}}}}
\put(9301,-5836){\makebox(0,0)[b]{\smash{{\SetFigFont{12}{24.0}
{\rmdefault}{\mddefault}{\updefault}{\color[rgb]{0,0,0}$A$}%
}}}}
\put(8251,-5686){\makebox(0,0)[b]{\smash{{\SetFigFont{12}{24.0}
{\rmdefault}{\mddefault}{\updefault}{\color[rgb]{0,0,0}$x$}%
}}}}
\put(8251,-6586){\makebox(0,0)[b]{\smash{{\SetFigFont{12}{24.0}
{\rmdefault}{\mddefault}{\updefault}{\color[rgb]{0,0,0}$C$}%
}}}}
\end{picture}%
\nop{
      +---------+
      V         |
B --> x ---+    |
|          +--> A
+---> C ---+
}

Since $F \models A \rightarrow x$, the equivalence
{} $F \models C \cup \{x\} \equiv C \cup A$
holds. This proves that the problematic cases involve equivalences.

\item Equivalences do not always forbid single-head equivalence. For example,
{} $\{a \rightarrow b, b \rightarrow a, b \rightarrow c, c \rightarrow b\}$
implies the equivalence of $a$, $b$ and $c$, yet it is equivalent to the
single-head formula
{} $\{a \rightarrow b, b \rightarrow c, c \rightarrow a\}$.
Equivalences are realized by loops of clauses, like in this case.

What single-head equivalence forbids is that variables outside the loop entail
variables in the loop. This was exactly what happened in the counterexample
{} $F = \{a \rightarrow b, b \rightarrow c, c \rightarrow b\}$,
where $b$ and $c$ are equivalent and $a$ entails one of them without being
equivalent to them.

The converse is not a problem: a variable in a loop may entail a variable
outside it.

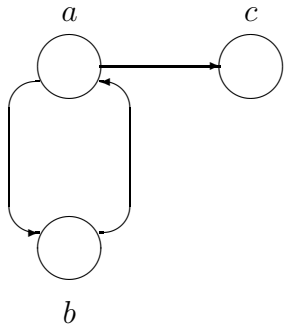
\[
\{a \rightarrow b, b \rightarrow a, a \rightarrow c\}
\]

\setlength{\unitlength}{5000sp}%
\begingroup\makeatletter\ifx\SetFigFont\undefined%
\gdef\SetFigFont#1#2#3#4#5{%
  \reset@font\fontsize{#1}{#2pt}%
  \fontfamily{#3}\fontseries{#4}\fontshape{#5}%
  \selectfont}%
\fi\endgroup%
\begin{picture}(1385,1635)(7042,-6451)
\thinlines
{\color[rgb]{0,0,0}\put(7201,-5611){\oval(300,750)[tl]}
\put(7201,-5611){\oval(300,750)[bl]}
\put(7201,-5986){\vector( 1, 0){0}}
}%
{\color[rgb]{0,0,0}\put(7501,-5611){\oval(300,750)[br]}
\put(7501,-5611){\oval(300,750)[tr]}
\put(7501,-5236){\vector(-1, 0){0}}
}%
{\color[rgb]{0,0,0}\put(7351,-5161){\circle{336}}
}%
{\color[rgb]{0,0,0}\put(7351,-6061){\circle{336}}
}%
{\color[rgb]{0,0,0}\put(8251,-5161){\circle{336}}
}%
{\color[rgb]{0,0,0}\put(7501,-5161){\vector( 1, 0){600}}
}%
\put(7351,-4936){\makebox(0,0)[b]{\smash{{\SetFigFont{12}{24.0}
{\rmdefault}{\mddefault}{\updefault}{\color[rgb]{0,0,0}$a$}%
}}}}
\put(7351,-6436){\makebox(0,0)[b]{\smash{{\SetFigFont{12}{24.0}
{\rmdefault}{\mddefault}{\updefault}{\color[rgb]{0,0,0}$b$}%
}}}}
\put(8251,-4936){\makebox(0,0)[b]{\smash{{\SetFigFont{12}{24.0}
{\rmdefault}{\mddefault}{\updefault}{\color[rgb]{0,0,0}$c$}%
}}}}
\end{picture}%
\nop{
c <-- a <--> b
}

According to this argument, the second condition for single-head equivalence
would be: if some sets of variables $A$ and $B$ are equivalent to each other
and are entailed by another set $C$, then $C$ is equivalent to them. This is
however too restrictive as a condition, as the next point shows.

\item When equivalences involve only single variables, the second condition
works when modified according to the point above. However, two equivalent sets
of variables may have a common part. As an example, the following formula
contains two equivalent sets $B$ and $C$, yet the non-equivalent set $A$
entails them.

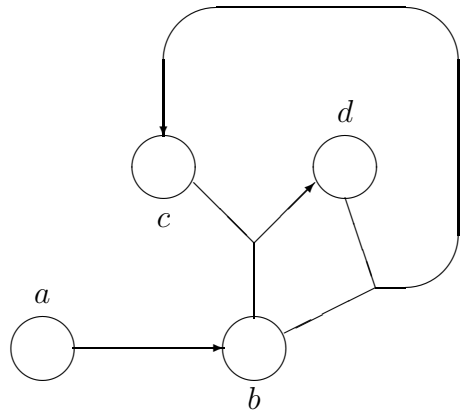
\[
F = \{
a \rightarrow b ,~
b c \rightarrow d ,~
b d \rightarrow c
\}
\]

\setlength{\unitlength}{5000sp}%
\begingroup\makeatletter\ifx\SetFigFont\undefined%
\gdef\SetFigFont#1#2#3#4#5{%
  \reset@font\fontsize{#1}{#2pt}%
  \fontfamily{#3}\fontseries{#4}\fontshape{#5}%
  \selectfont}%
\fi\endgroup%
\begin{picture}(2247,2018)(7025,-6076)
\thinlines
{\color[rgb]{0,0,0}\put(8851,-4801){\oval(824,1320)[br]}
\put(8529,-4801){\oval(1468,1468)[tr]}
\put(8529,-4711){\oval(1456,1288)[tl]}
\put(7801,-4711){\vector( 0,-1){0}}
}%
{\color[rgb]{0,0,0}\put(7201,-5761){\circle{336}}
}%
{\color[rgb]{0,0,0}\put(8251,-5761){\circle{336}}
}%
{\color[rgb]{0,0,0}\put(7801,-4861){\circle{336}}
}%
{\color[rgb]{0,0,0}\put(8701,-4861){\circle{336}}
}%
{\color[rgb]{0,0,0}\put(7351,-5761){\vector( 1, 0){750}}
}%
{\color[rgb]{0,0,0}\put(7951,-4936){\line( 1,-1){300}}
\put(8251,-5236){\line( 0,-1){375}}
}%
{\color[rgb]{0,0,0}\put(8251,-5236){\vector( 1, 1){300}}
}%
{\color[rgb]{0,0,0}\put(8701,-5011){\line( 1,-3){150}}
\put(8851,-5461){\line(-2,-1){450}}
}%
\put(7201,-5536){\makebox(0,0)[b]{\smash{{\SetFigFont{12}{24.0}
{\rmdefault}{\mddefault}{\updefault}{\color[rgb]{0,0,0}$a$}%
}}}}
\put(8251,-6061){\makebox(0,0)[b]{\smash{{\SetFigFont{12}{24.0}
{\rmdefault}{\mddefault}{\updefault}{\color[rgb]{0,0,0}$b$}%
}}}}
\put(7801,-5161){\makebox(0,0)[b]{\smash{{\SetFigFont{12}{24.0}
{\rmdefault}{\mddefault}{\updefault}{\color[rgb]{0,0,0}$c$}%
}}}}
\put(8701,-4636){\makebox(0,0)[b]{\smash{{\SetFigFont{12}{24.0}
{\rmdefault}{\mddefault}{\updefault}{\color[rgb]{0,0,0}$d$}%
}}}}
\end{picture}%
\nop{
      c <------------+
       \             |
        +---> d -+   |
       /         +---+
a --> b  --------+
}

This formula implies the equivalence of $B = \{b,c\}$ and $C = \{b,d\}$. The
set $A = \{a, c\}$ implies them without being equivalent to them. Yet, the
formula is single-head.

The rationale of the condition is that an equivalences between sets of
variables requires their entailment; therefore, their variables cannot be
entailed by other sets. According to this argument, $b$ being in $B$ that is
equivalent to $C$ requires $b$ to be entailed by $C$. As a result, it could not
be entailed by $A$. It can, instead. It can because it is in all equivalent
sets $B$ and $C$. Their equivalence does not require a clause like $C
\rightarrow b$. This leaves $b$ free as a head. The clause $a \rightarrow b$ is
allowed.

\item The second condition for single-head equivalence is that a set of
variables $A$ may entail a set $B$ that is equivalent to some other set $C$,
but only if $A$ is also equivalent to $B$ and $C$, at least regarding the
variables that are not in all sets equivalent to $B$ and $C$. The definitions
of equivalent sets and their common variables is necessary to formalize this
condition:

\begin{eqnarray*}
\equiset(A,F) &=& \{B \mid F \models A \equiv B\}		\\
\equiall(A,F) &=& \{x \mid \forall C \in \equiset(A,F) ~.~ x \in C\}
\end{eqnarray*}

The first definition is all sets of variables that are equivalent to $A$. The
second is all variables that are in all of them. The ``all`` in $\equiall(A,F)$
stands for ``variables in all sets``.

These two concepts allows formalizing the second condition to common
equivalence.

\begin{condition}
\label{second-condition}

If $F \models A \rightarrow B$ then there exists $C \in \equiset(B,F)$ such
that $C \backslash \equiall(B,F) \subseteq A$.

\end{condition}

This condition does not depend on the syntax of the formula since the clause it
mentions only occurs in an entailment, $\equiall()$ is based on $\equiset()$
and $\equiset()$ again only mentions clauses in an entailment. As a result, if
$F \equiv F'$, the condition holds for $F$ if and only if it holds for $F'$.

\end{enumerate}

Unfortunately, even this condition is only necessary to single-head
equivalence. It is not sufficient.

\subsection{Necessary conditions}

Conditions~\ref{first-condition} and~\ref{second-condition} are proved
necessary to single-head equivalence. They are based on entailment, which is
unaffected by the syntax of the formula. Therefore, if they hold for a formula,
they hold for every equivalent one. Therefore, they are not only satisfied by
single-head formulae; they are satisfied by every formula that is equivalent to
a single-head one.

The proof requires a lemma proved in a previous manuscript~\cite{libe-20-a}.

\begin{lemma}
\label{set-implies-set}

If $F$ is a definite Horn formula, the following three conditions are
equivalent, where $P' \rightarrow x$ is not a tautology ($x \not \in P'$).

\begin{enumerate}

\item $F \models P' \rightarrow x$;

\item $F^x \cup P' \models P$ where $P \rightarrow x \in F$;

\item $F \cup P' \models P$ where $P \rightarrow x \in F$.

\end{enumerate}

\end{lemma}

The necessity of Condition~\ref{first-condition} is proved by the following
lemma.

\begin{lemma}
\label{first-necessary}

If $F$ is a single-head formula, then for each variable $x$ there exists a set
of variables $A$ such that $x \not\in A$ and such that for all sets $B$ of
variables such that $x \not\in B$, $F \models B \rightarrow x$ implies $F
\models B \rightarrow A$ and $F \models A \rightarrow x$.

\end{lemma}

\proof If $x$ is the head of no clause in $F$ then $F \models B \rightarrow x$
never holds. The condition is vacuously satisfied by $A = \emptyset$.

Otherwise, $x$ is the head of a clause $A \rightarrow x \in F$. By assumption,
clauses are not tautologic: $x \not\in A$. By Lemma~\ref{set-implies-set}, if
$F \models B \rightarrow x$ and $x \not\in B$ then there exists a set of
variables $C$ such that $F \models B \rightarrow C$ and $C \rightarrow x \in
F$. Since $F$ is single-head, $C \rightarrow x$ is $A \rightarrow x$. This
implies $F \models B \rightarrow A$ and $A \rightarrow x \in F$, and the latter
implies $F \models A \rightarrow x$.~\qed

This lemma states that Condition~\ref{first-condition} holds for every
single-head formula. Extending it to every formula that is equivalent to a
single-head one is straightforward. Condition~\ref{first-condition} is
semantical, it only refers to entailments of clauses from the formula. Since it
holds for every single-head formula it also holds for every formula that is
equivalent to it. The same goes for Condition~\ref{second-condition} and the
following lemma.

\begin{lemma}
\label{second-necessary}

If $F$ is a single-head formula, then
{} $F \models A \rightarrow B$
implies
{} $\exists C \in \equiset(B,F) ~.~ C \backslash \equiall(B,F) \subseteq A$.

\end{lemma}

\proof A sketch precedes the formal proof. The entailment
{} $F \models A \rightarrow B$
is assumed. The claim is the existence of a set $C \in \equiset(B,F)$ that is
contained in $\equiall(B,F) \cup A$.

The set $B$ is by definition in $\equiset(B,F)$, but may not be contained in
$\equiall(B,F) \cup A$. Its variables that are not in this set are replaced by
others.

Let $d$ be a variable of $B$ that is neither in $\equiall(B,F)$ nor in $A$.
Since it is not in $\equiall(B,F)$, a set $C$ in $\equiset(B,F)$ does not
contain it. Since $C$ is equivalent to $B$, it entails $d$. Therefore, both $C$
and $A$ entail $d$ while not containing it.

By Lemma~\ref{set-implies-set} and because $F$ is single-head, $F$ contains a
clause $D \rightarrow d$ such that both $A$ and $C$ entail $D$. Let $B'$ be $B$
where $d$ is replaced by $D$. It has the same properties of $B$: it is in
$\equiset(B,F)$ and is entailed by both $A$ and $C$. This replacement is
iterated until the set is contained in $\equiall(B,F) \cup A$.

\

The proof is by induction on the size of $F$. It hinges on the following
Property~(\ref{structural}). This property holds for a subset of $F$ if it
holds for all smaller subsets. Induction on the size of the subset proves it
for $F$ itself.

\begin{eqnarray}
\nonumber
&& F' \models A \rightarrow d			\\
\nonumber
&& F' \subseteq F				\\
\nonumber
&& d \in D \in \equiset(B,F)			\\
& \Rightarrow &
\exists E ~.~
	F \models A \rightarrow E ,~
	F \models E \rightarrow d ,~
	F \models D \rightarrow E ,~
	E \subseteq A \cup \equiall(B,F)
\label{structural}
\end{eqnarray}

The induction step assumes that Property~\ref{structural} holds when $F'$ is
replaced by every subset of $F$ smaller than $F'$; it requires proving
Property~\ref{structural} for $F'$.

If $d$ is in either $A$ or $\equiall(B,F)$, Property~\ref{structural} holds for
$F'$ with $E=\{d\}$. Indeed, $F' \models A \rightarrow d$ and $F' \subseteq F$
imply $F \models A \rightarrow E$; the condition $F \models E \rightarrow d$
holds because $d \in E$; the condition $F \models D \rightarrow E$ holds
because $E = \{d\} \subseteq D$; finally, $E \subseteq A \cup \equiall(B,F)$
holds by assumption.

The other case is that $d$ is neither in $A$ nor in $\equiall(B,F)$. Since $d$
is not in $\equiall(B,F)$, some $D' \in \equiset(B,F)$ does not include it.
Since $D'$ is in $\equiset(B,F)$, it implies all literals of $D$, including
$d$. In formulae, $F \models D' \rightarrow d$. Lemma~\ref{set-implies-set}
applies to both this implication and the assumption $F' \models A \rightarrow
d$. Since $F$ is single-head and $F'$ is a subset of its, the clause in $F$ and
$F'$ is the same: $D'' \rightarrow d \in F'$ with $F^d \models D' \rightarrow
D''$ and $F'^d \models A \rightarrow D''$.

Since $D$ and $D'$ are in $\equiset(B,F)$, both $F \models B \rightarrow D$ and
$F \models B \rightarrow D'$ hold. Because of $F \models D' \rightarrow D''$,
it holds $F \models B \rightarrow D \backslash \{d\} \cup D''$. Since $F
\models D'' \rightarrow d$, it holds $F \models D \backslash \{d\} \cup D''
\rightarrow D$. Since $F \models D \rightarrow B$ because of $D \in
\equiset(B,F)$, it follows $F \models D \backslash \{d\} \cup D'' \rightarrow
B$. The conclusion is $D \backslash \{d\} \cup D'' \in \equiset(B,F)$.

Every variable $d'' \in D''$ is in an element $D \backslash \{d\} \cup D''$ of
$\equiset(B,F)$. Proved above is $F'^d \models A \rightarrow D''$, which
implies $F'^d \models A \rightarrow d''$. Since $F'^d \subset F' \subseteq F$,
this formula $F'^d$ is a subset of $F$ smaller than $F$. By the induction
assumption, every subset $F'^d$ of $F$ smaller than $F'$ and variable $d'' \in
D''$ satisfy Property~\ref{structural}. By induction, since its premise holds
also its conclusion holds: $
{} \exists E'' ~.~
{} 	F \models A \rightarrow E'' ,~
{} 	F \models E'' \rightarrow d'' ,~
{} 	F \models D \rightarrow E'' ,~
{} 	E'' \subseteq A \cup \equiall(B,F)
$.

The claim is that some set $E$ satisfies this conclusion for $d$. This is the
case for the set $E$ that is the union of the sets $E''$ for all $d'' \in D$.
The conditions $F \models A \rightarrow E''$, $F \models D \rightarrow E''$ and
$E'' \subseteq A \cup \equiall(B,F)$ all extends from individual sets to their
union. The only condition that requires some proof is $F \models E \rightarrow
d$; since $F \models E'' \rightarrow d''$ for every $d'' \in D''$ and $E$ is
the union of the sets $E''$, it follows $F \models E \rightarrow D''$. Since
$D'' \rightarrow d \in F$, the conclusion $F \models E \rightarrow d$ follows.

This almost proves Property~\ref{structural} by induction: it is true for $F'
\subseteq F$ if it is true all subsets of $F$ smaller than $F'$. The missing
bit is the base case: $F' = \emptyset$. The premise of
Property~\ref{structural} includes $F' \models A \rightarrow d$, which implies
$d \in A$ because $F'$ is empty. Its conclusion holds for $E=\{d\}$.

\

Induction proves Property~\ref{structural} for every $F' \subseteq F$. In
particular, it proves it for $F'=F$.

\begin{eqnarray*}
&& F \models A \rightarrow d			\\
&& d \in D \in \equiset(B,F)			\\
&& \Rightarrow
\exists E ~.~
	F \models A \rightarrow E ,~
	F \models E \rightarrow d ,~
	F \models D \rightarrow E ,~
	E \subseteq A \cup \equiall(B,F)
\end{eqnarray*}

Let $D$ be a set such that $F \models A \rightarrow D$ and $D \in
\equiset(B,F)$. The premises $F \models A \rightarrow d$ and $d \in D \in
\equiset(B,F)$ of the property hold for all elements $d \in D$. As a result,
for each $d \in D$ the conditions $F \models A \rightarrow E$, $F \models E
\rightarrow d$, $F \models D \rightarrow E$ and $E \subseteq A \cup
\equiall(B,F)$ all holds for some set $E$.

One part of the definition of $D \in \equiset(B,F)$ is $F \models B \rightarrow
D$. Since $F \models D \rightarrow E$, it follows $F \models B \rightarrow D
\backslash \{d\} \cup E$. The other part of the definition of $D \in
\equiset(B,F)$ is $F \models D \rightarrow B$. Since $F \models E \rightarrow
d$, it follows $F \models D \backslash \{d\} \cup E \rightarrow B$. The
conclusion is $D \backslash \{d\} \cup E \in \equiset(B,F)$.

In summary, if $F \models A \rightarrow D$ and $D \in \equiset(B,F)$, replacing
an arbitrary variable $d$ in $D$ but not in $A \cup \equiall(B,F)$ with its set
$E \subseteq A \cup \equiall(B,F)$ results in another equiset $D \backslash
\{d\} \cup E$ entailed by $A$. In other words, $D \backslash \{d\} \cup E$ has
the same properties of $D$: they are both equisets entailed by $A$. At the same
time, $D \backslash \{d\} \cup E$ contains one less variable that is not in $A
\cup \equiall(B,F)$. This replacement can be iterated, decreasing the number of
variables not in $A \cup \equiall(B,F)$ until it reaches zero. Induction on
this number proves that $F \models A \rightarrow D$ and $D \in \equiset(B,F)$
imply $F \models A \rightarrow C$ with $C \in \equiset(B,F)$ and $C \subseteq A
\cup \equiall(B,F)$.

This is the claim of the lemma when $D=B$.~\qed

\subsection{Insufficient conditions}

The two conditions are proved insufficient to ensure single-head equivalence.
The following formula satisfies both, but is not equivalent to any single-head
formula.

\begin{equation}
\label{cyclic-with-ab}
F = \{
	ab \rightarrow x,
	bx \rightarrow c,
	ac \rightarrow d,
	d \rightarrow x
\}
\end{equation}

As expected, this formula hinges around a loop. Yet, it is not a simple loop
where each variable entails another. The heads of the clauses in the loop are
still the individual variables $x, c, d, x$, but the first step of the loop
requires $a$ and the second $b$. This means that both $a$ and $b$ are required
to close the loop.


%
%
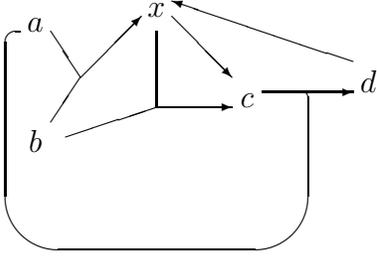
\setlength{\unitlength}{5000sp}%
\begingroup\makeatletter\ifx\SetFigFont\undefined%
\gdef\SetFigFont#1#2#3#4#5{%
  \reset@font\fontsize{#1}{#2pt}%
  \fontfamily{#3}\fontseries{#4}\fontshape{#5}%
  \selectfont}%
\fi\endgroup%
\begin{picture}(1826,1284)(7340,-7225)
\thinlines
{\color[rgb]{0,0,0}\put(7426,-6466){\oval(152,660)[tl]}
\put(8101,-6466){\oval(1502,1502)[bl]}
\put(8101,-6466){\oval(1502,1502)[br]}
\put(8851,-6466){\oval(  2, 60)[tr]}
}%
{\color[rgb]{0,0,0}\put(7576,-6136){\line( 2,-3){150}}
\put(7726,-6361){\line(-2,-3){150}}
}%
{\color[rgb]{0,0,0}\put(7726,-6361){\vector( 1, 1){300}}
}%
{\color[rgb]{0,0,0}\put(8176,-6061){\vector( 1,-1){300}}
}%
{\color[rgb]{0,0,0}\put(9076,-6286){\vector(-3, 1){900}}
}%
{\color[rgb]{0,0,0}\put(7651,-6661){\line( 3, 1){450}}
\put(8101,-6511){\line( 0, 1){375}}
}%
{\color[rgb]{0,0,0}\put(8101,-6511){\vector( 1, 0){375}}
}%
{\color[rgb]{0,0,0}\put(8626,-6436){\vector( 1, 0){450}}
}%
\put(7501,-6736){\makebox(0,0)[b]{\smash{{\SetFigFont{12}{24.0}
{\rmdefault}{\mddefault}{\updefault}{\color[rgb]{0,0,0}$b$}%
}}}}
\put(8101,-6061){\makebox(0,0)[b]{\smash{{\SetFigFont{12}{24.0}
{\rmdefault}{\mddefault}{\updefault}{\color[rgb]{0,0,0}$x$}%
}}}}
\put(7501,-6136){\makebox(0,0)[b]{\smash{{\SetFigFont{12}{24.0}
{\rmdefault}{\mddefault}{\updefault}{\color[rgb]{0,0,0}$a$}%
}}}}
\put(8551,-6511){\makebox(0,0)[b]{\smash{{\SetFigFont{12}{24.0}
{\rmdefault}{\mddefault}{\updefault}{\color[rgb]{0,0,0}$c$}%
}}}}
\put(9151,-6436){\makebox(0,0)[b]{\smash{{\SetFigFont{12}{24.0}
{\rmdefault}{\mddefault}{\updefault}{\color[rgb]{0,0,0}$d$}%
}}}}
\end{picture}%
\nop{
+-- a
|   +--> x <---------------+
|   b     |                |
|    -----+---> c ---+---> d
|                    |
+--------------------+
}

The formula is in the {\tt conditiontwo.py} test file of the {\tt
singlehead.py} program. The following lemma proves that it satisfies the first
condition.

\begin{lemma}
\label{ab-first}

Formula~\ref{cyclic-with-ab} satisfies Condition~\ref{first-condition}.

\end{lemma}

\proof All variables but $x$ are already the head of a single clause;
therefore, Condition~\ref{first-condition} holds with $A$ equal to the body of
that clause thanks to Lemma~\ref{set-implies-set}.

It also holds for $x$ with $A=\{d\}$. Condition~\ref{first-condition} is that
every set of variables entailing $x$ also entails $\{d\}$ and that $\{d\}$
entails $x$ if such a set exists. The latter holds because $d \rightarrow x$ is
in $F$. The sets entailing $x$ are $\{a,b\}$, $\{a,c\}$, $\{d\}$ and their
supersets. All three entail $d$ according to $F$. Therefore, their supersets do
as well.~\qed

Condition~\ref{first-condition} suggests that $ab \rightarrow x$ is entailed
only thanks to $ab \rightarrow d$, but this clause only holds thanks to the
chain of implications $ab \rightarrow x$, $bx \rightarrow c$ and $ac
\rightarrow d$, which requires $ab \rightarrow x$ as its first step. The plan
was to only retain $d \rightarrow x$ and to obtain $ab \rightarrow x$ as a
consequence of $ab \rightarrow d$ and $d \rightarrow x$, but the first premise
requires $ab \rightarrow x$, a second clause with $x$ as the head.

Proving that $F$ satisfies Condition~\ref{second-condition} is more complicated
because it requires considering sets of variables rather than single variables.
A concept introduced below, that of inequivalence, helps in that. This is why
the formal proof is delayed after some other results, in
Lemma~\ref{inequivalent-second-condition}.

While $F$ satisfies both Condition~\ref{first-condition}
and~\ref{second-condition}, it is not single-head equivalent. This is proved by
Lemma~\ref{ab-nonsingle}, which is later in the article because it requires
some other results.

\subsection{Equivalence}

Another necessary condition for single-head equivalence relates the premises of
the clauses that entail a variable and the same premises in the single-head
equivalent formula.

\begin{lemma}
\label{equivalent-preconditions}

If $F$ is equivalent to a single-head formula $F'$ that contains the clause $P
\rightarrow x$, then $F$ contains $P' \rightarrow x$ with $F \models P \equiv
P'$.

\end{lemma}

\proof Containment $P \rightarrow x \in F'$ implies entailment $F' \models P
\rightarrow x$. By equivalence, also $F$ entails $P \rightarrow x$.
Lemma~\ref{set-implies-set} implies the existence of a set of variables $P'$
such that $x \not\in P'$, $F \models P \rightarrow P'$ and $P' \rightarrow x
\in F$. The latter condition implies $F \models P' \rightarrow x$. By
equivalence, $F' \models P' \rightarrow x$. Again, Lemma~\ref{set-implies-set}
implies that $F' \models P' \rightarrow P''$ for some $P'' \rightarrow x \in
F'$. Since $F'$ is single-head and contains $P \rightarrow x$, this is only
possible if $P'' = P$. As a result, $F' \models P' \rightarrow P''$ is the same
as $F' \models P' \rightarrow P$. Since $F \models P \rightarrow P'$ and the
two formulae are equivalent, $F \models P \equiv P'$ is proved.~\qed

\subsection{Redundancy}

Sometimes, a formula can be made single-head just by removing some redundant
clauses. For example, $\{a \rightarrow b, b \rightarrow c, a \rightarrow c\}$
is not single-head, but removing $a \rightarrow c$ makes it so while preserving
its semantics.

Is this always the case?

If so, a formula could be made single-head by non-deterministically removing
redundant clauses until the formula either becomes irredundant or single-head.

The following theorem proves that this is not always possible. Sometimes the
single-head equivalent formula is a subset of the given formula, sometimes it
is not.

\begin{theorem}
\label{no-subset}

Some formulae are equivalent to single-head formulae, but no equivalent subset
of them is single-head.

\end{theorem}

\proof Such a formula is
{} $\{a \rightarrow b, b \rightarrow a, b \rightarrow c, c \rightarrow b\}$,
in the {\tt twoequiv.py} test file of the {\tt singlehead.py} program. Its
three variables are equivalent. The same can be achieved by a cycle of clauses,
like in the single-head formula
{} $\{a \rightarrow b, b \rightarrow c, c \rightarrow a\}$.

All clauses of the formula are irredundant. As a result, the formula is
equivalent to none of its proper subsets, only to itself. Since it is not
single-head, the claim is proved.~\qed

Turning a formula into its single-head form cannot always be done just by
removing clauses. Sometimes it requires adding new ones as well. The proof of
the theorem shows such a case: the formula does not contain $c \rightarrow a$,
but its equivalent single-head formula does.

If a formula does not contain cycles of clauses, removing redundant clauses
always leads to a single-head equivalent formula if any. This is proved in the
next section, where cyclicity is formally defined.

\section{Cyclicity}
\label{cyclic}

All examples that show that single-head equivalence is complicated contain
cycles of variables: a variable implies another which implies yet another which
implies the first, for example. When a formula is acyclic,
Condition~\ref{first-condition} is not only necessary to single-head
equivalence but also sufficient.

\subsection{Syntactical and semantical cyclicity}

Cycles are made of variables implying others, like $a \rightarrow b$. Other
preconditions do not matter, so $acd \rightarrow b$ is the same as $a
\rightarrow b$ from this point of view. More precisely, it is the same as $a
\rightarrow b$, $c \rightarrow b$ and $d \rightarrow b$. This set can be taken
as denoting a directed graph, where an edge from $a$ to $b$ is denoted $a
\rightarrow b$.



A formula defines two graphs:

\begin{description}

\item[syntactic graph:]
{} $Y(F) =
{}  \{y \rightarrow x \mid \exists P ~.~ P \cup \{y\} \rightarrow x \in F\}$;
an edge from $y$ to $x$ comes from a clause with $y$ in the body and $x$ in the
head; the syntactic graph is the graph associated to the Horn formula according
to Hammer and Kogan~\cite{hamm-koga-95}.

\item[semantic graph:]
{} $E(F) =
{}  \{y \rightarrow x \mid
{}      \exists P ~.~ F \models P \cup \{y\} \rightarrow x \mbox{ and }
{}                    F \not\models P \rightarrow x\}$;
the edge from $y$ to $x$ means that $y$ participates in entailing $x$;
participates means that it entails $x$ with something else and is necessary to
that.

\end{description}

The first definition is purely syntactic: it only involves the clauses of the
formula and what they contain; it does not even mention models or implications.
The second is purely semantical: it only involves entailments, and does not
mention any clause of the formula or its variables; it could as well be defined
from a set of models instead of a formula.

The difference is evident on redundant clauses. The first definition makes an
edge from $a$ to $b$ out of a clause $acd \rightarrow b$ even when another
clause $cd \rightarrow b$ is present. The second does not because $a$ is
redundant in implying $b$. Yet, they coincide when infusing some semantics in
the first. The semantical graph of a formula is identical to the syntactical
graph of the set of its prime implicates.

The job of joining the gap between the two definitions is done by primality,
which removes redundant clauses like $acd \rightarrow b$ and consequently bars
the edge from $a$ to $b$. This is proved by the following theorem, where
$PI(F)$ is the set of prime implicates of a formula $F$.

\begin{lemma}
\label{prime}

If $F$ is a formula that coincides with the set of its prime implicates
$PI(F)$, then $Y(F)$ coincides with $E(F)$.

\end{lemma}

\proof The first part of the claim is that $y \rightarrow x \in Y(F)$ implies
$y \rightarrow x \in E(F)$. By definition, $y \rightarrow x \in Y(F)$ means $P
\cup \{y\} \rightarrow x \in F$ for some set of variables $P$. As a result, $F
\models P \cup \{y\} \rightarrow x$. This is the first point in the definition
of $y \rightarrow x \in E(F)$. The second is $F \not\models P \rightarrow x$.
Its contrary $F \models P \rightarrow x$ means that $F$ implies a subset of $P
\cup \{y\} \rightarrow x$, which is therefore not a prime implicate. This
contradicts the assumption that $F$ only contains its prime implicates.

The second part of the claim is that $y \rightarrow x \in E(F)$ implies $y
\rightarrow x \in Y(F)$. By definition, $y \rightarrow x \in E(F)$ means that
for some set of variables $P$ the clause $P \cup \{y\} \rightarrow x$ is
entailed by $F$ while $P \rightarrow x$ is not. The first fact implies that $F$
contains a prime implicate contained in $P \cup \{y\} \rightarrow x$. Such a
prime implicate contains $x$ because otherwise $F$ would not be a definite Horn
formula. It also contains $y$, because otherwise $F$ would imply $P \rightarrow
x$. Therefore, this prime implicate has the form $P' \cup \{y\} \rightarrow x$.
Since $F$ contains all its prime implicates, it also contains $P' \cup \{y\}
\rightarrow x$. This means that $y \rightarrow x \in Y(F)$.~\qed

The goal is to show some sort of equality between syntactic and semantic
cyclicity. This lemma proves part of that: if the semantic graph of $F$ does
not contain cycles, neither the syntactic graph of $PI(F)$ does. Semantic
acyclic implies acyclicity of one syntactic form.

Only one, not all of them. For example,
{} $\{a \rightarrow b, b \rightarrow c\}$
is semantically acyclic, yet its equivalent formula
{} $\{a \rightarrow b, b \rightarrow c, ac \rightarrow b\}$
is syntactically cyclic. Most formulae can be made cyclic by adding some
redundant clauses like $ac \rightarrow b$. The resulting formula is equivalent
but cyclic.

When looking for a syntactic counterpart of semantic cyclicity, better looking
for something else. The exact opposite makes sense: if a formula is
semantically cyclic, all equivalent formulae are syntactically cyclic, and the
other way around.

This is actually the case for arbitrary paths, not just cycles. If the semantic
graph of a formula contains a path from $x$ to $y$, the same happens for the
syntactic graph of every equivalent formula, and vice versa. Paths are
formalized by the transitive closure of graphs.

\begin{definition}

The transitive closure $G^*$ of a graph $G$ is the graph that contains an edge
$y \rightarrow x$ if and only if $G$ has a path from $y$ to $x$.

\end{definition}

The transitive closure $Y^*(F)$ of the syntactic graph is very close to the
semantic graph: $E(F) \subseteq Y^*(F)$. The transitive closure of the
syntactic graph contains the semantic graph. A very simple intuitive argument
tells why: the syntactic graph is based on clauses; a path in it is a sequence
of clauses; a sequence of clauses creates an implication, which is the root of
the semantic graph.

The formal proof of this containment is based on the syntactic graph being
monotone with respect to the formula, which the semantic graph is not by
Lemma~\ref{semantic-monotone}.

\begin{lemma}
\label{syntactic-monotone}

If $F \subseteq F'$ then $Y(F) \subseteq Y(F')$.

\end{lemma}

\proof The definition of $y \rightarrow x \in Y(F)$ is that $F$ contains a
clause $P \rightarrow x$ such that $y \in P$. Since $F \subseteq F'$, this
clause $P \rightarrow x$ is also in $F'$. The definition of $y \rightarrow x
\in Y(F')$ is met because $y \in P$. \qed

Monotonicity allows proving $E(F) \subseteq Y^*(F)$.

\begin{lemma}
\label{syntactic-semantic}

If $y \rightarrow x \in E(F)$, then $Y(F)$ contains a path from $y$ to $x$.

\end{lemma}

\proof If $y = x$ the claim holds because a zero-length path between every node
and itself always exists in every graph.

A consequence is that the claim holds when the formula only comprises one
variable, because $x$ and $y$ coincide in this case. This is the base case of
an induction over the number of variables of the formula.

The inductive case assumes $y \rightarrow x \in E(F)$ and that the claim holds
for every formula smaller than $F$; the conclusion to prove is $y \rightarrow x
\in Y^*(F)$.

The assumption $y \rightarrow x \in E(F)$ is defined as
{} $F \models P \cup \{y\} \rightarrow x$ and
{} $F \not\models P \rightarrow x$
for some set of variables $P$.

The claim is already proved when $x = y$. The rest of the proof is for the case
$x \not= y$. Since
{} $F \not\models P \rightarrow x$,
the variable $x$ is not in $P$. It is not $y$ either; therefore, $x \not\in P
\cup \{y\}$.

Since
{} $F \models P \cup \{y\} \rightarrow x$
and $x \not\in P \cup \{y\}$, Lemma~\ref{set-implies-set} tells that $F$
contains a clause $P' \rightarrow x$ such that $F^x \models P \cup \{y\}
\rightarrow P'$.

If $y \in P'$ the claim is proved because $P' \rightarrow x \in F$ and $y \in
P'$ define $y \rightarrow x \in Y(F)$, which implies $y \rightarrow x \in
Y*(F)$.

The other case is $y \not\in P'$. Since $P' \rightarrow x \in F$, if $F^x
\models P \rightarrow P'$ then $F \models P \rightarrow x$, which is false. As
a result, $F^x \not\models P \rightarrow P'$: for at least some variable $p \in
P'$, it holds $F^x \not\models P \rightarrow p$. The entailment $F^x \models P
\cup \{y\} \rightarrow p$ instead holds because $F^x \models P \cup \{y\}
\rightarrow P'$ and $p \in P'$. This proves that $y \rightarrow p \in E(F^x)$.

Summarizing: $y \rightarrow x \in E(F)$ implies $y \rightarrow p \in E(F^x)$
and $P' \rightarrow x \in F$ with $p \in P'$. The latter implies $p \rightarrow
x \in Y(F)$. The former implies the existence of a path from $y$ to $p$ in
$Y(F^x)$ by induction, since $F^x$ has one variable less than $F$. Since $F^x
\subset F$, the same path is in $Y(F)$ by Lemma~\ref{syntactic-monotone}.

Since $Y(F)$ contains a path from $y$ to $p$ and the edge $p \rightarrow x$, it
also contains the path from $y$ to $x$.~\qed

The following results are about cycles. Assuming that formulae do not contain
tautologies, the syntactic graphs never contain loops, i.e., cycles comprising
only one node. The semantical graph contain most of them instead, since $F
\models x \rightarrow x$ always holds and $F \models \emptyset \rightarrow x$
holds only if $F \models x$. Cycles in the syntactical and semantical graph
match only when loops are excluded.

From this point on, only cycles containing at least two nodes are considered.

\begin{lemma}
\label{cycle-semantic-syntactic}

If $E(F)$ contains a nontrivial cycle, so does $Y(F)$.

\end{lemma}

\proof Nontrivial cycles contain at least two variables. Such a cycle in $E(F)$
comprises a set of all-different variables $x_1,\ldots,x_m$ such that $m \geq
2$ and $E(F)$ contains $x_i \rightarrow x_{i+1}$ for every $i = 1,\ldots,m-1$
and $x_m \rightarrow x_1$. By Lemma~\ref{syntactic-semantic}, the syntactic
graph $Y(F)$ contains a path from each $x_i$ to $x_{i+1}$ and from $x_m$ to
$x_1$. This is a path of at least $2$ variables starting from $x_1$ and ending
in $x_1$, a nontrivial cycle.~\qed

Cyclicity in the semantic graph imposes cyclicity in the syntactic graph. Not
the other way around, as shown above. What imposes cyclicity in the semantic
graph is cyclicity in the syntactic graph of all equivalent formulae.

\begin{theorem}
\label{cycle-syntax-semantic}

For every formula $F$, the graph $E(F)$ contains nontrivial cycles if and only
if $Y(F')$ contains cycles for every $F' \equiv F$.

\end{theorem}

\proof The first part of the proof shows that if $E(F)$ contains a nontrivial
cycle and $F'$ is equivalent to $F$, then $Y(F')$ contains a nontrivial cycle.
Since the semantic graph is defined semantically, it is the same for equivalent
formulae. In the present case, $E(F') = E(F)$. Since $E(F)$ contains a
nontrivial cycle, do does $E(F')$. And so does $Y(F')$ thanks to
Lemma~\ref{cycle-semantic-syntactic}.

The second part of the proof shows the converse: if $E(F)$ does not contain any
nontrivial cycle, then $Y(F')$ does not either for some formula $F'$ equivalent
to $F$. Such a formula $F'$ is the set of the prime implicates $PI(F)$ of $F$.
Since $PI(F)$ is equivalent to $F$, it has the same semantic graph: $E(F) =
E(PI(F))$. Since $PI(F)$ coincides with its set of prime implicates, its
semantic graph $E(PI(F))$ coincides with its syntactic graph $Y(PI(F))$ by
Lemma~\ref{prime}. Therefore, $E(F) = Y(PI(F))$. Since $E(F)$ does not contain
nontrivial cycles, $Y(PI(F))$ does not either.~\qed

Hammer and Kogan~\cite{hamm-koga-95} introduced the syntactic graph of a
formula, with its consequent definition of cyclicity. They also defined its
semantical counterpart: a Boolean Horn function is acyclic if a Horn formula
realizing it is acyclic. Since Boolean functions are realized by equivalent
formulae, this is the same as defining acyclicity as equivalence with a
syntactically acyclic formula. Lemma~\ref{syntactic-semantic} proves that this
definition is the same as semantical acyclicity as defined in this article, in
terms of implications from minimal premises ($F \models P \cup \{y\}
\rightarrow x$ and $F \not\models P \rightarrow x$).

The similarity is further emphasized by how close Lemma~\ref{prime} and
Lemma~\ref{syntactic-semantic} relate to Lemma~4.1 by Hammer and
Kogan~\cite{hamm-koga-95}: the syntactic graph of a formula has a path between
each variable of the body of a prime implicate and its head. They also relate
to Theorem~4.3 by Boros et~al.~\cite{boro-etal-98}: if two formulae are prime
and equivalent, their reachability between variables is the same.

As a related note, a semantically acyclic formula may contain a syntactically
cyclic subset. An example is
{} $F =
{}  \{a \rightarrow b, a \rightarrow c, ac \rightarrow b, ab \rightarrow c\}$.
This formula is semantically acyclic, as it is equivalent to its subset $\{a
\rightarrow b, a \rightarrow c\}$. Removing these two clauses makes the other,
previously redundant two matter: $ab \rightarrow c$ and $ac \rightarrow b$.
This is a cycle, from $b$ to $c$ and back.

\begin{tabular}{ccc}
\setlength{\unitlength}{5000sp}%
\begingroup\makeatletter\ifx\SetFigFont\undefined%
\gdef\SetFigFont#1#2#3#4#5{%
  \reset@font\fontsize{#1}{#2pt}%
  \fontfamily{#3}\fontseries{#4}\fontshape{#5}%
  \selectfont}%
\fi\endgroup%
\begin{picture}(2004,2007)(10923,-7273)
\thinlines
{\color[rgb]{0,0,0}\put(11807,-5536){\oval(388, 44)[tr]}
\put(11807,-6388){\oval(1748,1748)[tl]}
\put(11776,-6388){\oval(1686,1746)[bl]}
}%
{\color[rgb]{0,0,0}\put(11551,-6511){\circle{336}}
}%
{\color[rgb]{0,0,0}\put(12751,-6511){\circle{336}}
}%
{\color[rgb]{0,0,0}\put(12151,-5611){\circle{336}}
}%
{\color[rgb]{0,0,0}\put(12151,-5761){\line( 0,-1){750}}
\put(12151,-6511){\vector(-1, 0){450}}
}%
{\color[rgb]{0,0,0}\put(12601,-6511){\line(-1, 0){450}}
}%
{\color[rgb]{0,0,0}\put(11626,-6661){\line( 1,-4){150}}
\put(11776,-7261){\vector( 3, 2){900}}
}%
\put(12151,-5386){\makebox(0,0)[b]{\smash{{\SetFigFont{12}{24.0}
{\rmdefault}{\mddefault}{\updefault}{\color[rgb]{0,0,0}$a$}%
}}}}
\put(11551,-6886){\makebox(0,0)[b]{\smash{{\SetFigFont{12}{24.0}
{\rmdefault}{\mddefault}{\updefault}{\color[rgb]{0,0,0}$b$}%
}}}}
\put(12751,-6886){\makebox(0,0)[b]{\smash{{\SetFigFont{12}{24.0}
{\rmdefault}{\mddefault}{\updefault}{\color[rgb]{0,0,0}$c$}%
}}}}
{\color[rgb]{0,0,0}\put(12751,-6361){\vector( 0,-1){0}}
\put(12301,-6361){\oval(900,1500)[tr]}
}%
{\color[rgb]{0,0,0}\put(12001,-6361){\oval(900,1500)[tl]}
\put(11551,-6361){\vector( 0,-1){0}}
}%
\end{picture}%
\nop{
+-------- a --------+
|     +--   --+     |
|     |       |     |
| +---|-------+---+ |
V |   |           V V
 b    |            c
  ^   |           |
  +---+-----------+
}
&
\hbox to 2cm{\hfill}
&
\setlength{\unitlength}{5000sp}%
\begingroup\makeatletter\ifx\SetFigFont\undefined%
\gdef\SetFigFont#1#2#3#4#5{%
  \reset@font\fontsize{#1}{#2pt}%
  \fontfamily{#3}\fontseries{#4}\fontshape{#5}%
  \selectfont}%
\fi\endgroup%
\begin{picture}(2004,2007)(10923,-7273)
\thinlines
{\color[rgb]{0,0,0}\put(11807,-5536){\oval(388, 44)[tr]}
\put(11807,-6388){\oval(1748,1748)[tl]}
\put(11776,-6388){\oval(1686,1746)[bl]}
}%
{\color[rgb]{0,0,0}\put(11551,-6511){\circle{336}}
}%
{\color[rgb]{0,0,0}\put(12751,-6511){\circle{336}}
}%
{\color[rgb]{0,0,0}\put(12151,-5611){\circle{336}}
}%
{\color[rgb]{0,0,0}\put(12151,-5761){\line( 0,-1){750}}
\put(12151,-6511){\vector(-1, 0){450}}
}%
{\color[rgb]{0,0,0}\put(12601,-6511){\line(-1, 0){450}}
}%
{\color[rgb]{0,0,0}\put(11626,-6661){\line( 1,-4){150}}
\put(11776,-7261){\vector( 3, 2){900}}
}%
\put(12151,-5386){\makebox(0,0)[b]{\smash{{\SetFigFont{12}{24.0}
{\rmdefault}{\mddefault}{\updefault}{\color[rgb]{0,0,0}$a$}%
}}}}
\put(11551,-6886){\makebox(0,0)[b]{\smash{{\SetFigFont{12}{24.0}
{\rmdefault}{\mddefault}{\updefault}{\color[rgb]{0,0,0}$b$}%
}}}}
\put(12751,-6886){\makebox(0,0)[b]{\smash{{\SetFigFont{12}{24.0}
{\rmdefault}{\mddefault}{\updefault}{\color[rgb]{0,0,0}$c$}%
}}}}
\end{picture}%
\nop{
          a
      +--   --+ 
      |       |
  +---|-------+---+
V |   |           V
 b    |            c
  ^   |           |
  +---+-----------+
}
\\
a semantically acyclic formula
&
&
a syntactically cyclic subset
\end{tabular}

\subsection{Single-head equivalence of acyclic formulae}

All attempts at finding a necessary and sufficient condition to single-head
equivalence failed, but they all failed because of cyclic counterexamples ---
formulae that satisfy a candidate condition but were not equivalent to any
single-head formulae. This is why cyclicity is in this article: because it
looks like the root of all evil. If so, disallowing it should give a simple
necessary and sufficient condition to single-head equivalence. The following
lemma proves that to be the case.

\begin{lemma}
\label{first-acyclic-sufficient}

Every semantically acyclic formula that satisfies
Condition~\ref{first-condition} is equivalent to a single-head formula.

\end{lemma}

\proof Let $F'$ be a formula such that $E(F')$ is nontrivially acyclic and
satisfies Condition~\ref{first-condition}. By
Theorem~\ref{cycle-syntax-semantic}, there exists a formula $F$ such that $F
\equiv F'$ and $Y(F)$ is acyclic.

The claim is proved by showing that all clauses of $F$ with head $x$ are
redundant but one at most.

If $F$ does not contain any non-tautological clause of head $x$, the claim is
trivially true. Otherwise, let $B \rightarrow x$ a non-tautological clause of
$F$. Since $F'$ satisfies Condition~\ref{first-condition}, a set of variables
$A$ satisfies $x \not\in A$, $F' \models B \rightarrow A$ and $F' \models A
\rightarrow x$. Since $F \equiv F'$, the entailments imply $F \models B
\rightarrow A$ and $F \models A \rightarrow x$.

Lemma~\ref{set-implies-set} applies because of $x \not\in A$ and $F \models A
\rightarrow x$, and implies that $F$ contains a clause $A' \rightarrow x$ such
that $F \models A \rightarrow A'$. This entailment and $F \models B \rightarrow
A$ imply $F \models B \rightarrow A'$ by transitivity.

If $B \rightarrow x$ is different from $A' \rightarrow x$, it is proved
redundant:
{} $F \backslash \{B \rightarrow x\} \models B \rightarrow x$.

Since $A' \rightarrow x$ is in $F$ but is not the same as $B \rightarrow x$, it
is in $F \backslash \{B \rightarrow x\}$. Therefore, $A' \rightarrow x$ is
entailed by $F \backslash \{B \rightarrow x\}$. Let $A''$ be a minimal subset
of $A'$ such that $F \backslash \{B \rightarrow x\} \models A'' \rightarrow x$.

Because of minimality, $F \backslash \{B \rightarrow x\} \not\models (A''
\backslash \{a\}) \rightarrow x$ holds for every $a \in A''$. At the same time,
$F \backslash \{B \rightarrow x\}$ entails $(A'' \backslash \{a\} \cup a)
\rightarrow x$ because this clause is the same as $A'' \rightarrow x$. These
two conditions define $a \rightarrow x \in E(F \backslash \{B \rightarrow
x\})$ for every $a \in A''$.

The claim
{} $F \backslash \{B \rightarrow x\} \models B \rightarrow x$
is proved by contradiction. Its contrary is
{} $F \backslash \{B \rightarrow x\} \not\models B \rightarrow x$.
It implies
{} $F \backslash \{B \rightarrow x\} \not\models B \rightarrow A''$,
as otherwise $B$ would imply $x$ via $A''$ since
{} $F \backslash \{B \rightarrow x\} \models A'' \rightarrow x$,
As a result, there exists $a \in A'' \backslash B$ such that
{} $F \backslash \{B \rightarrow x\} \not\models B \rightarrow a$.
This is one part of the condition $x \rightarrow a \in E(F \backslash \{B
\rightarrow x\})$, which would prove the formula cyclic. The other part is
{} $F \backslash \{B \rightarrow x\} \models (B \cup \{x\} )\rightarrow a$,
which is now proved. Since $x \models B \rightarrow x$ holds in propositional
logic, it holds
{} $F \backslash \{B \rightarrow x\} \cup \{x\} \models B \rightarrow x$,
which means that
{} $F \backslash \{B \rightarrow x\} \cup \{x\}$ implies the only clause of $F$
it does not contain, and consequently it implies $F$. Since $F$ implies $B
\rightarrow A'$, it also implies $B \rightarrow a$ because $a$ is in $A''$,
which is a subset of $A'$. The condition
{} $F \backslash \{B \rightarrow x\} \cup \{x\} \models B \rightarrow a$
can be rewritten as
{} $F \backslash \{B \rightarrow x\} \models (B \cup \{x\}) \rightarrow a$.
This is the second part of
{} $x \rightarrow a \in E(F \backslash \{B \rightarrow x\})$.

The above two paragraphs prove
{} $a \rightarrow x \in E(F \backslash \{B \rightarrow x\})$
for every $a \in A''$ and
{} $x \rightarrow a \in E(F \backslash \{B \rightarrow x\})$
some $a \in A''$. Since $x \not\in A$ and $a \in A''$ with $A'' \subseteq A$,
it holds $x \not= a$. Therefore, $E(F \backslash \{B \rightarrow x\})$ contains
a nontrivial cycle. By Lemma~\ref{cycle-semantic-syntactic}, $Y(F \backslash
\{B \rightarrow x\})$ contains a nontrivial cycle. Since syntactic cyclicity is
monotonic by Lemma~\ref{syntactic-monotone}, also $Y(F)$ is cyclic. This is
contrary to the acyclicity of $Y(F)$. The assumption that let to this
contradiction is $F \backslash \{B \rightarrow x\} \not\models B \rightarrow
x$, which is therefore false.

The conclusion $F \backslash \{B \rightarrow x\} \models B \rightarrow x$
implies that $F \backslash \{B \rightarrow x\}$ is equivalent to $F$. This
proves that a clause $B \rightarrow x$ with $B \not= A''$ is redundant in $F$.
Since $F \backslash \{B \rightarrow x\}$ is equivalent to $F$, it has the same
properties. Therefore, every other clause $C \rightarrow x \in F \backslash \{B
\rightarrow x\}$ with $C \not= A''$ is redundant in it. This procedure can be
iterated and shows that all clauses with head $x$ but one can be removed one at
time.~\qed

The converse is always the case as shown by Lemma~\ref{first-necessary}. The
conclusion is that Condition~\ref{first-condition} is the same as single-head
equivalence on semantically acyclic formulae.

\begin{corollary}

A semantically acyclic formula is single-head equivalent if and only if it
satisfies Condition~\ref{first-condition}.

\end{corollary}

\subsection{Cyclicity and equivalence}

Simple cycles of clauses like $a \rightarrow b$, $b \rightarrow c$ and $c
\rightarrow a$ are easy to detect semantically: some variable (like $a$) is
equivalent to another (for example, $c$) thanks to the given clauses. In
general, formula $F$ is cyclic if $F \models x \equiv y$ for some variables $x$
and $y$. Unfortunately, this is only the case for cycles of binary clauses.

A cycle may also contain clauses with more than one variable in their body. For
example,
{} $F = \{ab \rightarrow c, ab \rightarrow d,
{}        cd \rightarrow a, cd \rightarrow b\}$
require extending the condition from single variables to sets:
{} $F \models \{a,b\} \equiv \{c,d\}$.
In general, the condition is $F \models A \equiv B$, where $A$ and $B$ are
different set of variables. Unfortunately, this condition is also satisfied by
$F = \{a \rightarrow b\}$, which does not contain cycles, with $A=\{a\}$ and
$B=\{a,b\}$.

The two sets of variables being different is not enough. If $A$ is contained in
$B$ then $F \models A \equiv B$ is the same as $F \models A \rightarrow B$:
this is a one-way implication, not a cycle. The other direction $F \models B
\rightarrow A$ is a consequence of monotonicity ($B \models A$), not of the
clauses of $F$.

A further refinement is to define the converse of cyclicity as ``if $F \models
A \equiv B$'' then either $A \subseteq B$ or $B \subseteq A$. This way, if $A$
and $B$ are made equivalent by $F$, then either one contains the other. The
intent is to allow equivalences only if one direction is due to monotonicity of
entailment. A counterexample fails this attempt: $F = \{a \rightarrow b, a
\rightarrow c\}$ makes $\{a,b\}$ equivalent to $\{a,c\}$ in spite of the
absence of cycles. A variable implies other two, making two sets equivalent in
spite of none being contained in the other. While $\{a\}$ is contained in both
$\{a,b\}$ and $\{a,c\}$, transitivity makes $\{a,b\}$ equivalent to $\{a,c\}$
but breaks containment. In general, two sets may be equivalent to each other
while not being contained one in the other; however, they must both be
equivalent to a common subset of them for the formula to be acyclic.

\begin{condition}
\label{condition-inequivalent}

A formula $F$ is {\em inequivalent} if $F \models A \equiv B$ for two sets of
variables $A$ and $B$ implies $F \models A \equiv A \cap B$.

\end{condition}

The word ``inequivalence'' may hint ``not equivalent to something'', maybe also
``not equivalent to a single head-formula''. This is a wrong hint. It is not
the formula that is not equivalent. An inequivalent formula makes no two sets
of variables equivalent. The case $F \models A \equiv (A \cap B)$ is a
necessary exception, as it implies $F \models A \equiv B$ regardless of $F$.

Every acyclic formula is inequivalent, but not the other way around.
Inequivalence fails at expressing acyclicity; yet, a following section shows
an efficient algorithm for single-head equivalence of inequivalent formulae.
Inequivalence is broader than acyclicity (includes more formulae) but still
maintains polynomiality of single-head equivalence. This is why inequivalence
is further analyzed in spite of not expressing acyclicity.

\begin{lemma}
\label{acyclic-inequivalent}

Every semantically acyclic formula is inequivalent, that is,
satisfies Condition~\ref{condition-inequivalent}.

\end{lemma}

\proof If a formula is not inequivalent, two equivalent sets $A$ and $B$ are
not equivalent to their intersection. If one of them is contained in the other,
it is equal to their intersection, violating the assumption. As a result, each
contains some elements that are not in the other. These elements are all
entailed by the other set.

This is almost a semantic cycle, but not quite. That requires not only a mutual
entailment, but also that the preconditions are necessary. This is what
complicates the proof: while some elements of $A$ participate in entailing some
elements of $B$ and vice versa, they may not be necessary to the entailments.
It can however be proved that some are.

\

The claim is proved by contraposition: if a formula is not inequivalent, its
semantic graph contains a nontrivial cycle.

The assumption is that $F \models A \equiv B$ holds, but $F \models A \equiv A
\cap B$ does not for some sets of variables $A$ and $B$.

If $F \models B \equiv A \cap B$ then $F \models A \equiv A \cap B$ because of
$F \models A \equiv B$. This is contrary to assumption. Therefore, $F
\not\models B \equiv A \cap B$.

The set $A$ may be redundant: $F \models A \equiv A \backslash \{a\}$ may be
the case for some $a \in A$. If so, $A \backslash \{a\}$ has the same
semantical properties of $A$ because of equivalence:
{} $F \models A \backslash \{a\} \equiv B$
and
{} $F \not\models A \backslash \{a\} \equiv A \cap B$.
The latter implies
{} $F \not\models A \backslash \{a\} \equiv (A \backslash \{a\}) \cap B$.
This is proved by contradiction: its converse
{} $F \models A \backslash \{a\} \equiv (A \backslash \{a\}) \cap B$
includes
{} $F \models (A \backslash \{a\}) \cap B \rightarrow A \backslash \{a\}$,
which implies
{} $F \models (A \backslash \{a\}) \cap B \rightarrow A$
because $A \backslash \{a\}$ is equivalent to $A$. By monotonicity of
entailment,
{} $F \models A \cap B \rightarrow A$.
The other direction
{} $F \models A \rightarrow A \cap B$
of this entailment holds because $A \cap B \subseteq A$. The result
{} $F \models A \cap B \equiv A$ contradicts the assumption, proving
{} $F \not\models A \backslash \{a\} \equiv (A \backslash \{a\}) \cap B$.
The conclusion is that if $A$ is equivalent to $A \backslash \{a\}$, the
assumptions of the claim also hold for $A \backslash \{a\}$. Iteratively, all
redundant elements of $A$ can be removed without affecting the assumptions. By
symmetry, the same holds for $B$. If the assumptions of the lemma hold for $A$
and $B$, they also hold for some irredundant subsets of them. Only
irredundant formulae $A$ and $B$ are considered from now on.

If $A \subseteq B$ then $A \cap B = A$, which contradicts $F \not\models A
\equiv A \cap B$. As a result, $A$ contains some elements that are not in $B$.
By symmetry, the same holds for $B$. The sets of these elements are denoted $A'
= A \backslash B$ and $B' = B \backslash B$.

Let $a \in A'$. The assumption $F \models A \equiv B$ implies $F \models A
\rightarrow B$. This entailment can be rewritten as
{} $F \models (A \backslash \{a\} \cup \{a\}) \rightarrow B$.
This is the first part of the definition of an edge from $a$ to every element
$b \in B$ in the semantic graph of $F$. The second part is
{} $F \not\models (A \backslash \{a\}) \rightarrow b$.
It may not be the case for all $b \in B$, but it is for some. Otherwise, $F
\models (A \backslash \{a\}) \rightarrow b$ for all $b \in B$ would imply $F
\models (A \backslash \{a\}) \rightarrow B$. Since $F \models B \equiv A$, this
implies $F \models (A \backslash \{a\}) \rightarrow A$, while $A$ is assumed
irredundant. As a result,
{} $F \not\models (A \backslash \{a\}) \rightarrow b$
holds for at least a variable $b \in B$. This variable $b$ is not the same as
$a$ because $a \in A' = A \backslash B$ while $b \in B$. Also, $b \not\in A
\backslash \{a\}$ because otherwise $(A \backslash \{a\}) \rightarrow b$ would
be a tautology. Since $b \in B$ but $b \not\in A$, it follows $b \in B
\backslash A = B'$. All of this proves that for every $a \in A'$ the semantical
graph of $F$ contains at least an edge from $a$ to an element of $B'$.

By symmetry, the same applies to every element of $B'$. Following these edges,
an element of $A'$ leads to an element of $B'$, which leads to an element of
$A'$ and so on. Since the number of variables is finite, at some point this
path leads to a previous variable, forming a cycle.~\qed

It would be nice if the converse of this lemma also holds, giving a necessary
and sufficient condition to cyclicity. Unfortunately, this is not the case, as
Formula~\ref{cyclic-with-ab} shows.

\[
F = \{ab \rightarrow x, bx \rightarrow c, ac \rightarrow d, d \rightarrow x\}
\]

This formula $F$ is inequivalent and cyclic as proved by the next lemma.
Cyclicity is evident when $F$ is show graphically, but has to be proved since
semantic cyclicity does not follow from syntactic cyclicity.

%
%
\setlength{\unitlength}{5000sp}%
\begingroup\makeatletter\ifx\SetFigFont\undefined%
\gdef\SetFigFont#1#2#3#4#5{%
  \reset@font\fontsize{#1}{#2pt}%
  \fontfamily{#3}\fontseries{#4}\fontshape{#5}%
  \selectfont}%
\fi\endgroup%
\begin{picture}(1826,1284)(7340,-7225)
\thinlines
{\color[rgb]{0,0,0}\put(7426,-6466){\oval(152,660)[tl]}
\put(8101,-6466){\oval(1502,1502)[bl]}
\put(8101,-6466){\oval(1502,1502)[br]}
\put(8851,-6466){\oval(  2, 60)[tr]}
}%
{\color[rgb]{0,0,0}\put(7576,-6136){\line( 2,-3){150}}
\put(7726,-6361){\line(-2,-3){150}}
}%
{\color[rgb]{0,0,0}\put(7726,-6361){\vector( 1, 1){300}}
}%
{\color[rgb]{0,0,0}\put(8176,-6061){\vector( 1,-1){300}}
}%
{\color[rgb]{0,0,0}\put(9076,-6286){\vector(-3, 1){900}}
}%
{\color[rgb]{0,0,0}\put(7651,-6661){\line( 3, 1){450}}
\put(8101,-6511){\line( 0, 1){375}}
}%
{\color[rgb]{0,0,0}\put(8101,-6511){\vector( 1, 0){375}}
}%
{\color[rgb]{0,0,0}\put(8626,-6436){\vector( 1, 0){450}}
}%
\put(7501,-6736){\makebox(0,0)[b]{\smash{{\SetFigFont{12}{24.0}
{\rmdefault}{\mddefault}{\updefault}{\color[rgb]{0,0,0}$b$}%
}}}}
\put(8101,-6061){\makebox(0,0)[b]{\smash{{\SetFigFont{12}{24.0}
{\rmdefault}{\mddefault}{\updefault}{\color[rgb]{0,0,0}$x$}%
}}}}
\put(7501,-6136){\makebox(0,0)[b]{\smash{{\SetFigFont{12}{24.0}
{\rmdefault}{\mddefault}{\updefault}{\color[rgb]{0,0,0}$a$}%
}}}}
\put(8551,-6511){\makebox(0,0)[b]{\smash{{\SetFigFont{12}{24.0}
{\rmdefault}{\mddefault}{\updefault}{\color[rgb]{0,0,0}$c$}%
}}}}
\put(9151,-6436){\makebox(0,0)[b]{\smash{{\SetFigFont{12}{24.0}
{\rmdefault}{\mddefault}{\updefault}{\color[rgb]{0,0,0}$d$}%
}}}}
\end{picture}%
\nop{
+-- a
|   +--> x <---------------+
|   b     |                |
|    -----+---> c ---+---> d
|                    |
+--------------------+
}

\begin{lemma}
\label{inequivalent-cyclic}

Formula~\ref{cyclic-with-ab}
{} ($F =
{} \{ab \rightarrow x, bx \rightarrow c, ac \rightarrow d, d \rightarrow x\}$)
is inequivalent (Condition~\ref{condition-inequivalent}) and semantically
cyclic.

\end{lemma}

\proof The cycle in $F$ is
{} $x \rightarrow c, c \rightarrow d, d \rightarrow x \in E(F)$.
The first edge $x \rightarrow c$ is in $E(F)$ because $F$ contains $xb
\rightarrow c$ but does not entail $b \rightarrow c$. The second edge $c
\rightarrow d$ is in $E(F)$ because $F$ contains $ac \rightarrow d$ but does
not entail $a \rightarrow d$. The third edge $d \rightarrow x$ is in $E(F)$
because $F$ contains $d \rightarrow x$ but does not entail $x$.

\nojournal

This formula contains a cycle, which implies some sort of equivalence. Why is
$F$ inequivalent, then? For example, a cycle comprising $x$ and $d$ means that
$x$ entails $d$ with other variables and $d$ entails $x$ with other variables.
If the set of all these other variables is $O$, then $O \cup \{x\}$ and $F$
imply $O \cup \{d\}$ and vice versa: $F \models (O \cup \{x\}) \equiv (O \cup
\{d\})$. Differing but equivalent sets of variables suggest that inequivalence
does not hold, but are not enough: they should also not be equivalent to their
intersection. In this case, $O$ comprises $a$ and $b$, which alone imply all
other variables. Therefore, $F$ also implies $O \equiv (O \cup \{x\}) \equiv (O
\cup \{d\})$, making inequivalence true. The same applies to every other pair
of variables in the cycle. The cycle does not contradict inequivalence because
it requires some variables that imply all variables in the cycle.

\endnojournal

Inequivalence is formally proved for every pair of sets of variables $A$ and
$B$: if $F$ entails $A \equiv B$ then it also entails $A \equiv A \cap B$.

If $A$ contains both $a$ and $b$, it entails all variables. If $B$ does not
contain both $a$ and $b$, it is not equivalent to $A$ since $a$ and $b$ are not
heads of any clause. Therefore, both $A$ and $B$ contain both $a$ and $b$.
Their intersection contains $\{a,b\}$, and is therefore equivalent to $A$
because it entails all variables like $A$ does. This proves inequivalence for
all pairs of sets where one of them contains both $a$ and $b$.

The remaining sets may contain $a$ or $b$ but not both. Since $a$
is not entailed by any other variables, if $a \in A$ then $a \in B$, since
otherwise $F \models B \rightarrow A$ does not hold. The same holds in the
other direction, and also for $b$ for the same reason. In other words, the
remaining cases are: both $A$ and $B$ contain $a$ but not $b$, they both contain
$b$ but not $a$, and they contain neither.

In the first case, since neither $A$ nor $B$ contain $a$ and no clause has $a$
in the head, the clauses $ab \rightarrow x$ and $ac \rightarrow d$ are not
relevant. Therefore, inequivalence of $F$ is the same as inequivalence of
{} $F \backslash \{ab \rightarrow x, ac \rightarrow d\} = 
{}  \{bx \rightarrow c, d \rightarrow x\}$.
This formula is syntactically acyclic. Therefore, it is semantically acyclic by
Lemma~\ref{cycle-semantic-syntactic} and consequently inequivalent by
Lemma~\ref{acyclic-inequivalent}.

The same applies to the second case, where the clauses not containing $b$ as
preconditions are $\{ac \rightarrow d, d \rightarrow x\}$, and are again
acyclic and therefore inequivalent. In the third case the only clause left is
$\{d \rightarrow x\}$, and again acyclicity implies inequivalence. This shows
that $F$ is inequivalent.~\qed

Inequivalence is not the same as semantic acyclicity. Yet, a following section
shows it useful in extending the range of tractability of single-head
equivalence, from semantically acyclic to inequivalent formulae.

A further use of inequivalence is to answer a question from a previous section:
are Condition~\ref{first-condition} and~\ref{second-condition} sufficient to
single-head equivalence? Lemma~\ref{inequivalent-cyclic} shows that
Formula~\ref{cyclic-with-ab} is inequivalent; the next lemma shows that
inequivalence implies Condition~\ref{second-condition}.

\begin{lemma}
\label{inequivalent-second-condition}

Every inequivalent formula satisfies Condition~\ref{second-condition}.

\end{lemma}

\proof The condition to be proved is that $F \models A \rightarrow B$ implies
the existence of a set $C \in \equiset(B,F)$ such that $C \backslash
\equiall(B,F) \subseteq A$, where:

\begin{eqnarray*}
\equiset(A,F) &=& \{B \mid F \models A \equiv B\}		\\
\equiall(A,F) &=& \bigcap \equiset(B,F)
\end{eqnarray*}

The set $\bigcap \equiset(B,F)$ is proved equivalent to $B$ if $F$ is
inequivalent. By definition, $B' \in \equiset(B,F)$ if and only if $F \models B
\equiv B'$. By inequivalence, $F \models B \equiv B \cap B'$. Let $B''$ be
another element of $\equiset(B,F)$. Since $F \models B \equiv B''$,
transitivity implies $F \models B \cap B' \equiv B''$. By inequivalence again,
$F \models B \cap B' \equiv B \cap B' \cap B''$, and then by transitivity $F
\models B \equiv B \cap B' \cap B''$. This procedure can be iterated over all
elements of $\equiset(B,F)$, proving that $F$ makes their intersection
equivalent to $B$, that is, $F \models B \equiv \bigcap \equiset(B,F)$. This
entailment can be rewritten $F \models B \equiv \equiall(B,F)$, which proves
$\equiall(B,F) \in \equiset(B,F)$.

The required set $C$ is $C = \equiall(B,F)$, since it is in $\equiset(B,F)$ and
$C \backslash \equiall(B,F) = \emptyset \subseteq A$.~\qed

Lemma~\ref{ab-first} proves that Formula~\ref{cyclic-with-ab} satisfies
Condition~\ref{first-condition}. Lemma~\ref{inequivalent-cyclic} shows that it
is inequivalent, and all inequivalent formulae satisfy
Condition~\ref{second-condition} by Lemma~\ref{inequivalent-second-condition}.
Although this formula satisfies both conditions, it is not single-head
equivalent as proved by Lemma~\ref{ab-nonsingle}.

\begin{corollary}

Some inequivalent formulae satisfy both Condition~\ref{first-condition} and
Condition~\ref{second-condition} but are not single-head equivalent.

\end{corollary}

\subsection{Acyclicity and redundancy}

Theorem~\ref{no-subset} proves that removing clauses may not make a single-head
equivalent formula single-head. It proves that some formulae are equivalent to
a single-head formula but none of their equivalent subsets is single-head. Its
proof uses a cyclic formula. Does it hold on acyclic formulae as well? For
semantically acyclic formula, yes. A counterexample shows a formula that is
irredundant and not single-head but equivalent to a single-head formula.

\begin{lemma}

Some semantically acyclic irredundant formulae are not single-head, but they
are equivalent to a single-head formula.

\end{lemma}

\proof A formula that meets the statement of the lemma is
{} $F = \{bx \rightarrow a, b \rightarrow x, a \rightarrow x\}$.
None of its clauses is entailed by the others.

In spite of the syntactic cycle made of $x \rightarrow a$ and $a \rightarrow
x$, it is semantically acyclic because the first edge $x \rightarrow a$ is only
due to $bx \rightarrow a$, but the formula entails $b \rightarrow a$.

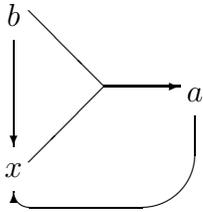
\setlength{\unitlength}{5000sp}%
\begingroup\makeatletter\ifx\SetFigFont\undefined%
\gdef\SetFigFont#1#2#3#4#5{%
  \reset@font\fontsize{#1}{#2pt}%
  \fontfamily{#3}\fontseries{#4}\fontshape{#5}%
  \selectfont}%
\fi\endgroup%
\begin{picture}(957,1079)(5209,-4425)
\put(5251,-3511){\makebox(0,0)[b]{\smash{{\SetFigFont{12}{24.0}
{\rmdefault}{\mddefault}{\updefault}{\color[rgb]{0,0,0}$b$}%
}}}}
\put(5251,-4261){\makebox(0,0)[b]{\smash{{\SetFigFont{12}{24.0}
{\rmdefault}{\mddefault}{\updefault}{\color[rgb]{0,0,0}$x$}%
}}}}
\put(6151,-3886){\makebox(0,0)[b]{\smash{{\SetFigFont{12}{24.0}
{\rmdefault}{\mddefault}{\updefault}{\color[rgb]{0,0,0}$a$}%
}}}}
\thinlines
{\color[rgb]{0,0,0}\put(5251,-4336){\vector( 0, 1){0}}
\put(5554,-4336){\oval(606,158)[bl]}
\put(5554,-3961){\oval(1194,908)[br]}
}%
{\color[rgb]{0,0,0}\put(5251,-3586){\vector( 0,-1){525}}
}%
{\color[rgb]{0,0,0}\put(5326,-3436){\line( 1,-1){375}}
\put(5701,-3811){\line(-1,-1){375}}
}%
{\color[rgb]{0,0,0}\put(5701,-3811){\vector( 1, 0){375}}
}%
\end{picture}%
\nop{
b ---+
|    |
|    +---> a
V    |     |
x ---+     |
^          |
+----------+
}

The variable $x$ is the head of two clauses of $F$. An equivalent formula is
{} $F' = \{b \rightarrow a, a \rightarrow x\}$,
since $bx \rightarrow a$ and $b \rightarrow x$ resolve into $b \rightarrow a$,
which subsumes $bx \rightarrow a$.

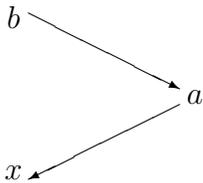
\setlength{\unitlength}{5000sp}%
\begingroup\makeatletter\ifx\SetFigFont\undefined%
\gdef\SetFigFont#1#2#3#4#5{%
  \reset@font\fontsize{#1}{#2pt}%
  \fontfamily{#3}\fontseries{#4}\fontshape{#5}%
  \selectfont}%
\fi\endgroup%
\begin{picture}(957,1079)(5209,-4425)
\put(5251,-3511){\makebox(0,0)[b]{\smash{{\SetFigFont{12}{24.0}
{\rmdefault}{\mddefault}{\updefault}{\color[rgb]{0,0,0}$b$}%
}}}}
\put(5251,-4261){\makebox(0,0)[b]{\smash{{\SetFigFont{12}{24.0}
{\rmdefault}{\mddefault}{\updefault}{\color[rgb]{0,0,0}$x$}%
}}}}
\put(6151,-3886){\makebox(0,0)[b]{\smash{{\SetFigFont{12}{24.0}
{\rmdefault}{\mddefault}{\updefault}{\color[rgb]{0,0,0}$a$}%
}}}}
\thinlines
{\color[rgb]{0,0,0}\put(5326,-3436){\vector( 2,-1){750}}
}%
{\color[rgb]{0,0,0}\put(6076,-3886){\vector(-2,-1){750}}
}%
\end{picture}%
\nop{
b ---+
     |
     +---> a
           |
x          |
^          |
+----------+
}

This formula is single-head and equivalent to $F$.~\qed

The statement of this lemma is the same as that of Theorem~\ref{no-subset} with
the additional condition of semantical acyclicity. It shows a semantically
acyclic formula that is not equivalent to any of its proper subsets. Its only
equivalent subset is itself. It is not single-head but is equivalent to a
single-head formula.

\begin{corollary}

Some semantically acyclic formulae are equivalent to single-head formulae, but
no equivalent subset of them is single-head.

\end{corollary}

The formula in the proof of the lemma is semantically acyclic but syntactically
cyclic. Irredundancy is a syntactical property: no clause of the formula is
entailed by the others; it is not maintained when switching from a formula to
an equivalent one. Being syntactic, irredundancy matches syntactical acyclicity
more than semantical acyclicity. The property that is disproved for general and
semantically acyclic formula holds for syntactically acyclic formulae: they are
single-head equivalent if and only if they have a single-head subset.

More generally, syntactically acyclic formulae do not present the main trap in
single-head equivalence, the possibility that a clause $B \rightarrow x$ seems
to follow from $A \rightarrow x \in F$ via $F \models B \rightarrow A$ but it
does not because $F \models B \rightarrow A$ requires $B \rightarrow x$ itself.
The following lemma excludes such a dependency in certain conditions.

\begin{lemma}
\label{keep-subset}

If
{} $F \models A \rightarrow x$
and
{} $y \rightarrow x \not\in E(F \backslash \{B \rightarrow y\})$
then
{} $F \backslash \{B \rightarrow y\} \models A \rightarrow x$.

\end{lemma}

\proof The starting point is the trivial entailment $y \models B \rightarrow
y$. A consequence of it is
{} $F \backslash \{B \rightarrow y\} \cup \{y\} \models
{}  F \backslash \{B \rightarrow y\} \cup \{B \rightarrow y\}$.
The entailed formula is $F$, which entails $A \rightarrow x$ by assumption.
Transitivity implies
{} $F \backslash \{B \rightarrow y\} \cup \{y\} \models A \rightarrow x$,
which is the same as
{} $F \backslash \{B \rightarrow y\} \models (A \cup \{y\}) \rightarrow x$.

This entailment with
{} $F \backslash \{B \rightarrow y\} \not\models A \rightarrow x$
implies
{} $y \rightarrow x \in E(F \backslash \{B \rightarrow y\})$.
Since this consequence is false, the first premise is true; therefore, the
second premise is false:
{} $F \backslash \{B \rightarrow y\} \models A \rightarrow x$.
This is the claim.~\qed

This lemma involves a semantic graph, flashing the vision of properties of
semantically acyclic formulae. It indeed proves that certain entailments from
$F$ carry over to $F \backslash \{B \rightarrow x\}$, but its premise is on the
semantical graph of $F \backslash \{B \rightarrow x\}$, not of $F$.

At a first sight, the premise $y \rightarrow x \not\in E(F \backslash \{B
\rightarrow x\})$ of the lemma looks like a consequence of $y \rightarrow x \in
E(F)$. It is not.

This would be the case if the semantic graph were monotonic: a subformula has a
subset of the edges of the superformula. This is disproved by the following
lemma.

\begin{lemma}
\label{semantic-monotone}

There exists two formulae $F$ and $F'$ such that $F \subseteq F'$ but $E(F)
\not\subseteq E(F')$.

\end{lemma}

\proof The two formulae are
{} $F = \{ay \rightarrow x\}$
and
{} $F' = \{ay \rightarrow x, a \rightarrow x\}$.
The former has the edge $y \rightarrow x \in E(F)$ since it entails $ay
\rightarrow x$ but not $a \rightarrow x$. The second does not have that edge
because it entails $a \rightarrow x$.~\qed

A similar counterexample fails Lemma~\ref{keep-subset} if its premise $y
\rightarrow x \not\in E(F \backslash \{B \rightarrow y\})$ is replaced by $y
\rightarrow x \not\in E(F)$. The formula is
{} $F = \{ay \rightarrow x, a \rightarrow b, b \rightarrow y\}$,
the removed clause is $B \rightarrow x = b \rightarrow y$. The premises of the
modified lemma are satisfied: $F \models a \rightarrow x$ and $y \rightarrow x
\not\in E(F)$; the second holds in spite of $ay \rightarrow x \in F$ because of
$F \models a \rightarrow x$. Yet, its conclusion $F \backslash \{b \rightarrow
y\} \models a \rightarrow x$ is false.

For its intended usage the premise $y \rightarrow x \not\in E(F)$ should hold
on the whole formula, not the formula without the clause. The counterexample
shows that the lemma does not extend this way. It does not because the semantic
graph is not monotonic with respect to the formula as proved by
Lemma~\ref{semantic-monotone}. The syntactic graph is, as proved by
Lemma~\ref{syntactic-monotone}. This allows extending Lemma~\ref{keep-subset}
to
{} $y \rightarrow x \not\in Y^*(F)$.

\begin{lemma}
\label{keep}

If
{} $F \models A \rightarrow x$
and
{} $y \rightarrow x \not\in Y^*(F)$
then
{} $F \backslash \{B \rightarrow y\} \models A \rightarrow x$.

\end{lemma}

\proof The definition of
{} $y \rightarrow x \not\in Y^*(F)$
is that no sequence of clauses connects $y$ to $x$. By
Lemma~\ref{syntactic-monotone}, removing clauses may only remove edges. This
implies
{} $y \rightarrow x \not\in Y^*(F \backslash \{B \rightarrow y\})$.

Lemma~\ref{syntactic-semantic} tells that
{} $y \rightarrow x \not\in Y^*(F \backslash \{B \rightarrow y\})$
implies
{} $y \rightarrow x \not\in E(F \backslash \{B \rightarrow y\})$.

The preconditions of Lemma~\ref{keep-subset} are satisfied by the formula $F
\backslash \{B \rightarrow y\}$. Its consequence is
{} $F \backslash \{B \rightarrow y\} \models A \rightarrow x$.~\qed

Lemma~\ref{keep} cuts the vicious loop of single-head equivalence:
syntactically acyclic formulae never require $B \rightarrow x$ to entail $B
\rightarrow A$ if they entail both $B \rightarrow x$ and $A \rightarrow x$. It
kills all the subtlety of single-head equivalence.

A consequence is that irredundancy makes single-head the same as single-head
equivalence for syntactically acyclic formula.

\begin{theorem}
\label{ayclic-irredundant}

If an irredundant syntactically acyclic formula is single-head equivalent,
it is single-head.

\end{theorem}

\proof The proof assumes that $F$ is syntactically acyclic, not single-head but
equivalent to a single-head formula $F'$; the conclusion is that $F$ is
redundant.

Let $x$ be a variable that is the head of two clauses of $F$.

Since $F'$ is single-head, it contains either a single clause $A \rightarrow x$
with head $x$ or none.

The latter case is analyzed first: $F'$ does not contain any clause with head
$x$. Let $B \rightarrow x$ be one of the two clauses of $F$ with head $x$. If
$x \in B$ then $B \rightarrow x$ is a tautology and $F$ is redundant, which is
the claim. Otherwise, $x \not\in B$. Since $F'$ is equivalent to $F$, it
entails $B \rightarrow x$. Lemma~\ref{set-implies-set} proves $F'^x \models B
\rightarrow A$ with $A \rightarrow x \in F'$, contradicting the assumption that
$F'$ does not contain any clause of head $x$.

The other case is that $F'$ contains a single clause $A \rightarrow x$ with
head $x$.

If $x \in A$ then $A \rightarrow x$ is tautologic and therefore redundant: $F'$
is equivalent to $F' \backslash \{A \rightarrow x\}$. This formula is
equivalent to $F$ and does not contain clauses with $x$ in the head. The
argument in the paragraph above proves that if $F$ is equivalent to a formula
that has no clause with head $x$ then $F$ is redundant.

The remaining case is that $F'$ contains a single clause $A \rightarrow x$ and
$x \not\in A$.

If $F'$ also entails a clause $A' \rightarrow x$ with $A' \subset A$ then $F'$
is equivalent to
{} $F' \equiv F' \backslash \{A \rightarrow x\} \cup \{A' \rightarrow x\}$
since $A' \rightarrow x \models A \rightarrow x$ always holds and $F' \models
A' \rightarrow x$ holds by assumption. Since the head of these two clauses is
the same, the second formula is also single-head. Iteratively, this argument
proves that if $F$ is single-head equivalent it is also equivalent to a
single-head formula whose clauses $A \rightarrow x$ are minimal: this formula
does not entail any clause $A' \rightarrow x$ with $A' \subset A$.

Membership implies entailment: $A \rightarrow x \in F'$ implies $F' \models A
\rightarrow x$. Since $F'$ is equivalent to $F$, they entail the same clauses,
including $A \rightarrow x$. Since $F'$ does not entail any clause $A'
\rightarrow x$ with $A' \subset A$, the same holds for $F$. As a result, $F$
entails $A \rightarrow x$ but not $A \backslash \{a\} \rightarrow x$ if $a \in
A$. This is the definition of $a \rightarrow x \in E(F)$. It implies $a
\rightarrow x \in Y^*(F)$ by Lemma~\ref{syntactic-semantic}. This holds for
every $a \in A$.

Since $F$ is equivalent to $F'$, it entails $A \rightarrow x$. Since the case
$x \in A$ is already excluded, Lemma~\ref{set-implies-set} proves that $F$
contains a clause $B \rightarrow x$ such that $F^x \models A \rightarrow B$.

By assumption, $F$ contains two clauses of head $x$. One is $B \rightarrow x$.
Let $C \rightarrow x$ be the other. If $x \in C$ then this clause is a
tautology and $F$ is redundant, which is the claim. Otherwise, $x \not\in C$.
Since $F'$ is equivalent to $F$, it entails $C \rightarrow x$. By
Lemma~\ref{set-implies-set}, $F'$ contains a clause $D \rightarrow x$ such that
$F' \models C \rightarrow D$. Since $F'$ is single-head and contains $A
\rightarrow x$, this clause $D \rightarrow x$ is $A \rightarrow x$. The
conclusion is $F' \models C \rightarrow A$. Since $F$ is equivalent to $F'$, it
entails $C \rightarrow A$.

The entailment $F \models C \rightarrow A$ is the same as $F \models C
\rightarrow a$ for every $a \in A$. By Lemma~\ref{keep}, if $F \backslash \{C
\rightarrow x\} \not\models C \rightarrow a$ then $x \rightarrow a \in Y^*(F)$.
With the edge $a \rightarrow x$ already proved in $Y^*(F)$, this forms a cycle
in $Y^*(F)$. Since $F$ is syntactically acyclic, the assumption $F \backslash
\{C \rightarrow x\} \not\models C \rightarrow a$ is false. The contrary is
true: $F \backslash \{C \rightarrow x\} \models C \rightarrow a$. This being
the case for every $a \in A$, the conclusion is $F \backslash \{C \rightarrow
x\} \models C \rightarrow A$.

The following are proved so far:
{} $F \backslash \{C \rightarrow x\} \models C \rightarrow A$,
{} $F^x \models A \rightarrow B$ and
{} $B \rightarrow x \in F$.
Since $C \rightarrow x$ contains $x$, it is not in $F^x$. Therefore, $F^x
\subseteq F \backslash \{C \rightarrow x\}$. A consequence is $F \backslash \{C
\rightarrow x\} \models A \rightarrow B$. Since $C \rightarrow x$ is a clause
different from $B \rightarrow x$, the condition $B \rightarrow x \in F$
implies $B \rightarrow x \in F \backslash \{C \rightarrow x\}$.

The three conclusions
{} $F \backslash \{C \rightarrow x\} \models C \rightarrow A$,
{} $F \backslash \{C \rightarrow x\} \models A \rightarrow B$ and
{} $B \rightarrow x \in F \backslash \{C \rightarrow x\}$
imply $F \backslash \{C \rightarrow x\} \models C \rightarrow x$. Since $C
\rightarrow x$ is a clause of $F$ by assumption, $F$ is redundant.~\qed

An immediate consequence of this theorem is that single-head equivalent is the
same as single-head for syntactically acyclic and irredundant formulae. It
provides a simple algorithm for checking single-head equivalence on
syntactically acyclic formulae: their redundant clauses are removed one by one
in any order and the result is checked for the presence of clauses with the
same head.

Does something similar exist for semantically acyclic formulae?

Irredundancy is syntactic, like syntactic acyclicity. The semantic version of
syntactic acyclicity is semantic acyclicity. What is the semantic version of
irredundancy?

Irredundancy is syntactic minimality. A formula is irredundant if it has no
equivalent proper subset; it has no smaller equivalent formula made only of its
clauses. It is syntactic because of the latter point: ``made only of its
clauses''. When the specific clauses of a formula matter, syntax is in.

If irredundancy is syntactic minimality, what is semantic minimality? When the
individual clauses of the formula do not matter, only size remains. Minimal is
minimal by number of literal occurrencies. Which is the same as minimal by
number of clauses as Hammer and Kogan~\cite{hamm-koga-95} proved for
semantically acyclic formula.

%
%
%

If this analogy works, minimality matches semantical acyclicity in the same way
irredundancy matches syntactic acyclicity. A semantical acyclic and minimal
formula is single-head equivalent if and only if it is single-head. That this
analogy works is what the next two lemmas prove.

The first lemma shows that semantic acyclicity implies syntactic acyclicity for
minimal formulae. The converse is always the case.

\begin{lemma}
\label{minimal-semantic-syntactic}

If a definite Horn formula is not equivalent to any formula with less literal
occurrences and is semantically acyclic then it is also syntactically acyclic.

\end{lemma}

\proof Let $F$ be a minimal formula. The first step of the proof is that it
only contains prime implicates. First, it only contains some of its implicates
because it entails every clause it contains. Contrary to the claim, it is
assumed to entail a proper subset $C'$ of a clause $C$ it contains. Since
subclauses entail superclauses, $C'$ entails $C$. As a result,
{} $F \backslash \{C\} \cup \{C'\}$
entails
{} $F \backslash \{C\} \cup \{C\} = F$.
The converse is also the case since by assumption $F$ entails $C'$, the only
clause of
{} $F \backslash \{C\} \cup \{C'\}$
it does not contain. This makes $F$ equivalent to $F \backslash \{C\} \cup
\{C'\}$, which is smaller than $F$ because it contains $C' \subset C$ in place
of $C$. This contradicts the assumption that $F$ is minimal. The conclusion is
that $F$ only contains some of its prime implicates.

This conclusion can be written $F \subseteq PI(F)$ where $PI(F)$ is the set of
the prime implicates of $F$. This implies $PI(F) \models F$ by monotonicity.
The converse is also the case since $PI(F)$ is only made of clauses of $F$ by
definition. Entailment in both direction is equivalence: $F \equiv PI(F)$.

If $F$ is semantically acyclic, also $PI(F)$ is semantically acyclic because
they are equivalent and semantic acyclicity does not depend on the syntax.

By Lemma~\ref{prime}, since $PI(F)$ coincides with its set of prime implicates
and is semantically acyclic it is also syntactically acyclic. Since $F$ is a
subset of $PI(F)$, the same containment holds on their syntactic graphs by
Lemma~\ref{syntactic-monotone}: $Y(F) \subseteq Y(PI(F))$. If $Y(F)$ contains a
cycle, then $PI(F)$ contains the same cycle. This is not possible because
$PI(F)$ is syntactically acyclic. Therefore, $F$ is syntactically acyclic.~\qed

Why this lemma? Proving that semantic acyclicity implies syntactic acyclicity,
it makes Theorem~\ref{ayclic-irredundant} applicable: a syntactic acyclic
irredundant formula is single-head if it is single-head equivalent.

\begin{lemma}

If a definite Horn formula is equivalent to no formula with less literal
occurrences, is semantically acyclic and single-head equivalent, then it is
single-head.

\end{lemma}

\proof By Lemma~\ref{minimal-semantic-syntactic}, a formula that is minimal and
semantically acyclic is also syntactically acyclic. Since it is minimal, it is
irredundant. Being irredundant, syntactically acyclic and single-head
equivalent, it is single-head by Theorem~\ref{ayclic-irredundant}.~\qed

On minimal semantically acyclic formulae, single-head equivalent implies
single-head. Single-head implies single-head equivalent by definition.
Therefore, single-head equivalence coincides with single-head on minimal
semantically acyclic formulae. Except that checking for being single-head
requires only a simple scan of the formula for clauses with the same head.

The algorithm for single-head equivalence of a semantically acyclic formula is
similar to that for syntactically acyclic formulae. Minimality takes the place
of irredundancy: first the formula is made minimal, then the presence of
clauses of the same head is checked.

Hammer and Kogan~\cite{hamm-koga-95} provided a quadratic algorithm for finding
the minimal formula equivalent to a given one. Checking for clauses with the
same head is quadratic as well, making single-head equivalence checking
quadratic in running time.

\section{The order and the algorithm}
\label{order}

\draft

summary of results:

\begin{itemize}

\item even a single-head equivalent formula may take exponential time to the
body replace algorithm; turning it into its single-head form decreases time to
polynomial

\item $A \leq_F B$ means $F \models B \rightarrow A$

\item this order is relevant because $A$ is minimal among the sets implying $x$
in every single-head formula that contains $A \rightarrow x$; therefore, it is
also minimal in all equivalent formulae

\item the order is not total even on single-head formulae

\item lemma convergence:

If $F$ is a single-head equivalent formula and
{} $F \cup A \models x$ and $F \cup B \models x$,
there exists $C \subseteq A \cup B$ such that
{}   $F \cup A \models C$,
{}   $F \cup B \models C$ and
{}   $F \cup C \models x$.

\item $MIN(F)$ is the formula that contains $A \rightarrow x$ if $A$ is minimal
among the sets entailing $x$; minimal means that there is no $<_F$ or $\subset$
lower set

\begin{itemize}

\item lemma min-equivalent-single:

If $MIN(F)$ is single-head and equivalent to $F$,
then $F$ is single-head equivalent.

\item lemma single-min-equivalent:

If $F$ is single-head, then $F \equiv MIN(F)$.

\item lemma single-min-single:

If $F$ is single-head and inequivalent
(Condition~\ref{condition-inequivalent})
then $MIN(F)$ is single-head.

\item lemma inequivalent-min:

If $F$ is inequivalent, it is single-head equivalent if and only if $MIN(F)$ is
single-head and equivalent to $F$.

\end{itemize}

\item lemma ab-nonsingle proves that a formula previously used is not
single-head equivalent:

Formula~\ref{cyclic-with-ab} is not single-head equivalent.

\item the aim shifts to finding the elements of $MIN(F)$; this can be done
starting from the sets $A$ such that $A \rightarrow x \in F$, replacing it with
sets that are first $<_F$-less and then $\subset$-less that it; this can be
done by checking a variable at time:

\begin{itemize}

\item $BCN(A,F) = \{x \mid F \models A \rightarrow x\}$

\item checking for subsets of $A$ is easy, since it can be done by removing an
element at time from $A$ and checking whether $F \models A \backslash \{a\}
\rightarrow x$

\item this is not the case for checking the existence of lower elements $B <_F
A$ since the number of candidate sets $B$ is exponential

\item lemma less-minus-one-sufficient tells that some sets $B$ lower than $A$
are obtained by considering a variable at time:

If
{} $F \models BCN(A,F) \backslash \{a,x\} \rightarrow x$ and
{} $F \not\models BCN(A,F) \backslash \{a,x\} \rightarrow a$
where $a \in A$, then $B = BCN(A,F) \backslash \{a,x\}$ satisfies
{} $F \models B \rightarrow x$ and $B <_F A$.

\item lemma less-minus-one-not-necessary tells the converse: lower sets may
exist even if none of them is obtained by removing a single variable

There exists a single-head formula $F$, a set of variable $A$ and a variable
$x$ such that both
{} $F \models A \backslash \{a\} \rightarrow x$ and
{} $F \not\models BCN(A,F) \backslash \{a,x\} \rightarrow a$
are false for every $a \in A$, but there exists $B$ such that
{} $F \models B \rightarrow x$ and $B <_F A$.

\item lemma less-minus-one:

If $F$ is inequivalent (Condition~\ref{condition-inequivalent}) and entails $A
\rightarrow x$ with $x \not\in A$, it entails a clause $B \rightarrow x$ with
$x \not\in B$ and $B <_F A$ if and only if
{} $F \models BCN(A,F) \backslash \{a,x\} \rightarrow x$ and
{} $F \not\models BCN(A,F) \backslash \{a,x\} \rightarrow a$
for some $a \in A$.

\end{itemize}

\item $RCN(A,F) = \{a \mid F \cup (BCN(A,F) \backslash \{a\}) \models a\}$

\item lemma use-clause:

It holds $x \in RCN(B,F)$ if and only if $F$ contains a clause $B' \rightarrow
x$ such that $F \models B \rightarrow B'$.

\item lemma b-rcn:

For every formula $F$ and set of variables $B$, it holds
{} $BCN(B,F) = B \cup RCN(B,F)$.

\item lemma third-rcn:

If
{} $F \models BCN(A,F) \backslash \{a,x\} \rightarrow x$
with $a \in A$, then
{} $F \not\models BCN(A,F) \backslash \{a,x\} \rightarrow a$
is equivalent to 
$a \in A \backslash RCN(A,F)$.

\item lemma rcn-less

If $a \in A \backslash RCN(A,F)$ then $A \backslash \{a\} <_F A$.

\item algorithm for $RCN(B,F)$

\item lemma rcn-first:

The first return value of Algorithm~\ref{algorithm-rcnucl} is $RCN(B,F)$.

\item lemma rcn-second:

The second return value of Algorithm~\ref{algorithm-rcnucl} is
$\{B' \rightarrow x \in F \mid B' \subseteq BCN(B,F)\}$;

this result is not relevant to this article;
it is here just because the algorithm is here

\item the improved algorithm

\item it may look like two incomparable bodies or two incomparable minimal
bodies prove that the formula is not single-head equivalent, but this is not
the case; a valid condition is that no body in the formula is less than any
body of a generated clause

\end{itemize}

\enddraft

\subsection{Why bothering with single heads?}

After all this talking about single-head equivalence, about necessary
conditions to single-head equivalence, about the pitfalls of sufficient
conditions to single-head equivalence, about cyclicity and single-head
equivalence, about redundancy and minimality and single-head equivalence, after
all of this it may look like the question is only to establish single-head
equivalence --- decide whether a formula is equivalent to a single-head
formula. It says in advance whether the forgetting algorithm takes polynomial
time.

This is not the main question. Single-head equivalence does not even make the
forgetting algorithm polynomial. That requires formulae that are single-head,
not single-head equivalent.

The goal is not to establish single-head equivalence, but to find a single-head
equivalent formula, if any.

If any, the replacing algorithm takes polynomial time on it even if it does not
on the original formula. The translation from single-head equivalent to
single-head reduces the running time from exponential to polynomial. It is an
efficiency improvement of the forgetting algorithm, not just an efficiency
estimation.

This section gives an algorithm for turning a formula into a single-head
equivalent one. It is incomplete: it always produces a single-head formula
which may or may not be equivalent to the given one. If it is, the outcome is
not only a proof of single-head equivalence, but also the required single-head
equivalent formula.

The forgetting algorithm only takes polynomial time on it even if it takes
exponential time on the original one. This happens for example on the following
formula.

\[
\{
a \rightarrow b, b \rightarrow c, c \rightarrow d,
d \rightarrow e, e \rightarrow f,
a \rightarrow c, b \rightarrow d, c \rightarrow e, d \rightarrow f
\}
\]

A pictorial representation shows what is wrong with $F$: it is a chain of
clauses surrounded by redundant clauses. The redundant clauses are all entailed
by transitivity, and can be removed without affecting the semantics of the
formula. Yet, they are processed by the forgetting algorithm as if they were
essential.

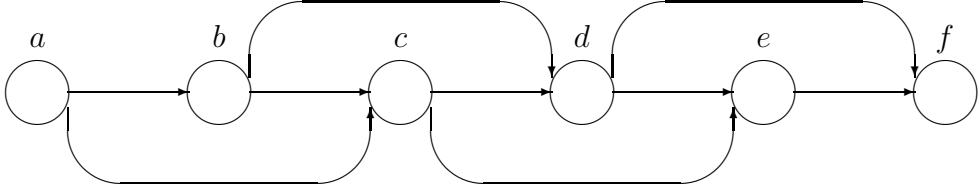
\setlength{\unitlength}{5000sp}%
\begingroup\makeatletter\ifx\SetFigFont\undefined%
\gdef\SetFigFont#1#2#3#4#5{%
  \reset@font\fontsize{#1}{#2pt}%
  \fontfamily{#3}\fontseries{#4}\fontshape{#5}%
  \selectfont}%
\fi\endgroup%
\begin{picture}(4852,918)(7625,-6520)
\thinlines
{\color[rgb]{0,0,0}\put(8701,-6136){\oval(1500,750)[bl]}
\put(8701,-6136){\oval(1500,750)[br]}
\put(9451,-6136){\vector( 0, 1){0}}
}%
{\color[rgb]{0,0,0}\put(10501,-6136){\oval(1500,750)[bl]}
\put(10501,-6136){\oval(1500,750)[br]}
\put(11251,-6136){\vector( 0, 1){0}}
}%
{\color[rgb]{0,0,0}\put(10351,-5986){\vector( 0,-1){0}}
\put(9601,-5986){\oval(1500,750)[tr]}
\put(9601,-5986){\oval(1500,750)[tl]}
}%
{\color[rgb]{0,0,0}\put(12151,-5986){\vector( 0,-1){0}}
\put(11401,-5986){\oval(1500,750)[tr]}
\put(11401,-5986){\oval(1500,750)[tl]}
}%
{\color[rgb]{0,0,0}\put(11401,-6061){\circle{336}}
}%
{\color[rgb]{0,0,0}\put(12301,-6061){\circle{336}}
}%
{\color[rgb]{0,0,0}\put(10501,-6061){\circle{336}}
}%
{\color[rgb]{0,0,0}\put(9601,-6061){\circle{336}}
}%
{\color[rgb]{0,0,0}\put(8701,-6061){\circle{336}}
}%
{\color[rgb]{0,0,0}\put(7801,-6061){\circle{336}}
}%
{\color[rgb]{0,0,0}\put(8851,-6061){\vector( 1, 0){600}}
}%
{\color[rgb]{0,0,0}\put(9751,-6061){\vector( 1, 0){600}}
}%
{\color[rgb]{0,0,0}\put(10651,-6061){\vector( 1, 0){600}}
}%
{\color[rgb]{0,0,0}\put(11551,-6061){\vector( 1, 0){600}}
}%
{\color[rgb]{0,0,0}\put(7951,-6061){\vector( 1, 0){600}}
}%
\put(7801,-5836){\makebox(0,0)[b]{\smash{{\SetFigFont{12}{24.0}
{\rmdefault}{\mddefault}{\updefault}{\color[rgb]{0,0,0}$a$}%
}}}}
\put(8701,-5836){\makebox(0,0)[b]{\smash{{\SetFigFont{12}{24.0}
{\rmdefault}{\mddefault}{\updefault}{\color[rgb]{0,0,0}$b$}%
}}}}
\put(9601,-5836){\makebox(0,0)[b]{\smash{{\SetFigFont{12}{24.0}
{\rmdefault}{\mddefault}{\updefault}{\color[rgb]{0,0,0}$c$}%
}}}}
\put(10501,-5836){\makebox(0,0)[b]{\smash{{\SetFigFont{12}{24.0}
{\rmdefault}{\mddefault}{\updefault}{\color[rgb]{0,0,0}$d$}%
}}}}
\put(11401,-5836){\makebox(0,0)[b]{\smash{{\SetFigFont{12}{24.0}
{\rmdefault}{\mddefault}{\updefault}{\color[rgb]{0,0,0}$e$}%
}}}}
\put(12301,-5836){\makebox(0,0)[b]{\smash{{\SetFigFont{12}{24.0}
{\rmdefault}{\mddefault}{\updefault}{\color[rgb]{0,0,0}$f$}%
}}}}
\end{picture}%
\nop{
+----------+ +----------+
|          V |          V
a --> b --> c --> d --> e --> f
      |          ^ |          ^
      +----------+ +----------+
}

The algorithm replaces each variable to forget with the body of a clause having
it as its head. If the variable is in the head of two clauses, each of them
provides a way to forget it. The algorithm non-deterministically tries both.

It forgets $e$ by replacing it in $e \rightarrow f$. Since $e$ is the head of
$c \rightarrow e$ and $d \rightarrow e$, it replaces it with $c$ in a
nondeterministic branch and with $d$ in another.

If $d$ is also to be forgotten, it branches again to replace it with either $c$
or $b$. To forget $c$ it branches once again in each of the four deterministic
branches, making them eight. This behavior can be observed on the first formula
of the {\tt singlehead.py} test file of the {\tt forget-fork.py} program, which
implements the replacing algorithm for forgetting~\cite{libe-20-a}. The version
of this formula with $n$ variables requires $O(2^n)$ branches.

This is not the case for the following formula, which is single-head and
equivalent to the previous.
It is in the {\tt chain.py} test file of the {\tt singlehead.py} program.

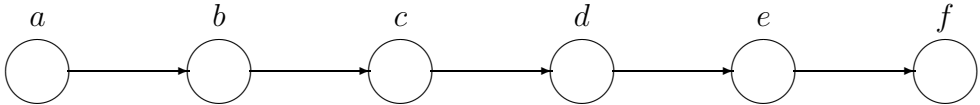
\[
\{a \rightarrow b, b \rightarrow c, c \rightarrow d,
d \rightarrow e, e \rightarrow f\}
\]

\setlength{\unitlength}{5000sp}%
\begingroup\makeatletter\ifx\SetFigFont\undefined%
\gdef\SetFigFont#1#2#3#4#5{%
  \reset@font\fontsize{#1}{#2pt}%
  \fontfamily{#3}\fontseries{#4}\fontshape{#5}%
  \selectfont}%
\fi\endgroup%
\begin{picture}(4852,918)(7625,-6520)
{\color[rgb]{0,0,0}\thinlines
\put(11401,-6061){\circle{336}}
}%
{\color[rgb]{0,0,0}\put(12301,-6061){\circle{336}}
}%
{\color[rgb]{0,0,0}\put(10501,-6061){\circle{336}}
}%
{\color[rgb]{0,0,0}\put(9601,-6061){\circle{336}}
}%
{\color[rgb]{0,0,0}\put(8701,-6061){\circle{336}}
}%
{\color[rgb]{0,0,0}\put(7801,-6061){\circle{336}}
}%
{\color[rgb]{0,0,0}\put(8851,-6061){\vector( 1, 0){600}}
}%
{\color[rgb]{0,0,0}\put(9751,-6061){\vector( 1, 0){600}}
}%
{\color[rgb]{0,0,0}\put(10651,-6061){\vector( 1, 0){600}}
}%
{\color[rgb]{0,0,0}\put(11551,-6061){\vector( 1, 0){600}}
}%
{\color[rgb]{0,0,0}\put(7951,-6061){\vector( 1, 0){600}}
}%
\put(7801,-5836){\makebox(0,0)[b]{\smash{{\SetFigFont{12}{24.0}
{\rmdefault}{\mddefault}{\updefault}{\color[rgb]{0,0,0}$a$}%
}}}}
\put(8701,-5836){\makebox(0,0)[b]{\smash{{\SetFigFont{12}{24.0}
{\rmdefault}{\mddefault}{\updefault}{\color[rgb]{0,0,0}$b$}%
}}}}
\put(9601,-5836){\makebox(0,0)[b]{\smash{{\SetFigFont{12}{24.0}
{\rmdefault}{\mddefault}{\updefault}{\color[rgb]{0,0,0}$c$}%
}}}}
\put(10501,-5836){\makebox(0,0)[b]{\smash{{\SetFigFont{12}{24.0}
{\rmdefault}{\mddefault}{\updefault}{\color[rgb]{0,0,0}$d$}%
}}}}
\put(11401,-5836){\makebox(0,0)[b]{\smash{{\SetFigFont{12}{24.0}
{\rmdefault}{\mddefault}{\updefault}{\color[rgb]{0,0,0}$e$}%
}}}}
\put(12301,-5836){\makebox(0,0)[b]{\smash{{\SetFigFont{12}{24.0}
{\rmdefault}{\mddefault}{\updefault}{\color[rgb]{0,0,0}$f$}%
}}}}
\end{picture}%
\nop{
a --> b --> c --> d --> e --> f
}

This formula is a chain of clauses.
As shown by the second formula of the {\tt singlehead.py} test file of the
{\tt forget-fork.py} program~\cite{libe-20-a},
the forgetting algorithm starts with $e \rightarrow f$ and turns it into $d
\rightarrow f$, then $c \rightarrow f$, $b \rightarrow f$ and finally into $a
\rightarrow f$, the expected result. It does not branch. It takes linear time.

A variant of the algorithm that adds all possible replacements instead of
branching on each would solve the problem, but the solution is worse than the
problem since it requires exponential space in general while the
nondeterministic algorithm runs in polynomial space.

The algorithm $SHMIN(F)$ shown at the end of this section turns $F$ into $F'$.
Success is not guaranteed: while $SHMIN(F)$ is always single-head, it is not
always equivalent to $F$. If it is, the replacing algorithm can be run on
$SHMIN(F)$ instead of $F$, reducing time from exponential to polynomial.
Otherwise, it is run on $F$ itself; the unsuccessful call to $SHMIN(F)$ only
adds polynomial time to the bare algorithm.

Even in this case, it is not always wasted time. While $SHMIN(F)$ may not be
equivalent to $F$, it is always entailed by it. Therefore, equivalence can be
achieved by repeatedly adding clauses of $F$ that are not entailed by
$SHMIN(F)$. The resulting formula is no longer single-head, but may still have
fewer clauses with the same head. If so, the non-deterministic branching of the
replacing algorithm decreases, improving efficiency.

\subsection{The induced order}

The single-head form (if any) of a formula can be found by ordering the sets of
literals.

If $F$ is single-head equivalent, it is equivalent to a single-head formula
$F'$. For each variable $x$, this formula $F'$ may only contain a single clause
$A \rightarrow x$ with $x$ as its head. If $F$ contains another clause $B
\rightarrow x$, then $F$ implies it. By equivalence, also $F'$ implies it: $F'
\models B \rightarrow x$. Since formulae are assumed not to contain
tautologies, Lemma~\ref{set-implies-set} implies $B' \rightarrow x \in F'$ with
$F \models B \rightarrow B'$. Since $F'$ is single-head, $B' \rightarrow x$ is
the same as $A \rightarrow x$. Therefore, $F' \models B \rightarrow A$. By
equivalence, $F \models B \rightarrow A$.

In summary, if $F$ contains a clause $B \rightarrow x$, then $F$ implies $B
\rightarrow A$ where $A \rightarrow x$ is the clause of head $x$ in $F'$.

The goal is to build $F'$ from $F$. To generate its clauses, like $A
\rightarrow x$, the above property helps. It tells that $A$ is a set of
literals such that $B \rightarrow x \in F$ implies $F \models B \rightarrow A$.
This restricts the range of possible sets $A$ to the ones at the end of chains
of implications from all other sets $B$ such that $B \rightarrow x \in F$.

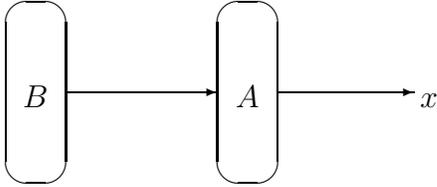
\setlength{\unitlength}{5000sp}%
\begingroup\makeatletter\ifx\SetFigFont\undefined%
\gdef\SetFigFont#1#2#3#4#5{%
  \reset@font\fontsize{#1}{#2pt}%
  \fontfamily{#3}\fontseries{#4}\fontshape{#5}%
  \selectfont}%
\fi\endgroup%
\begin{picture}(2127,924)(7189,-6223)
\thinlines
{\color[rgb]{0,0,0}\put(7306,-6106){\oval(210,210)[bl]}
\put(7306,-5416){\oval(210,210)[tl]}
\put(7396,-6106){\oval(210,210)[br]}
\put(7396,-5416){\oval(210,210)[tr]}
\put(7306,-6211){\line( 1, 0){ 90}}
\put(7306,-5311){\line( 1, 0){ 90}}
\put(7201,-6106){\line( 0, 1){690}}
\put(7501,-6106){\line( 0, 1){690}}
}%
{\color[rgb]{0,0,0}\put(8356,-6106){\oval(210,210)[bl]}
\put(8356,-5416){\oval(210,210)[tl]}
\put(8446,-6106){\oval(210,210)[br]}
\put(8446,-5416){\oval(210,210)[tr]}
\put(8356,-6211){\line( 1, 0){ 90}}
\put(8356,-5311){\line( 1, 0){ 90}}
\put(8251,-6106){\line( 0, 1){690}}
\put(8551,-6106){\line( 0, 1){690}}
}%
{\color[rgb]{0,0,0}\put(7501,-5761){\vector( 1, 0){750}}
}%
{\color[rgb]{0,0,0}\put(8551,-5761){\vector( 1, 0){675}}
}%
\put(7351,-5836){\makebox(0,0)[b]{\smash{{\SetFigFont{12}{24.0}
{\rmdefault}{\mddefault}{\updefault}{\color[rgb]{0,0,0}$B$}%
}}}}
\put(9301,-5836){\makebox(0,0)[b]{\smash{{\SetFigFont{12}{24.0}
{\rmdefault}{\mddefault}{\updefault}{\color[rgb]{0,0,0}$x$}%
}}}}
\put(8401,-5836){\makebox(0,0)[b]{\smash{{\SetFigFont{12}{24.0}
{\rmdefault}{\mddefault}{\updefault}{\color[rgb]{0,0,0}$A$}%
}}}}
\end{picture}%
\nop{
B --> A --> x
}

Writing $F \models B \rightarrow A$ to $A \leq_F B$ allows reformulating the
aim as: for each variable $x$, find a set of variables $A$ that is minimal
according to $\leq_F$.

What complicates the search is that the order may not be total. The
counterexample is
{} $F = \{a \rightarrow b, c \rightarrow d, bd \rightarrow e\}$.

\setlength{\unitlength}{5000sp}%
\begingroup\makeatletter\ifx\SetFigFont\undefined%
\gdef\SetFigFont#1#2#3#4#5{%
  \reset@font\fontsize{#1}{#2pt}%
  \fontfamily{#3}\fontseries{#4}\fontshape{#5}%
  \selectfont}%
\fi\endgroup%
\begin{picture}(3202,1765)(6125,-6536)
{\color[rgb]{0,0,0}\thinlines
\put(7651,-5161){\circle{336}}
}%
{\color[rgb]{0,0,0}\put(9151,-5761){\circle{336}}
}%
{\color[rgb]{0,0,0}\put(7651,-6361){\circle{336}}
}%
{\color[rgb]{0,0,0}\put(6301,-5161){\circle{336}}
}%
{\color[rgb]{0,0,0}\put(6301,-6361){\circle{336}}
}%
{\color[rgb]{0,0,0}\put(7801,-5236){\line( 1,-1){525}}
\put(8326,-5761){\line(-1,-1){525}}
}%
{\color[rgb]{0,0,0}\put(8326,-5761){\vector( 1, 0){675}}
}%
{\color[rgb]{0,0,0}\put(6451,-5161){\vector( 1, 0){1050}}
}%
{\color[rgb]{0,0,0}\put(6451,-6361){\vector( 1, 0){1050}}
}%
\put(6301,-4936){\makebox(0,0)[b]{\smash{{\SetFigFont{12}{24.0}
{\rmdefault}{\mddefault}{\updefault}{\color[rgb]{0,0,0}$a$}%
}}}}
\put(7651,-4936){\makebox(0,0)[b]{\smash{{\SetFigFont{12}{24.0}
{\rmdefault}{\mddefault}{\updefault}{\color[rgb]{0,0,0}$b$}%
}}}}
\put(6301,-6136){\makebox(0,0)[b]{\smash{{\SetFigFont{12}{24.0}
{\rmdefault}{\mddefault}{\updefault}{\color[rgb]{0,0,0}$c$}%
}}}}
\put(7651,-6136){\makebox(0,0)[b]{\smash{{\SetFigFont{12}{24.0}
{\rmdefault}{\mddefault}{\updefault}{\color[rgb]{0,0,0}$d$}%
}}}}
\put(9151,-5536){\makebox(0,0)[b]{\smash{{\SetFigFont{12}{24.0}
{\rmdefault}{\mddefault}{\updefault}{\color[rgb]{0,0,0}$e$}%
}}}}
\end{picture}%
\nop{
a ----> b --
            +--> e
c ----> d --
}

The order is not total because it does not compare $\{a,d\}$ and $\{c,b\}$.
None implies the other; for example, $F \not\models ad \rightarrow cb$ because
$F \not\models ad \rightarrow c$. Still, both sets entail $e$. Graphically,
they form a sort of ``cross'' in the diagram, so that each set contains a
literal that remains on the left of the other set, while $e$ is on the right of
both.

The figure also suggests a property they possess: following the arrow from them
leads to a common set $\{b,d\}$ that is still within their union. In terms of
entailment, they have a common consequence that is part of their union and
still implies $e$.

That they have a common consequence is obvious: $e$. The point is that they
also have a common consequence that only contains their elements and still
entails $e$: the set $\{b,d\}$. The following lemma proves that this is not a
coincidence.

\begin{lemma}
\label{convergence-syntax}

If $F$ is a single-head formula such that
{} $F \models A \rightarrow x$ and $F \models B \rightarrow x$,
then there exists $C \subseteq A \cup B$ such that
{}   $F^x \models A \rightarrow C$,
{}   $F^x \models B \rightarrow C$ and
{}   $F \models C \rightarrow x$.

\end{lemma}

\proof The proof is by induction on the size of $F$. If $F$ is empty the
preconditions
{} $F \models A \rightarrow x$ and $F \models B \rightarrow x$
imply $x \in A$ and $x \in B$. The claim holds with $C=\{x\}$.

The induction case assumes
{} $F \models A \rightarrow x$ and $F \models B \rightarrow x$
and postulates that the lemma holds for every formula smaller than $F$. If $x
\in A$ or $x \in B$ the claim holds with $C=\{x\}$. The rest of the proof
assumes $x \not\in A$ and $x \not\in B$.

Applying Lemma~\ref{set-implies-set} to $F \models A \rightarrow x$ and $F
\models B \rightarrow x$ proves
{} $F^x \models A \rightarrow P$ and $F^x \models B \rightarrow P'$
with $P \rightarrow x, P' \rightarrow x \in F$. Since $F$ is single-head, $P'$
is equal to $P$. As a result, $P$ satisfies most of the requirements of this
lemma: $F^x \models A \rightarrow P$, $F^x \models B \rightarrow P$ and $F
\models P \rightarrow x$. It may not meet $P \subseteq A \cup B$.

The rest of the proof shows how to distill a set $C$ that meets all conditions
of the claim from $P$. Let $y$ be an element of $P \backslash (A \cup B)$.
Since it is in $P$, both
{} $F^x \models A \rightarrow y$ and $F^x \models B \rightarrow y$
hold. Since $F$ is single-head, $F^x$ is also single-head because it is a
subset of it. It is a proper subset because it does not contain $P \rightarrow
x$. The claim of the lemma applies by induction: some set $C_y$ satisfies
{} $C_y \subseteq A \cup B$, 
{} $F^{xy} \models A \rightarrow C_y$, 
{} $F^{xy} \models B \rightarrow C_y$, and
{} $F^{xy} \models C_y \rightarrow y$.
Since $F^{xy}$ is a subset of $F^x$, the latter three imply
{} $F^x \models A \rightarrow C_y$, 
{} $F^x \models B \rightarrow C_y$, and
{} $F^x \models C_y \rightarrow y$.

Let $D = P \backslash (A \cup B)$ and $C = \bigcup_{y \in D} C_y$. Combining
the above conditions for all $y \in D$ results in
{} $C \subseteq A \cup B$, 
{} $F^x \models A \rightarrow C$, 
{} $F^x \models B \rightarrow C$, and
{} $F^x \models C \rightarrow D$.
The first condition
{} $C \subseteq A \cup B$
implies
{} $P \backslash D \cup C \subseteq A \cup B$
since $D$ comprises all elements of $P$ that are not in $A \cup B$. The second
condition
{} $F^x \models A \rightarrow C$, 
combines with $F^x \models A \rightarrow P$ to produce
{} $F^x \models A \rightarrow (P \backslash D \cup C)$.
By symmetry, the third condition produces
{} $F^x \models B \rightarrow (P \backslash D \cup C)$.
The fourth condition
{} $F^x \models C \rightarrow D$
implies
{} $F \models (P \backslash D \cup C) \rightarrow (P \backslash D \cup D)$,
which can be rewritten as
{} $F \models (P \backslash D \cup C) \rightarrow P$.
Combined with $F \models P \rightarrow x$, it gives
{} $F \models (P \backslash D \cup C) \rightarrow x$.
The claim is proved.~\qed

If both $A$ and $B$ entail $x$ in a single-head formula, both $A$ and $B$
entail the body of the only clause having $x$ as its head. What is not obvious
is that this convergence of entailments can be tracked back earlier than that,
to a set only comprising variables of $A$ and $B$.

The lemma holds on single-head formulae, but its intended usage is on formulae
that are equivalent to single-head ones. The construction $F^x$ does not
survive equivalence: $F^x$ is not equivalent to $F'^x$ even if $F'$ is
equivalent to $F$. Still, $F^x \models A \rightarrow C$ implies $F \models A
\rightarrow C$, which implies $F' \models A \rightarrow C$. The lemma with $F$
in place of $F^x$ holds on single-head equivalent formulae.

\begin{lemma}
\label{convergence}

If $F$ is a single-head equivalent formula such that
{} $F \models A \rightarrow x$ and $F \models B \rightarrow x$,
there exists $C \subseteq A \cup B$ such that
{}   $F \models A \rightarrow C$,
{}   $F \models B \rightarrow C$ and
{}   $F \models C \rightarrow x$.

\end{lemma}

\proof Since $F$ is single-head equivalent, a single-head formula $F'$ with $F
\equiv F'$ exists. Lemma~\ref{convergence-syntax} applies to $F'$. Everything
entailed by $F'^x$ is also entailed by $F'$, and also by $F$ by equivalence.
Replacing $F'^x$ by $F$ in the statement of Lemma~\ref{convergence-syntax}
results in the claim.~\qed

This lemma implies that $\leq_F$ is a downwards directed order if $F$ is
single-head equivalent: every pair of elements it compares has a lower bound.
However, it proves more than this: a lower bound is contained in the union of
the pair.

The lemma can be seen as yet another necessary condition to single-head
equivalence, but again is not sufficient. A counterexample is
{} $\{a \rightarrow b, b \rightarrow c, c \rightarrow b\}$,
which is not single-head equivalent but satisfies it. This formula is already
defined as Formula~\ref{inloop}, and proved not single-head equivalent. Yet, it
satisfies the claim of the lemma. This is proved for each variable that is the
head of a clause, $b$ and $c$. For example, $c$ is entailed by $\{a\}$, $\{b\}$
and $\{a,b\}$. The union of one of these three sets with itself is itself; the
union of one with another contains $b$, which implies $c$. The same holds for
$b$ by symmetry.

\subsection{The formula of minimal bodies}

The order between sets of literals suggests how to find a single-head
equivalent formula, if any: for every variable $x$, if more than one set of
literals imply it, only the minimal one is taken. For example, if
{} $F \models A \rightarrow x$,
{} $F \models B \rightarrow x$ and
{} $F \models C \rightarrow x$,
the three sets $A$, $B$ and $C$ are compared according to $\leq_F$. If $C$ is
less than $A$ and $B$, then $F \models A \rightarrow C$ and $F \models B
\rightarrow C$ hold. Maybe $C \rightarrow x$ is sufficient, since the other two
clauses $A \rightarrow x$ and $B \rightarrow x$ are consequences of it and $A
\rightarrow C$ and $B \rightarrow C$. Why ``maybe''? At this point the traps of
single-head equivalence should be clear: $A \rightarrow C$ may be a consequence
of $A \rightarrow x$. Or not.

The method works in the other way around. Just because $C$ is minimal does not
mean that it is the body of the clause with $x$ in the head in the single-head
equivalent formula. But if $C$ is that body, it is minimal. If only one minimal
element exists, the clause of the single-head form is found. If the formula is
inequivalent (Condition~\ref{condition-inequivalent}), uniqueness is
guaranteed. The minimal bodies make the single-head equivalent formula.

\[
MIN(F) = \left\{A \rightarrow x
~\left|~
\begin{array}{l}
F \models A \rightarrow x \mbox{ and } \\
\not\exists B ~.~
(
x \not\in B ,~
F \models B \rightarrow x \mbox{ and }
(B <_F A \mbox{ or } B \subset A \})
)
\end{array}
\right.
\right\}
\]

An inequivalent formula $F$ is single-head equivalent if and only if it is
equivalent to $MIN(F)$. A mechanism to establish single-head equivalence is to
build $MIN(F)$ and check whether it is single-head and equivalent to $F$. This
not only proves single-head equivalence, but also produces the single-head
equivalent formula: $MIN(F)$.

This procedure is correct but not complete. If it tells that the formula is
single-head equivalent, it is. Otherwise, it may still be. An example is the
following formula, which is single-head equivalent
as proved by the {\tt singlehead.py} program on the {\tt incomplete.py} test
file,
but $MIN(F)$ is equal to $F$, which is not single-head.

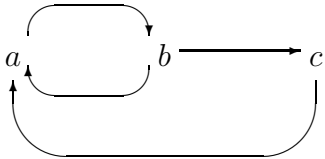
\begin{equation}
\label{minf-f}
F = \{a \rightarrow b, b \rightarrow a, b \rightarrow c, c \rightarrow a\}
\end{equation}

\setlength{\unitlength}{5000sp}%
\begingroup\makeatletter\ifx\SetFigFont\undefined%
\gdef\SetFigFont#1#2#3#4#5{%
  \reset@font\fontsize{#1}{#2pt}%
  \fontfamily{#3}\fontseries{#4}\fontshape{#5}%
  \selectfont}%
\fi\endgroup%
\begin{picture}(1530,768)(4786,-4270)
\thinlines
{\color[rgb]{0,0,0}\put(5476,-3661){\vector( 0,-1){0}}
\put(5176,-3661){\oval(600,300)[tr]}
\put(5176,-3661){\oval(600,300)[tl]}
}%
{\color[rgb]{0,0,0}\put(4876,-3811){\vector( 0, 1){0}}
\put(5176,-3811){\oval(600,300)[bl]}
\put(5176,-3811){\oval(600,300)[br]}
}%
{\color[rgb]{0,0,0}\put(4801,-3886){\vector( 0, 1){0}}
\put(5551,-3886){\oval(1500,750)[bl]}
\put(5551,-3886){\oval(1500,750)[br]}
}%
{\color[rgb]{0,0,0}\put(5626,-3736){\vector( 1, 0){600}}
}%
\put(4801,-3811){\makebox(0,0)[b]{\smash{{\SetFigFont{12}{24.0}
{\rmdefault}{\mddefault}{\updefault}{\color[rgb]{0,0,0}$a$}%
}}}}
\put(5551,-3811){\makebox(0,0)[b]{\smash{{\SetFigFont{12}{24.0}
{\rmdefault}{\mddefault}{\updefault}{\color[rgb]{0,0,0}$b$}%
}}}}
\put(6301,-3811){\makebox(0,0)[b]{\smash{{\SetFigFont{12}{24.0}
{\rmdefault}{\mddefault}{\updefault}{\color[rgb]{0,0,0}$c$}%
}}}}
\end{picture}%
\nop{
  --->
a       b ---> c 
^ <---         |
|              |
+--------------+
}

The following lemmas tell how $MIN(F)$ relates to $F$ and its single-head
equivalence. Their proofs sometimes involve sets that are strictly contained
one in the other, like $B \subset A$. A caveat is that $B \subset A$ implies $B
\leq_F A$ but not $B <_F A$. Even if $B$ is strictly contained into $A$, they
may still be equivalent: $B \equiv_F A$. As an example, $F \models \{a,b\}
\equiv \{b\}$ if $F$ is $\{b \rightarrow a\}$.

A first obvious property is that if $MIN(F)$ is single-head and equivalent to
$F$, then $F$ is single-head equivalent. This is almost the definition of
single-head equivalence, only restricted to $MIN(F)$. It makes a formal lemma
only for being referenced from the following proofs.

\begin{lemma}
\label{min-equivalent-single}

If $MIN(F)$ is single-head and equivalent to $F$,
then $F$ is single-head equivalent.

\end{lemma}

\proof By definition, $F$ is single-head equivalent if and only if it is
equivalent to a single-head formula. Such a formula is $MIN(F)$.~\qed

What is less obvious is that the converse holds in case of inequivalence
(Condition~\ref{condition-inequivalent}): if $F$ is inequivalent and
single-head equivalent, then $MIN(F)$ is single-head and equivalent to it.
Inequivalence makes $MIN(F)$ the single-head version of $F$, if any.

This claim can be broken in two: $F$ is equivalent to $MIN(F)$, and $MIN(F)$ is
single-head. Both claims require $F$ to be inequivalent and single-head
equivalent.

Some preliminary results are about single-head formulae.

\begin{lemma}
\label{no-less}

If $F$ is single-head and contains $A \rightarrow x$, it does not entail any
non-tautologic clause $B \rightarrow x$ with $B <_F A$.

\end{lemma}

\proof By contradiction, $F \models B \rightarrow x$ is assumed for some $B <_F
A$ with $x \not\in B$. The comparison $B <_F A$ is defined as $F \models A
\rightarrow B$ and $F \not\models B \rightarrow A$. By
Lemma~\ref{set-implies-set}, $F \models B \rightarrow x$ and $x \not\in B$
imply the existence of a clause $C \rightarrow x \in F$ such that $F \models B
\rightarrow C$. If $C$ were the same as $A$, it would contradict $F \not\models
B \rightarrow A$. As a result, $C$ is different from $A$. Therefore, $F$
contains two different clauses $A \rightarrow x$ and $C \rightarrow x$ with the
same head, contradicting the assumption that it is single-head.~\qed

Single-head formulae are always equivalent to their formulae of minimal bodies.

\begin{lemma}
\label{single-min-equivalent}

If $F$ is single-head, then $F \equiv MIN(F)$.

\end{lemma}

\proof Since $MIN(F)$ only contains clauses entailed by $F$, it is entailed by
$F$. The claim requires proving the converse: if $F$ is single-head, it is
entailed by $MIN(F)$.

This is the same as showing $MIN(F) \models A \rightarrow x$ for every $A
\rightarrow x \in F$. It is proved by contradiction, assuming $MIN(F)
\not\models A \rightarrow x$ for some clause $A \rightarrow x \in F$.

Since $A \rightarrow x$ is not entailed by $MIN(F)$, it is not in $MIN(F)$
either. Since it is entailed by $F$, the definition of $MIN(F)$ implies the
existence of a non-tautologic clause $B \rightarrow x$ such that $F \models B
\rightarrow x$ and either $B <_F A$ or $B \subset A$. These two possibilities
can be split differently: the first is $B <_F A$, the second is $B \not<_F A$
and $B \subset A$. The first makes $F \models B \rightarrow x$ and $x \not\in
B$ contradict Lemma~\ref{no-less}.

The second case is $B \not<_F A$ and $B \subset A$. The containment $B \subset
A$ makes $A \rightarrow B$ a tautology. The consequence $F \models A
\rightarrow B$ defines $B \leq_F A$. With $B \not<_F A$, it implies $B \equiv_F
A$. Let $C$ be a minimal subset of $A$ that is equivalent to it.
Such a minimal subset exists because the comparison is by the subset ordering.
If $C \rightarrow x \in MIN(F)$ then $MIN(F) \models C \rightarrow x$, which
implies $MIN(F) \models A \rightarrow x$ since $C \subset A$. A consequence of
$C \rightarrow x \not\in MIN(F)$ is $F \models D \rightarrow x$ with $D <_F C$
or $D \not<_F C$ and $D \subset C$ for some non-tautologic clause $D
\rightarrow x$.

If $D <_F C$ then $D <_F A$ because of $C \equiv_F A$; this makes $F \models D
\rightarrow x$ and $x \not\in D$ contradict Lemma~\ref{no-less}. The other case
is $D \not<_F C$ and $D \subset C$. The second condition is the start of the
chain of consequences $F \models C \rightarrow D$, $F \models C \rightarrow D$
and $D \leq_F C$. With $D \not<_F C$, the latter implies $D \equiv_F C$. This
makes $D \subset C$ contradict the minimality of $C$ among the subsets of $A$
that are equivalent to it.~\qed

The converse of this lemma requires $F$ to be inequivalent
Condition~\ref{condition-inequivalent}.

\begin{lemma}
\label{single-min-single}

If $F$ is single-head and inequivalent
(Condition~\ref{condition-inequivalent})
then $MIN(F)$ is single-head.

\end{lemma}

\proof The proof is by contradiction: $A \rightarrow x$ and $B \rightarrow x$
are assumed to both belong to $MIN(F)$, with $A \not= B$; this condition is
shown to contradict the assumptions.

By definition, all clauses of $MIN(F)$ are entailed by $F$, including $A
\rightarrow x$ and $B \rightarrow x$. By Lemma~\ref{convergence}, $F$ also
entails a clause $C \rightarrow x$ with $C \subseteq A \cup B$ such that both
$F \models A \rightarrow C$ and $F \models B \rightarrow C$ hold. These two
entailments define $C \leq_F A$ and $C \leq_F B$.

Since $F \models C \rightarrow x$, if either $C <_F A$ or $C \subset A$ were
true, then $A \rightarrow x \not\in MIN(F)$. As a result, both $C <_F A$ and $C
\subset A$ are false. By definition, $C <_F A$ is $C \leq_F A$ and $A
\not\leq_F C$; since $C <_F A$ is false, either $C \leq_F A$ is false or $A
\leq_F C$ is true. But the first is true. Therefore, the second is true: $A
\leq_F C$. With $C \leq_F A$, it proves $A \equiv_F C$. By symmetry, $B
\equiv_F C$.

This proves $A \equiv_F B$. Condition~\ref{condition-inequivalent} implies $A
\equiv_F A \cap B$. This equivalence and $F \models A \rightarrow x$ imply $F
\models A \cap B \rightarrow x$.

Since $A \cap B \subseteq A$, two cases are possible: either this containment
is strict or it is an equality. The first case, $A \cap B \subset A$,
contradicts the assumption $A \rightarrow x \in MIN(F)$ because $F \models A
\cap B \rightarrow x$. In the second case, $A \cap B = A$, it holds $A
\subseteq B$; since $A$ and $B$ are different, this containment is strict: $A
\subset B$; this contradicts the assumption $B \rightarrow x \in MIN(F)$
because $F \models A \rightarrow x$.~\qed

Combining the latter two lemmas tells that if $F$ is inequivalent, checking
whether $MIN(F)$ is single-head and equivalent to $F$ is a way to verify the
single-head equivalence of $F$.

\begin{lemma}
\label{inequivalent-min}

If $F$ is inequivalent, it is single-head equivalent if and only if $MIN(F)$ is
single-head and equivalent to $F$.

\end{lemma}

\proof If $MIN(F)$ is single-head and equivalent to $F$ then $F$ is single-head
equivalent by Lemma~\ref{min-equivalent-single}. In the other direction, if $F$
is single-head equivalent then it is equivalent to a single-head formula $F'$
by definition. By Lemma~\ref{single-min-equivalent}, $MIN(F)$ is equivalent to
$F'$, and to $F$ by transitivity. Since inequivalence is a semantical property,
$F'$ is also inequivalent. By Lemma~\ref{single-min-single}, $MIN(F')$ is
single-head. Since the formula of minimal bodies is defined semantically,
$MIN(F')$ is the same as $MIN(F)$ and is therefore single-head.~\qed

What happens if $F$ is not inequivalent? The previous counterexample
{} $F = \{a \rightarrow b, b \rightarrow a, b \rightarrow c, c \rightarrow a\}$
shows that the lemma does not extend: while $F$ is equivalent to the
single-head formula $\{a \rightarrow b, b \rightarrow c, c \rightarrow a\}$,
the formula $MIN(F)$ is not single-head. Indeed, $\{a\} \equiv_F \{b\}$, which
implies that both $a \rightarrow c$ and $b \rightarrow c$ are in $MIN(F)$
because they are entailed by $F$ and no set is less than $\{a\}$ or $\{b\}$
according to $\leq_F$.

A consequence of this lemma is that a formula used in a previous counterexample
is not single-head equivalent:
{} $F =
{}  \{ab \rightarrow x, bx \rightarrow c, ac \rightarrow d, d \rightarrow x\}$.
The proof was delayed to this point, where Lemma~\ref{inequivalent-min} makes
it easy.

\begin{lemma}
\label{ab-nonsingle}

Formula~\ref{cyclic-with-ab} is not single-head equivalent.

\end{lemma}

\proof Lemma~\ref{inequivalent-cyclic} proved Formula~\ref{cyclic-with-ab}
inequivalent. Lemma~\ref{inequivalent-min} proves that an inequivalent formula
$F$ is single-head equivalent if and only if $MIN(F)$ is single-head and
equivalent to it. The claim is proved by showing that the model $M = \{a,b\}$
that sets $a$ and $b$ to true and all other variables to false satisfies
$MIN(F)$ but not $F$.

It does not satisfy $F$ because it falsifies $ab \rightarrow x \in F$.

That $M$ satisfies $MIN(F)$ is proved by contradiction, by assuming it does
not. This means that it falsifies a clause of $MIN(F)$. Since $M$ assigns true
to $a$ and $b$ and false to the other variables, this clause may only have a
subset of $\{a,b\}$ in its body and a variable among $c$, $d$ and $x$ in the
head. The body is not empty because $MIN(F)$ only contains clauses entailed by
$F$, which does not entail any clause with an empty body. The body does not
comprise $a$ only because otherwise $MIN(F)$ contains a clause $a \rightarrow
h$ with $h \in \{c,d,x\}$ while $F$ entails no such clause. For the same
reason, the body does not comprise $b$ only.

Therefore, the clause of $MIN(F)$ is $ab \rightarrow h$ with $h \in \{c,d,x\}$.
For each of the three possible heads, a set $D$ that satisfies $F \models D
\rightarrow h$, $h \not\in D$ and $D <_F \{a,b\}$ is shown; this proves $ab
\rightarrow h \not\in MIN(F)$, contrary to assumption. Making $<_F$ explicit,
what is proved is $F \models D \rightarrow h$, $h \not\in D$, $F \models
\{a,b\} \rightarrow D$ and $F \not\models D \rightarrow \{a,b\}$.

\begin{description}

\item[$h=c$]; the required set $D$ is $\{b,x\}$; the clause $bx \rightarrow c$
is in $F$ and is therefore entailed by it; $F \models ab \rightarrow bx$ holds
because $\{a,b\}$ entails all variables; $F \not\models bx \rightarrow ab$
holds because no clause of $F$ has head $a$;

\item[$h=d$]; the required set $D$ is $\{a,c\}$; the clause $ac \rightarrow d$
is in $F$ and is therefore entailed by it; $F \models ab \rightarrow ac$ holds
because $\{a,b\}$ entails all variables; $F \not\models ac \rightarrow ab$
holds because no clause of $F$ has head $b$;

\item[$h=x$]; the required set $D$ is $\{d\}$; the clause $d \rightarrow x$
is in $F$ and is therefore entailed by it; $F \models ab \rightarrow d$ holds
because $\{a,b\}$ entails all variables; $F \not\models d \rightarrow ab$
holds because no clause of $F$ has head $a$.

\end{description}

In all three cases, a contradiction is reached from the assumption that $M$ is
not a model of $MIN(F)$. Therefore, it is. It is not a model of $F$, which
implies $F \not\equiv MIN(F)$. Since $F$ is inequivalent, it is not single-head
equivalent by Lemma~\ref{inequivalent-min}.~\qed

The proof is easy but not straighforward. While Formula~\ref{cyclic-with-ab}
is also proved not single-head equivalent by the {\tt reconstruct.py} program
on the {\tt conditiontwo.py} test file, the importance of the formula as a
counterexample calls for an explicit proof. The length of the proof for such
a simple formula shows the usefulness of the program.

\subsection{Computing a formula of minimal bodies}

\nojournal

If $MIN(F)$ is single-head and equivalent to $F$, forgetting can be computed in
polynomial time because it can be done on $MIN(F)$. Inequivalence is not
necessary. Forgetting on $MIN(F)$ instead of $F$ can be done even if $F$ is not
inequivalent, if it is equivalent to $MIN(F)$ and $MIN(F)$ is single-head.

Inequivalence enters into play when $MIN(F)$ is either not single-head or not
equivalent to $F$. Neither implies that $F$ is not single-head equivalent in
general. They only do if $F$ is also inequivalent. This information is
algorithmically useful because it ends the quest for a single-head equivalent
formula. Forgetting is done on $F$.

Checking inequivalence is not necessary. Checking whether $MIN(F)$ is
equivalent to $F$ is easy since both formulae are Horn. Computing $MIN(F)$ is
the problem. This formula may be large. An alternative is to find only a clause
$C \rightarrow x \in MIN(F)$ for each variable $x$, if any. Such a formula is
the same as $MIN(F)$ if $MIN(F)$ is single-head, which is the only case where
$MIN(F)$ is useful anyway.

This is the theme of this section: find a formula $SHMIN(F)$ that contains a
...

\endnojournal

The theme of this section is to find a formula $SHMIN(F)$ that contains a
clause $C \rightarrow x \in MIN(F)$ for each $x$, if any. Such a formula is
single-head by construction, and equivalent to $MIN(F)$ if $MIN(F)$ is
single-head. If $MIN(F)$ is also equivalent to $F$, then $SHMIN(F)$ is a
single-head version of $F$.

The starting point is a clause $A \rightarrow x \in F$ for each variable $x$.
If no other clause $B \rightarrow x$ entailed by $F$ is such that either $B <_F
A$ or $B \subset A$, then $A \rightarrow x$ is in $MIN(F)$; it is a valid
choice for the clause of $x$ in $SHMIN(F)$. Otherwise, $B \rightarrow x$ is
entailed by $F$ where $B$ is either less than or contained in $A$. The same
argument applies to $B \rightarrow x$ in place of $A \rightarrow x$. Other
clauses $B' \rightarrow x$ with the same properties are irrelevant because the
goal is to find a single minimal body for the head $x$, not all of them.

A subset $B \subset A$ can be found by looping over all variables $a \in A$ and
testing whether $F$ implies $A \backslash \{a\} \rightarrow x$. If so, $A
\rightarrow x$ is replaced by $A \backslash \{a\} \rightarrow x$ and the search
continues from there. The procedure stops when $A$ is subset-minimal: no $B
\subset A$ satisfies $F \models B \rightarrow x$.

This state is insufficient for $A \rightarrow x \in MIN(F)$ because $B <_F A$
and $F \models B \rightarrow x$ may still hold for some set $B$ that is not a
subset of $A$. The problem is that exponentially many sets of variables
$B${\plural} are to be checked. The set $BCN(A,F)$ is used to reduce this
number.

\[
BCN(A,F) = \{x \mid F \models A \rightarrow x\}
\]

This is the set of variables entailed by $A$ according to $F$~\cite{libe-20-a}.
It is the base of a condition that ensures that a set less than another exists.

\begin{lemma}
\label{less-minus-one-sufficient}

If
{} $F \models BCN(A,F) \backslash \{a,x\} \rightarrow x$ and
{} $F \not\models BCN(A,F) \backslash \{a,x\} \rightarrow a$
with $a \in A$, then $B = BCN(A,F) \backslash \{a,x\}$ satisfies
{} $F \models B \rightarrow x$ and $B <_F A$.

\end{lemma}

\proof The claim $F \models B \rightarrow x$ coincides with the assumption $F
\models BCN(A,F) \backslash \{a,x\} \rightarrow x$. The condition $B <_F A$ is
defined as $F \models A \rightarrow B$ and $F \not\models B \rightarrow A$. The
first holds because of $B \subset BCN(A,F)$, which means that $F \models A
\rightarrow b$ holds for every $b \in B$; the second holds because the
assumption
{} $F \not\models BCN(A,F) \backslash \{a,x\} \rightarrow a$
is the same as $F \not\models B \rightarrow a$, where $a$ is an element of
$A$.~\qed

This lemma gives a method for finding sets $B$ such that $B <_F A$ and $F
\models B \rightarrow x$. Similar to the loop over the sets $A \backslash
\{a\}$, it allows looping over all elements $a$ of $A$ instead of all sets of
variables. If the checks
{} $F \models BCN(A,F) \backslash \{a,x\} \rightarrow x$ and
{} $F \not\models BCN(A,F) \backslash \{a,x\} \rightarrow a$
succeed, $BCN(A,F) \backslash \{a,x\}$ replaces $A$ because it is less than $A$
according to $<_F$. The algorithm $SHMIN(F)$ uses this kind of loop. It takes a
definite Horn formula as input and produce one as output.

\

\begingroup

\renewcommand{\labelenumii}{\arabic{enumi}.\arabic{enumii}}
\renewcommand{\theenumii}{.\arabic{enumii}}

\begin{algorithm}
\label{shmin-algorithm}

\noindent $SHMIN(formula ~ F)$

\begin{enumerate}

\item $R = \emptyset$

\item for each $A \rightarrow x \in F$
\label{shmin-main-loop}

\begin{enumerate}

\item if $R$ contains a clause $B \rightarrow x$
\newline
then continue
\label{shmin-already}

\item $G = \true$

\item while $G$
\label{shmin-less-loop}

\begin{itemize}

\item $G = \false$

\item for each $a \in A$

\begin{itemize}

\item $B = BCN(A,F) \backslash \{a,x\}$

\item if $F \models B \rightarrow x$ and $F \not\models B \rightarrow a$ 
\newline
then $A = B$, $G = \true$, break

\end{itemize}

\end{itemize}

\item $G = \true$

\item while $G$
\label{shmin-contain-loop}

\begin{itemize}

\item $G = \false$

\item for each $a \in A$

\begin{itemize}

\item $B = A \backslash a$

\item if $F \models B \rightarrow x$
\newline
then $A = B$, $G = \true$, break

\end{itemize}

\end{itemize}

\item $R = R \cup \{A \rightarrow x\}$

\end{enumerate}

\item return $R$

\end{enumerate}

\end{algorithm}

\endgroup

\

The two loops in Step~\ref{shmin-less-loop} and Step~\ref{shmin-contain-loop}
are separated and in this order because of how minimality according to $<_F$
and $\subset$ interact. A subset of a body that is minimal according to $<_F$
is still minimal. Instead, a $\subset$-minimal body may still be greater than
another that is not $\subset$-minimal. First minimizing according to $<_F$ and
then to $\subset$ ensures that a second minimization according to $<_F$ is not
required.

Each iteration of the two loops checks a linear number of entailments, which
are polynomial-time because $F$ is Horn. The question is the number of
iterations: polynomial, exponential or infinite? Since the loops terminate if
$G$ is false and the only instruction that sets it to true is with $A = B$,
each iteration either is the last or replaces a set $A$ with another set $B$
that is strictly lower than it or strictly contained in it. Infinite chains are
impossible. Proving the iteration polynomial many is the purpose of the next
lemma.

\begin{lemma}
\label{shmin-poly}

The computation of $SHMIN(F)$ requires polynomial time.

\end{lemma}

\proof Each iteration of the two loops of Algorithm~\ref{shmin-algorithm} takes
polynomial time because it checks some entailments from Horn clauses, which
take polynomial time. The claim is a consequence of the number of iterations
being polynomial.

Both loops terminate if $G$ is false, and the only instructions that set it to
true are executed only after replacing $A$ with $B$. Apart from the last
iteration of each loop, all others replace $A$ with $B$.

The iterations of the first loop (Step~\ref{shmin-less-loop}) replace $A$ with
$B = BCN(A,F) \backslash \{a,x\}$ only if $B <_F A$, as proved by
Lemma~\ref{less-minus-one-sufficient}. A consequence of $B <_F A$ which is now
proved is $BCN(B,F) \subset BCN(A,F)$.

This claim is proved in two parts: first, $b \in BCN(B,F)$ implies $b \in
BCN(A,F)$; second, $BCN(B,F) = BCN(A,F)$ is a contradiction.

The definition of $B <_F A$ is $F \models A \rightarrow B$ and $F \not\models B
\rightarrow A$. The definition of $b \in BCN(B,F)$ is $F \models B \rightarrow
b$. Transitivity implies $F \models A \rightarrow b$, which defines $b \in
BNC(A,F)$. This proves $BNC(B,F) \subseteq BNC(A,F)$.

This containment is proved strict by implying a contradiction from its
converse: $BNC(B,F) = BCN(A,F)$. Since $A \rightarrow a$ is a tautology if $a
\in A$, it is valid and therefore entailed by $F$. This fact $F \models A
\rightarrow a$ defines $a \in BCN(A,F)$. This is the case for every $a \in A$,
which implies $A \subseteq BNC(A,F)$. Since $BNC(A,F)$ is equal to $BNC(B,F)$
by assumption, $A \subseteq BNC(B,F)$ follows. This is defined as $F \models B
\rightarrow a$ for every $a \in A$, or $F \models B \rightarrow A$.
Contradiction is reached since $F \not\models B \rightarrow A$ is part of the
definition of $B <_F A$.

The conclusion is that if $B$ replaces $A$ in an iteration of the first loop
(Step~\ref{shmin-less-loop}), then $BCN(B,F) \subset BCN(A,F)$. The set
$BCN(A,F)$ strictly decreases at every iteration. Since this set contains at
most all variables of $F$, the iterations cannot be more than the variables.
The number of iterations is linear.

A similar but simpler argument applies to the second loop
(Step~\ref{shmin-contain-loop}): the set $A$ decreases at every step since it
may only be replaced by $A \backslash \{a\}$ for some $a \in A$. Only a linear
number of iterations are possible.~\qed

Algorithm~\ref{shmin-algorithm} takes polynomial time. The next lemma is
proving that it always produces a single-head formula.

\begin{lemma}
\label{shmin-singlehead}

$SHMIN(F)$ is a single-head formula.

\end{lemma}

\proof The claim is proved by contradiction, assuming instead that
Algorithm~\ref{shmin-algorithm} returns two clauses with the same head. The
return value is $R$. This set is only changed at the end of each iteration of
the main loop at Step~\ref{shmin-main-loop}. The head $x$ of the second clause
that is added to $R$ did not change during the iteration of the main loop. It
is the same in Step~\ref{shmin-already}. Since this is the second clause of
head $x$ that is added to $R$, the check succeeds, cutting the loop short and
preventing the second clause to be added to $R$. This contradicts the
assumption that both clauses are eventually in $R$.~\qed

Algorithm~\ref{shmin-algorithm} takes polynomial time and produces a
single-head formula $SHMIN(F)$. If this set is equivalent to $F$, then $F$ is
single-head equivalent by definition.

The converse does not hold in general, it only holds if $F$ is inequivalent: if
$F$ is both single-head and inequivalent, then $F \equiv SHMIN(F)$. This shows
that $SHMIN(F)$ is not just a candidate for being the single-head form of $F$
if any, but a good candidate since it is the single-head form of $F$ when $F$
is inequivalent. This is proved in the next section.

When $F$ is not inequivalent, the algorithm may end with a body that is not
minimal according to $<_F$. Lemma~\ref{less-minus-one-sufficient} only gives a
sufficient condition, not a necessary and sufficient one. Even after exhausting
all sets $B = BCN(A,F) \backslash \{a,x\}$, the minimality of $A$ is not
guaranteed. Another set $B <_F A$ may still exist.

This drawback is not fatal to the intended usage of the final clause:
it is collected in a set $SHMIN(F)$ that is checked for equivalence with $F$.
If the system fails to find $B$ with $B <_F A$, the result will not be
equivalent to $F$, and $F$ is not replaced by $SHMIN(F)$ for computing
forgetting. Efficiency is harmed, not correctness.

Lemma~\ref{less-minus-one-sufficient} does not exclude that some sets less than
$A$ are missed when checking only the sets $BCN(A,F) \backslash \{a,x\}$ with
$a \in A$ instead of all sets of literals, but does not prove they may exist
either. The following lemma proves they may do. Additionally, it shows that
they may do even when no proper subset of $A$ entails $x$ and $F$ is
single-head.

\begin{lemma}
\label{less-minus-one-not-necessary}

There exists a single-head formula $F$, a set of variables $A$ and a variable
$x$ such that both
{} $F \models A \backslash \{a\} \rightarrow x$ and
{} $F \not\models BCN(A,F) \backslash \{a,x\} \rightarrow a$
are false for every $a \in A$, but there exists a set of variables $B$ such
that
{} $F \models B \rightarrow x$ and $B <_F A$.

\end{lemma}

\proof The formula $F$ and the set of variables $A${\plural} are as follows.

\begin{eqnarray*}
F &=& \{
ab \rightarrow d,
ad \rightarrow b,
bd \rightarrow a,
d \rightarrow x
\} \\
A &=& \{a,b\}
\end{eqnarray*}

Every single element of $BCN(A,F) \backslash \{x\} = \{a,b,d\}$ is entailed by
the other two, making these subsets equivalent to $A$. Yet, the two elements
$a$ and $b$ cannot be recovered once both removed: $\{d\}$ is strictly less
than $A$ while still satisfying $F \models d \rightarrow x$.

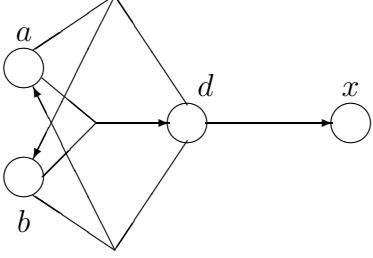
\setlength{\unitlength}{5000sp}%
\begingroup\makeatletter\ifx\SetFigFont\undefined%
\gdef\SetFigFont#1#2#3#4#5{%
  \reset@font\fontsize{#1}{#2pt}%
  \fontfamily{#3}\fontseries{#4}\fontshape{#5}%
  \selectfont}%
\fi\endgroup%
\begin{picture}(1838,1284)(5562,-4483)
{\color[rgb]{0,0,0}\thinlines
\put(5671,-3571){\circle{202}}
}%
{\color[rgb]{0,0,0}\put(5671,-4111){\circle{202}}
}%
{\color[rgb]{0,0,0}\put(6481,-3841){\circle{202}}
}%
{\color[rgb]{0,0,0}\put(7291,-3841){\circle{202}}
}%
{\color[rgb]{0,0,0}\put(5761,-3616){\line( 6,-5){270}}
\put(6031,-3841){\line(-1,-1){270}}
}%
{\color[rgb]{0,0,0}\put(6031,-3841){\vector( 1, 0){360}}
}%
{\color[rgb]{0,0,0}\put(6571,-3841){\vector( 1, 0){630}}
}%
{\color[rgb]{0,0,0}\put(6481,-3751){\line(-2, 3){360}}
\put(6121,-3211){\line(-3,-2){405}}
}%
{\color[rgb]{0,0,0}\put(6481,-3931){\line(-2,-3){360}}
\put(6121,-4471){\line(-3, 2){405}}
}%
{\color[rgb]{0,0,0}\put(6121,-4471){\vector(-1, 2){405}}
}%
{\color[rgb]{0,0,0}\put(6121,-3211){\vector(-1,-2){405}}
}%
\put(5671,-3436){\makebox(0,0)[b]{\smash{{\SetFigFont{12}{24.0}
{\rmdefault}{\mddefault}{\updefault}{\color[rgb]{0,0,0}$a$}%
}}}}
\put(6571,-3706){\makebox(0,0)[b]{\smash{{\SetFigFont{12}{24.0}
{\rmdefault}{\mddefault}{\updefault}{\color[rgb]{0,0,0}$d$}%
}}}}
\put(5671,-4381){\makebox(0,0)[b]{\smash{{\SetFigFont{12}{24.0}
{\rmdefault}{\mddefault}{\updefault}{\color[rgb]{0,0,0}$b$}%
}}}}
\put(7291,-3706){\makebox(0,0)[b]{\smash{{\SetFigFont{12}{24.0}
{\rmdefault}{\mddefault}{\updefault}{\color[rgb]{0,0,0}$x$}%
}}}}
\end{picture}%
\nop{
 +-----+---------+
 |     |         |
 |     V         |
 |  +- a ---+    |
 |  |       +--> d -----> x
 +--|- b ---+    |
    |  ^         |
    |  |         |
    +--+---------+
}

The claim is proved for $a$; it holds for $b$ by symmetry. The first
requirement is the falsity of
{} $F \models A \backslash \{a\} \rightarrow x$;
it holds because $A \backslash \{a\} = \{b\}$, and $b$ alone does not imply
$x$. The second is the falsity of
{} $F \not\models BCN(A,F) \backslash \{a,x\} \rightarrow a$,
which is the same as the truth of
{} $F \models BCN(A,F) \backslash \{a,x\} \rightarrow a$;
it holds because
{} $BCN(A,F) \backslash \{a,x\} = \{a,b,d,x\} \backslash \{a,x\} = \{b,d\}$;
this set implies $a$ thanks to $bd \rightarrow a \in F$.

The set $B$ of the statement of the lemma is $\{d\}$. It satisfies $F \models B
\rightarrow x$ because $B \rightarrow x$ is $d \rightarrow x$, which is in $F$.
The other requirement $B <_F A$ is defined as $F \models A \rightarrow B$ and
$F \not\models B \rightarrow A$. The first holds because $A \rightarrow B$ is
$ab \rightarrow d$, which is in $B$. The second holds because $B \rightarrow A$
is $d \rightarrow a$ and $d \rightarrow b$, and $F$ is consistent with the
model that assigns true to $d$ and $x$ and false to $a$ and $b$.~\qed

This lemma proves that no set $B = A \backslash \{a\}$ or $B = BCN(A,F)
\backslash \{x,a\}$ may replace $A$, yet a set $B$ strictly smaller than $A$
satisfies $F \models B \rightarrow x$. If Algorithm~\ref{shmin-algorithm}
starts from the clause $ab \rightarrow x$ when analyzing the formula $F \cup
\{ab \rightarrow x\}$, it outputs that clause because no strictly lower set or
proper subset is found by looping over a single variable in $A$. Since $F$ is
single-head and entails $ab \rightarrow x$, the formula $F \cup \{ab
\rightarrow x\}$ is single-head equivalent. Yet, $SHMIN(F)$ contains the wrong
clause $ab \rightarrow x$.

\subsection{Inequivalence and single-head equivalence}

Lemma~\ref{less-minus-one-not-necessary} shows that
{} $F \models B \rightarrow x$ and $B <_F A$
do not imply the existence of a variable $a \in A$ such that
{} $F \models BCN(A,F) \backslash \{a,x\} \rightarrow x$
and
{} $F \not\models BCN(A,F) \backslash \{a,x\} \rightarrow a$.
In short, the converse of Lemma~\ref{less-minus-one-sufficient} does not always
hold. Removing single variables from $BCN(A,F) \backslash \{x\}$ does not
always provide a body less than $A$ even if one exists.

It does when the formula is inequivalent: a lower body is always found this
way, if one exists. A minimal body eventually results.
Algorithm~\ref{shmin-algorithm} always finds a single-head equivalent formula
if one exists.

\begin{lemma}
\label{less-minus-one}

If $F$ is inequivalent (Condition~\ref{condition-inequivalent}) and entails $A
\rightarrow x$ with $x \not\in A$, it entails a clause $B \rightarrow x$ with
$x \not\in B$ and $B <_F A$ if and only if
{} $F \models BCN(A,F) \backslash \{a,x\} \rightarrow x$ and
{} $F \not\models BCN(A,F) \backslash \{a,x\} \rightarrow a$
both hold for some $a \in A$.

\end{lemma}

\proof Lemma~\ref{less-minus-one-sufficient} proves that
{} $F \models BCN(A,F) \backslash \{a,x\} \rightarrow x$ and
{} $F \not\models BCN(A,F) \backslash \{a,x\} \rightarrow a$
imply that $B = BCN(A,F) \backslash \{a,x\}$ satisfies $F \models B \rightarrow
x$ and $B <_F A$.

The rest of the proof is for the converse: $F \models B \rightarrow x$, $x
\not\in B$ and $B <_F A$ imply
{} $F \models BCN(A,F) \backslash \{a,x\} \rightarrow x$ and
{} $F \not\models BCN(A,F) \backslash \{a,x\} \rightarrow a$
for some $a \in A$ if $F$ is inequivalent.

Let $A'$ be the set of variables of $BCN(A,F)$ that are not entailed by $B$:

\[
{} A' = BCN(A,F) \backslash BCN(B,F) =
{}  \{a \in BCN(A,F) \mid F \not\models B \rightarrow a\}
\]

The first part
{} $F \models BCN(A,F) \backslash \{a,x\} \rightarrow x$
of the claim is proved for all elements $a \in A'$, the second
{} $F \not\models BCN(A,F) \backslash \{a,x\} \rightarrow a$
for at least an element $a \in A' \cap A$.

The assumption $B <_F A$ includes $F \models A \rightarrow B$, which implies $B
\backslash \{x\}$ thanks to the assumption $x \not\in B$. Since $B$ does not
entail any element of $A'$ by construction, it does not contain any. This makes
the containment further strengthen to $B \subseteq BCN(A,F) \backslash A'
\backslash \{x\}$, which is needed below. Since $F$ entails $B \rightarrow x$,
it also entails its superclause $BCN(A,F) \backslash \{a,x\} \rightarrow x$ for
every $a \in A'$.

The other part of the claim is proved by showing that
{} $F \models BCN(A,F) \backslash \{a,x\} \rightarrow a$
for all $a \in A'$ contradicts the assumption $B <_F A$. A consequence is
{} $F \not\models BCN(A,F) \backslash \{a,x\} \rightarrow a$
for at least a variable $a \in A'$; this variable is then proved to belong to
$A$.

By construction, $F$ entails $A \rightarrow BCN(A,F)$; this implies
{} $F \models A \rightarrow BCN(A,F) \backslash \{a,x\}$
{} for every variable $a$.
The converse implication
{} $F \models BCN(A,F) \backslash \{a,x\} \rightarrow A$
holds for all $a \in A'$ because $x$ does not belong to $A$ by assumption and
$BCN(A,F) \backslash \{a,x\}$ by assumption implies $a$, the only element of
$A$ it does not contain. Implication in both directions is equivalence: $F
\models BCN(A,F) \backslash \{a,x\} \equiv A$. This holds for every $a \in A'$:
all sets $BCN(A,F) \backslash \{a,x\}$ for $a \in A'$ are equivalent to $A$.
Therefore, they are also equivalent to each other. By
Condition~\ref{condition-inequivalent}, they are equivalent to their
intersection
{} $BCN(A,F) \backslash A' \backslash \{x\}$,
which is therefore equivalent to $A$.

By construction, $A'$ comprises all elements $BCN(A,F)$ that $F \cup B$ does
not imply; as a result, $F \cup B$ implies the others: $F \models B \rightarrow
BCN(A,F) \backslash A'$. An immediate consequence is $F \models B \rightarrow
BCN(A,F) \backslash A' \backslash \{x\}$. The converse implication is a
consequence of $B \subseteq BCN(A,F) \backslash A' \backslash \{x\}$, proved
above. This proves the equivalence
{} $F \models B \equiv BCN(A,F) \backslash A' \backslash \{x\}$.
Since $BCN(A,F) \backslash A' \backslash \{x\}$ is equivalent to $A$, so is
$B$. This contradicts the assumption $B <_F A$.

This contradiction disproves the assumption
{} $F \models BCN(A,F) \backslash \{a,x\} \rightarrow a$ for all $a \in A'$.
Therefore, it proves $F \not\models BCN(A,F) \backslash \{a,x\} \rightarrow a$
for at least an element $a \in A'$. If $a \not\in A$ then $A \subseteq BCN(A,F)
\backslash \{a,x\}$ since $x \not\in A$. This implies $F \not\models A
\rightarrow a$, which is a contradiction because the definition of $a \in A'$
includes $a \in BCN(A,F)$, which is $F \models A \rightarrow a$ by definition.
Therefore, $a \in A$.~\qed

This lemma proves that the mechanism of checking one variable at time when
searching for a lower body is correct in the inequivalent case.
Algorithm~\ref{shmin-algorithm} switches from a body to a lower body if any,
returning a body only when minimal.

\begin{lemma}
\label{min-shmin}

If $F$ is inequivalent then $SHMIN(F) \subseteq MIN(F)$.

\end{lemma}

\proof Let $A \rightarrow x$ be a clause in $SHMIN(F)$. It is proved to be in
$MIN(F)$. The claim requires $F \models A \rightarrow x$ and that $F \models B
\rightarrow x$ does not hold if $B <_F A$ or $B \subset A$.

The first part of the claim is an invariant in Algorithm~\ref{shmin-algorithm}:
the clause $A \rightarrow x$ is initially a clause of $F$, and $B$ replaces $A$
only if $F \models B \rightarrow x$.

The second part of the claim takes most of the proof.

Algorithm~\ref{shmin-algorithm} returns a clause $A \rightarrow x$ only when
its second loop at Step~\ref{shmin-contain-loop} ends. This is only the case
when $G$ remains false during an entire iteration, which happens only when $F
\models A \backslash \{a\} \rightarrow x$ holds for no variable $a \in A$. If
$F \models B \rightarrow x$ holds with $B \subset A$ then $A \backslash B$
contains at least a variable $b$ because containment is strict. Since $B$ does
not contain $b$, it is contained in $A \backslash \{b\}$. A consequence of $F
\models B \rightarrow x$ is $F \models A \backslash \{b\} \rightarrow x$, which
contradicts the assumption with $a=b$. This proves that if $A \rightarrow x$ is
in $SHMIN(F)$ then $F \models B \rightarrow x$ does not hold if $B \subset A$.

This is the first part of the definition of $A \rightarrow x \in MIN(F)$. The
second is that $F \models B \rightarrow x$ does not hold if $B <_F A$.

Let $A'$ be the value of $A$ at the end of the first loop of
Algorithm~\ref{shmin-algorithm} at Step~\ref{shmin-less-loop}. The loop ends
only when $G$ remains false for an entire iteration, which is only the case
when $BCN(A,F) \backslash \{a,x\} \models x$ and $BCN(A,F) \backslash \{a,x\}
\not\models a$ do not hold at the same time for any $a \in A$.
Lemma~\ref{less-minus-one} proves that $B <_F A$ and $F \models B \rightarrow
x$ with $x \not\in B$ implies the opposite of that. As a result, $B <_F A'$
holds for no non-tautologic clause $B \rightarrow x$ entailed by $F$.

This is not the claim yet, because the set $A'$ that is proved minimal is the
value of $A$ at the end of the first loop of the algorithm, not its final
value. Minimality extends to that thanks to the invariant $A \equiv_F A'$ of
the second loop of Algorithm~\ref{shmin-algorithm} at
Step~\ref{shmin-contain-loop}.

This invariant is proved by induction on the number of iterations. The claim
$A' \equiv_F A$ is proved at the beginning of the second loop of the algorithm
(base case) and is then assumed at the beginning of an iteration of the loop
and proved at the end (induction case).

The base case is the start of the second loop. Since $A'$ is the value of $A$
at the end of the first loop of Algorithm~\ref{shmin-algorithm}, it is the same
as the value of $A$ at the beginning of the second. The claim holds because
$A=A'$ implies $A \equiv_F A'$.

The induction case is about an arbitrary iteration of the loop. The inductive
assumption is $A \equiv_F A'$ when the iteration starts, the inductive claim is
the same at the end. The value of $A$ changes only if $F \models A \backslash
\{a\} \rightarrow x$ with $a \in A$. Since $A \rightarrow A \backslash \{a\}$
is a tautology, $F$ implies it. This defines $A \backslash \{a\} \leq_F A$. The
inductive assumption $A \equiv_F A'$ implies $A \backslash \{a\} \leq_F A'$.
Less than or equal to are two possibilities: less than, or equal to. In this
case, $A \backslash \{a\} \leq_F A'$ is either $A \backslash \{a\} <_F A'$ or
$A \backslash \{a\} \equiv_F A'$. The first is not possible because it
contradicts the previously proved property that $B <_F A'$ holds for no
non-tautologic clause $B \rightarrow x$ entailed by $F$. The only actual
possibility is the second: $A \backslash \{a\} \equiv_F A'$. Since $A
\backslash \{a\}$ replaces $A$, the inductive claim follows: the next value of
$A$ is equivalent to $A'$.

Since $B <_F A'$ holds for no non-tautologic clause $B \rightarrow x$ entailed
by $F$ and $A'$ is equivalent to all following values of $A$, this property
holds for the final value of $A$. This is the second part of the definition of
$A \rightarrow x$ being in $MIN(F)$.~\qed

Having proved that $SHMIN(F)$ only contains clauses of $MIN(F)$, remains to
prove the converse: it contains all of them. This claim requires an additional
assumption: $F$ is not only inequivalent but also single-head equivalent. The
second is necessary because $MIN(F)$ is not single-head otherwise, while
$SHMIN(F)$ always is.

Intuitively, $SHMIN(F)$ starts from a clause of $F$ and produces a clause
entailed by it with a minimal body. Instead, $MIN(F)$ contains all entailed
clauses with a minimal body. The difference is one versus all, which disappears
in the single-head equivalent case. With a caveat: $SHMIN(F)$ starts with a
clause in $F$ while $MIN(F)$ is purely semantical. This gap is filled by the
next lemma.

\begin{lemma}
\label{minf-head}

If $MIN(F)$ contains a clause with head $x$ then $F$ contains a clause with
head $x$.

\end{lemma}

\proof The proof is by contradiction: let $A \rightarrow x$ be a clause of
$MIN(F)$ such that $F$ contains no clause with head $x$. Since $MIN(F)$ only
contains clauses entailed by $F$, this particular clause $A \rightarrow x$ is
entailed by $F$.

Let $M$ be the model that sets all variables to true but $x$. This model
satisfies every clause $B \rightarrow y$ of $F$ because $y$ is by assumption
different from $x$, and is therefore assigned true by $M$. The clause $A
\rightarrow x$ is instead falsified by $M$ because all variables in $A$ are
different from $x$ and therefore assigned to true while $x$ is assigned to
false.~\qed

This lemma shows that all clauses in $MIN(F)$ have a head that is also a head
in $F$, and therefore a head in $SHMIN(F)$. In the single-head equivalent
case, this was the only missing bit holding $SHMIN(F)$ away from $MIN(F)$.

\begin{lemma}
\label{shmin-min-singlehead}

If $F$ is inequivalent and single-head equivalent, then $SHMIN(F) = MIN(F)$.

\end{lemma}

\proof Lemma~\ref{min-shmin} proves $SHMIN(F) \subseteq MIN(F)$ since $F$ is
inequivalent. The claim follows from $MIN(F) \subseteq SHMIN(F)$, which is
proved if $F$ is single-head equivalent.

Lemma~\ref{minf-head} proves that $A \rightarrow x \in MIN(F)$ implies that $F$
contains some clauses with the same head. In other words, $F$ contains one or
more clauses of head $x$. The iterations of the main loop of
Algorithm~\ref{shmin-algorithm} at Step~\ref{shmin-main-loop} are performed on
all clauses of $F$, including them. Since $x$ is never changed during an
iteration, it is the head of the clause added to $R$ if any. The first
iteration where the clause has head $x$ is such that: $R$ does not contain any
clause of head $x$ because this is the first; the iteration is therefore not
cut short at Step~\ref{shmin-already} and a clause is added to $R$, a clause of
head $x$. Let $B \rightarrow x \in SHMIN(F)$ be such a clause. Since $SHMIN(F)
\subseteq MIN(F)$ by Lemma~\ref{min-shmin}, $B \rightarrow x$ is in $MIN(F)$.
By Lemma~\ref{single-min-single}, $MIN(F)$ is single-head. Since it contains $A
\rightarrow x$, this is the only clause of head $x$ it contains. Therefore, $B$
is equal to $A$, proving $A \rightarrow x \in SHMIN(F)$.

This being the case for every clause of $MIN(F)$, it proves $MIN(F) \subseteq
SHMIN(F)$.~\qed

The final destination of this string of lemmas is the equality of $MIN(F)$ and
$SHMIN(F)$ if $F$ is both inequivalent and single-head equivalent:
Algorithm~\ref{shmin-algorithm} calculates $MIN(F)$ in this case. This gives a
way for checking single-head equivalence in the inequivalent case.

\begin{theorem}
\label{shmin-inequivalent-singlehead}

An inequivalent formula $F$ is single-head equivalent if and only if
$F \equiv SHMIN(F)$.

\end{theorem}

\proof The if direction of the lemma does not require inequivalence: $F \equiv
SHMIN(F)$ implies that $F$ is single-head equivalent.
Lemma~\ref{shmin-singlehead} proves that $SHMIN(F)$ is single-head. As a
result, if $F \equiv SHMIN(F)$ then $F$ is equivalent to the single-head
formula $SHMIN(F)$.

The only if direction of the lemma assumes $F$ inequivalent and single-head
equivalent and proves $F \equiv SHMIN(F)$. When $F$ is both inequivalent and
single-head equivalent, Lemma~\ref{shmin-min-singlehead} proves $SHMIN(F) =
MIN(F)$ and Lemma~\ref{inequivalent-min} proves $F \equiv MIN(F)$. Together,
these two facts imply $F \equiv SHMIN(F)$.~\qed

A consequence of this lemma is that single-head equivalence can be computed in
polynomial time on inequivalent formulae. The polynomial algorithm not only
tells whether a single-head equivalent formula exists but produces it. That
formula can be then fed to the forgetting algorithm, which runs in polynomial
time because the formula is single-head.

\begin{theorem}
\label{inequivalent-polynomial}

Computing a single-head formula equivalent to a given inequivalent formula if
any is polynomial-time.

\end{theorem}

\proof The polynomial algorithm computes $SHMIN(F)$ and checks $F \equiv
SHMIN(F)$. This establishes the single-head equivalence of $F$ by
Theorem~\ref{shmin-inequivalent-singlehead}. If this check succeeds, then
$SHMIN(F)$ is a single-head formula equivalent to $F$; that is single-head is
proved by Lemma~\ref{shmin-singlehead}. Generating it takes polynomial time
thanks to Lemma~\ref{shmin-poly}; checking equivalence also takes polynomial
time because both $F$ and $SHMIN(F)$ are Horn formulae.~\qed

Even if the formula is not inequivalent, producing $SHMIN(F)$ takes polynomial
time. Inequality guarantees its equality with $MIN(F)$ if $F$ is single-head
equivalent.

This theorem provides a way to forget variables from a formula that is both
inequivalent and single-head equivalent: first compute the single-head formula
that is equivalent to the given one, which can be done in polynomial time since
the formula is also inequivalent, and then forget the variables from that
formula.

\begin{corollary}
\label{inequivalent-polynomial-forget}

Forgetting variables from inequivalent and single-head equivalent formulae can
be computed in polynomial time.

\end{corollary}

\subsection{The real consequences}

The previous section shows how to calculate $SHMIN(F)$ by a sequence of steps
that check the following conditions for all $a \in A$:

\begin{itemize}

\item $F \models BCN(A,F) \backslash \{a,x\} \rightarrow x$,

\item $F \not\models BCN(A,F) \backslash \{a,x\} \rightarrow a$,

\item $F \models A \backslash \{a\} \rightarrow x$.

\end{itemize}

These conditions can be tested exploiting the real consequences, the variables
that are inferred from $A$ with at least a derivation step. This excludes the
variables of $A$ that are entailed just because they are in $A$, not because of
some clauses of $F$.

\[
RCN(A,F) = \{a \mid F \models BCN(A,F) \backslash \{a\} \rightarrow a\}
\]

While $BCN(A,F)$ contains all variables that $A$ implies, $RCN(A,F)$ only
contains its real consequences, those implied thanks to at least a clause of
$F$. This is formally proved by the following lemma.

\begin{lemma}
\label{use-clause}

A variable $x$ is in $RCN(A,F)$ if and only if $F$ contains a clause $B
\rightarrow x$ such that $F \models A \rightarrow B$.

\end{lemma}

\proof The definition of $x \in RCN(A,F)$ is $F \models BCN(A,F) \backslash
\{x\} \rightarrow x$.

Since $x \not\in BCN(A,F) \backslash \{x\}$, Lemma~\ref{set-implies-set}
implies the existence of a clause $B \rightarrow x \in F$ such that
{} $F \models BCN(A,F) \backslash \{x\} \rightarrow B$.
Since $F \models A \rightarrow BCN(A,F)$, transitivity and monotonicity tell $F
\models A \rightarrow B$, the required conclusion.

The other direction assumes the existence of a clause $B \rightarrow x \in F$
with $F \models A \rightarrow B$. The latter implies $B \subseteq BCN(A,F)$.
Since no clause in $F$ is tautologic by assumption, $x$ is not in $B$.
Therefore, $B \subseteq BCN(A,F) \backslash \{x\}$. By monotonicity, $F \models
B \rightarrow x$ implies $F \models BCN(A,F) \backslash \{x\} \rightarrow
x$.~\qed

This lemma clarifies the difference between $RCN(A,F)$ and $BCN(A,F)$: both
require each of their variables $x$ to be entailed from $F \cup A$, but the
first also imposes this entailment to result from a clause $B \rightarrow x$,
the second do not. Only when $x \in A$ this difference matters.

The next section shows an algorithm for $RCN(B,F)$. The rest of this one shows
how $RCN(B,F)$ is used. The first way is to calculate $BCN(B,F)$.

\begin{lemma}
\label{b-rcn}

For every formula $F$ and set of variables $A$, it holds
{} $BCN(A,F) = A \cup RCN(A,F)$.

\end{lemma}

\proof The claim is that $x \in BCN(A,F)$ is the same as either $x \in A$ or $x
\in RCN(A,F)$.

The condition $x \in BCN(A,F)$ is defined as $F \models A \rightarrow x$.
Lemma~\ref{set-implies-set} tells it equivalent to $F$ containing a clause $B
\rightarrow x$ such that $F \models A \rightarrow B$ if $x$ is not in $A$.
Reformulated, ``something exists if a condition is false'' is the same as
``either the condition is true or something exists''. In the present case,
either $x \in A$ or $F \models A \rightarrow B$ for some $B \rightarrow x \in
F$. As proved by Lemma~\ref{use-clause}, the second possibility is equivalent to
$x \in RCN(A,F)$. Overall, either $x \in A$ or $x \in RCN(A,F)$.~\qed

The three checks required to determine $SHMIN(F)$ can be expressed in terms of
$RCN()$.

The check $F \models A \backslash \{a\} \rightarrow x$ is equivalent to $x \in
BCN(A \backslash \{a\},F)$, which Lemma~\ref{b-rcn} proved the same as
{} $x \in A \backslash \{a\} \cup RCN(A \backslash \{a\},F)$.

The check
{} $F \models BCN(A,F) \backslash \{a,x\} \rightarrow x$
is equivalent to $x \in B \cup RCN(B,F)$ with $B = BCN(A,F) \backslash
\{a,x\}$ for the same reason.

The check
{} $F \not\models BCN(A,F) \backslash \{a,x\} \rightarrow a$
is only needed if the previous check succeeds, since
Lemma~\ref{less-minus-one-sufficient} requires both. In so, the following lemma
recasts it in terms of $RCN(A,F)$.

\begin{lemma}
\label{third-rcn}

If
{} $F \models BCN(A,F) \backslash \{a,x\} \rightarrow x$
holds with $a \in A$, then
{} $F \not\models BCN(A,F) \backslash \{a,x\} \rightarrow a$
is equivalent to 
$a \in A \backslash RCN(A,F)$.

\end{lemma}

\proof A consequence of
{} $F \models BCN(A,F) \backslash \{a,x\} \rightarrow x$
is
{} $F \models BCN(A,F) \backslash \{a,x\} \equiv BCN(A,F) \backslash \{a\}$
because the entailing set contains all elements of the entailed set but $x$ and
entails $x$ by assumption. The condition
{} $F \not\models BCN(A,F) \backslash \{a,x\} \rightarrow a$
is therefore the same as
{} $F \not\models BCN(A,F) \backslash \{a\} \rightarrow a$,
which is the exact opposite of the definition of $a \in RCN(A,F)$. Since $a$ is
by assumption an element of $A$, the condition $a \not\in RCN(A,F)$ is the same
as $a \in A \backslash RCN(A,F)$.~\qed

Another use of $RCN(A,F)$ is to prove some sets to be strictly greater than
certain subsets of them.


\begin{lemma}
\label{rcn-less}

If $a \in A \backslash RCN(A,F)$ then $A \backslash \{a\} <_F A$.

\end{lemma}

\proof The claim $A \backslash \{a\} <_F A$ is
{} $F \models A \rightarrow A \backslash \{a\}$
and
{} $F \not\models A \backslash \{a\} \rightarrow A$.
The first condition holds because $A \backslash \{a\}$ is a subset of $A$.

The second condition
{} $F \not\models A \backslash \{a\} \rightarrow A$
is proved as follows. Since $a$ is not in
{} $RCN(A,F) = \{a \mid F \models BCN(A,F) \backslash \{a\} \rightarrow a\}$,
the entailment
{} $F \models BCN(A,F) \backslash \{a\} \rightarrow a$
does not hold. Since $A \subseteq BCN(A,F)$, by monotonicity of entailment
{} $F \models A \backslash \{a\} \rightarrow a$
does not hold either. This implies
{} $F \not\models A \backslash \{a\} \rightarrow A$
since $A \backslash \{a\}$ does not imply $a$, which is in $A$.~\qed

\subsection{Finding the real consequences}

The following algorithm calculates the real consequence $RCN(B,F)$ of $B$
according to $F$. It employs unit propagation~\cite{craw-auto-93} but keeps the
given variables $B$ separated from the generated ones $H$.

\

\begin{algorithm}
\label{algorithm-rcnucl}

\noindent $rcnucl(variables ~ B, formula ~ F)$:

\begin{enumerate}

\item $H = \emptyset$

\item $U = \emptyset$

\item while $H$ changes:

\begin{enumerate}

\item[~] for every $B' \rightarrow x \in F$ such that $B' \subseteq B \cup H$:

\begin{enumerate}

\item $H = H \cup \{x\}$

\item $U = U \cup \{B' \rightarrow x\}$

\end{enumerate}

\end{enumerate}

\item return $H,U$

\end{enumerate}

\end{algorithm}

\

The first return value is proved to be $RCN(B,F)$ by the following lemma. The
algorithm also returns the clauses it used, which prove useful later.

\begin{lemma}
\label{rcn-first}

The first return value of Algorithm~\ref{algorithm-rcnucl} is $RCN(B,F)$.

\end{lemma}

\proof The proof relies on Lemma~\ref{use-clause}: $x \in RCN(B,F)$ is
equivalent to $F \models B \rightarrow B'$ for a clause $B' \rightarrow x \in
F$; the entailment is equivalent to $B' \subseteq BCN(B,F)$. Since $x$ enters
$H$ if and when $B' \subseteq B \cup H$ holds for a clause $B' \rightarrow x
\in F$, the claim follows from $B \cup H$ always being a subset of $BCN(B,F)$
and eventually becoming equal to it. The first fact implies that $x$ enters $H$
only if $x \in RCN(B,F)$, the second implies that if $x \in RCN(B,F)$ then $x$
eventually enters $H$. Once containment $B \cup H \subseteq BCN(B,F)$ is
proved, equality becomes equivalent to containment in the other direction:
$BCN(B,F) \subseteq B \cup H$.

The first part of the claim is that $B \cup H \subseteq BCN(B,F)$ holds during
the entire run of the algorithm. It is proved by induction on the number of
iterations of the loop. It holds at the beginning because $B \subseteq
BCN(B,F)$ holds and $H$ is initially empty. While $B$ does not change, $H$ is
updated by the instruction $H = H \cup \{x\}$ when a clause $B' \rightarrow x
\in F $ satisfies $B' \subseteq B \cup H$. The inductive assumption $B \cup H
\subseteq BCN(B,F)$ and transitivity of containment imply $B' \subseteq
BCN(B,F)$. This is defined as $F \models B \rightarrow b$ for every $b \in B'$,
or $F \models B \rightarrow B'$. Together with $B' \rightarrow x \in F$, it
implies $F \models B \rightarrow x$, which defines $x \in BCN(B,F)$. Therefore,
$B \cup H \subseteq BCN(B,F)$ still holds after the addition of $x$ to $H$.

The second part of the claim is that $B \cup H$ eventually contains all of
$BCN(B,F)$.

A part of the proof is to show that $B \cup H$ is larger when running the
algorithm on a larger formula. Formally, if $x$ is in $B \cup H$ at some point
when running the algorithm on $B$ and $F'$ then it is also in $B \cup H$ when
running it on $B$ and $F$ if $F' \subseteq F$,

This is proved by induction on the number of iterations of the loop. Initially
this number is zero, providing the base case of induction. In both runs $H$ is
empty. Since $B$ is the same in both runs, also $B \cup H$ is the same. This
proves the base case of induction. The inductive step assumes that $B \cup H$
on $F'$ is contained in $B \cup H$ on $F$ at the beginning of an iteration, and
requires proving the same at its end. If $x$ is added to $H$ when running on
$F'$, then $B' \rightarrow x \in F'$ holds for some $B' \subseteq B \cup H$;
the same conditions are also true when running on $F$: the first $B'
\rightarrow x \in F$ because $F' \subseteq F$, the second $B' \subseteq B \cup
H$ because of the inductive assumption. Therefore, $x$ is also added when
running on $F$.

This proves that $B \cup H$ monotonically increases with $F$.

The claim that $BCN(B,F) \subseteq B \cup H$ holds at some point of the
algorithm is proved by induction on the size of $F$. When $F$ is empty
$BCN(B,F) = B$. The claim $BCN(B,F) \subseteq B \cup H$ follows.

In the inductive case, the claim is that $BCN(B,F) \subseteq B \cup H$ holds at
some point when running on $B$ and $F$; the assumption is that the same holds
for every formula smaller than $F$.

The claim is the same as $x \in BCN(B,F)$ implying $x \in B \cup H$.
Lemma~\ref{set-implies-set} tells that $x \in BCN(B,F)$ implies either $x \in
B$ or $F^x \models B \rightarrow B'$ for some clause $B' \rightarrow x \in F$.
In the first case, $x \in B$ implies $x \in B \cup H$ and the claim is proved.
In the second case, $F^x \models B \rightarrow B'$ is the same as $b \in
BCN(F^x,B)$ for every $b \in B'$. The inductive assumption tells that $b \in
BCN(F^x,B)$ implies $b \in B \cup H$ at some point of running the algorithm on
$B$ and $F^x$. It has been proved that $B \cup H$ is larger when running on a
larger formula. Therefore, $b \in B \cup H$ also holds when running the
algorithm on $B$ and $F$. This holds for every $b \in B'$, implying $B'
\subseteq B \cup H$. Since $B' \rightarrow x \in F$, the algorithms adds $x$ to
$H$, making the inductive claim $x \in B \cup H$ true.~\qed

The algorithm collects all clauses it uses in its second return value. They
could be found by a separate algorithm, but producing them while calculating
$RCN(B,F)$ only requires the extra time of adding them to a set.

They are the clauses that are relevant to proving that something is a
consequence of $F \cup B$. All others are irrelevant. Formally, this set
contains a clause if and only if its precondition is entailed by $F \cup B$.

\begin{lemma}
\label{rcn-second}

The second return value of Algorithm~\ref{algorithm-rcnucl} is
{} $\{B' \rightarrow x \in F \mid B' \subseteq BCN(B,F)\}$.

\end{lemma}

\proof By Lemma~\ref{b-rcn}, $BCN(B,F) = B \cup RCN(B,F)$. The claim is
therefore equivalent to
{} $\{B' \rightarrow x \in F \mid B' \subseteq BCN(B,F)\}$
being the set of clauses $B' \rightarrow x \in F$ such that $B' \subseteq B
\cup RCN(B,F)$.

A clause $B' \rightarrow x$ enters the set $U$ if and only if $B' \subseteq B
\cup H$. The set $H$ is never removed elements, and its final value is
$RCN(B,F)$ by Lemma~\ref{rcn-first}. Therefore, $H \subseteq RCN(B,F)$ holds
during the entire run of the algorithm. This proves that all clauses in $U$
satisfy $B' \subseteq B \cup RCN(B,F)$.

In the other direction, let $B' \rightarrow x \in F$ be a clause satisfying $B'
\subseteq B \cup RCN(B,F)$. In the final iteration, $H = RCN(B,F)$ by
Lemma~\ref{rcn-first}. As a result, $B' \subseteq B \cup H$. The clause is
added to $U$ if not already.~\qed

\subsection{Finding the real consequences, quickly}

The {\tt rcnucl()} implementation in the {\tt singlehead.py} program uses the
standard Python functions for sets: a formula is a set of clauses, each
clause is a set of literals; the main loop is over the clauses of the formula,
each iteration checks whether the negative variables of a clause are contained
in $B \cup H$.

From the point of view of code simplicity, such an implementation is
unbeatable: apart from the initializations and the return statement, all is
done in six lines of code.

Yet, its efficiency is not optimal.

Since it implements unit propagation, it benefits from its optimizations:
clause indexing and unassigned variables counts~\cite{craw-auto-93}.

Scanning the formula each time in search for a clause containing some variables
is inefficient since a variable may be contained in just few clauses. A reverse
index avoids such a wasteful scan. Also, a clause may itself be large to scan
every time.

Not all clauses are actually needed, and not their entire content. Only the
clauses that contain variables in $B$ or in $H$ in their body matter, and what
matters of them is only whether their body only comprises such variables. The
first point is achieved by pointers from variables to clauses; the second by
storing the number of variables that are in the body of each clause but not in
$B \cup H$. Such an index comprises:

\begin{itemize}

\item for each clause, the index holds a record comprising an integer and a
variable; the integer is initialized with the size of its body, the variable is
its head;

\item for each variable, the index contains a list of pointers; each points to
the record of a clause containing the variable in its body.

\end{itemize}

The variables in $B$ are added to a queue, which is processed a variable at
time. For each variable, its list of pointers is scanned; for each pointer, the
integer in the pointed record is decreased; if this number reaches zero, the
variable in the record is added to $H$ and to the queue if not already in $B
\cup H$; the latter check requires constant time thanks to a vector
representing $B \cup H$.

The pointers avoid the loop over all clauses by pointing directly to the
clauses containing each variable in their body. The integer avoids the scan of
the clause. The variable is already guaranteed to be in the body thanks to the
pointer. Of the other variables of the body, what matters is only whether they
are all in $B \cup H$. This is the same as the number of variables in the body
but not in $B \cup H$ reaching zero.

\subsection{The algorithm, improved}

Algorithm~\ref{shmin-algorithm} makes a clause of $SHMIN(F)$ from each clause
$A \rightarrow x$ of $F$ by iteratively replacing $A$ with another set $B$ such
that $B <_F A$ and $F \models B \rightarrow x$; when such a set $B$ no longer
exists, it continues with sets $B \subset A$ such that $F \models B \rightarrow
x$. Each phase allows for some improvements.

\begin{itemize}

\item In the first phase, only the sets $B = BCN(A,F) \backslash \{a,x\}$ for
every $a \in A$ are checked. For such sets,
Lemma~\ref{less-minus-one-sufficient} proves that
{} $F \models B \rightarrow x$ and
{} $F \not\models B \rightarrow a$
imply $B <_F A$, which makes $B$ a valid replacement for $A$.

The first condition $F \models B \rightarrow x$ is the same as $x \in
BCN(B,F)$, or $x \in B \cup RCN(B,F)$. Since $B$ is $BCN(A,F) \backslash
\{a,x\}$, it does not contain $x$. Checking $x \in RCN(B,F)$ is enough.

If the first condition $F \models B \rightarrow x$ is met, the second $F
\not\models B \rightarrow a$ is the same as $a \in A \backslash RCN(A,F)$ by
Lemma~\ref{third-rcn}. This check is also expressed in terms of $RCN()$.

If they both hold, $B$ replaces $A$; the next step employs $RCN(A,F)$, but this
set needs not to be calculated again since it the same as $RCN(B,F)$ for the
set $B$ that replaced $A$.

\item In the second phase, only the sets $B = A \backslash \{a\}$ for every $a
\in A$ are checked. This is not a restriction since $B \subset A$ and $F
\models B \rightarrow x$ imply the same for every $B' = A \backslash \{a\}$
with $a \in A \backslash B$.

The check $F \models B \rightarrow x$ is the same as $x \in BCN(B,F)$, or $x
\in B \cup RCN(B,F)$. This is the same as $x \in RCN(B,F)$ since $x \not\in B$.
Indeed, $B$ is a subset of $A$, either the result of the first phase or one of
its subsets. In turn, this is either the original clause or a set $BCN(A,F)
\backslash \{a,x\}$; the second does not contain $x$ by construction, the first
because no clause of $F$ is tautologic by assumption.

The second phase begins only when the first cannot continue. No $B$ satisfies
both $F \models B \rightarrow x$ and $B <_F A$ if $F$ is inequivalent;
otherwise, completeness is not guaranteed anyway. The second phase searches for
sets $B$ such that $F \models B \rightarrow x$ and $B \subset A$. Since the
first phase is over, $F \models B \rightarrow x$ implies that $B <_F A$ is not
possible; $B \subset A$ implies $F \models A \rightarrow B$, which defines $B
\leq_F A$. A consequence of $B \leq_F A$ and the impossibility of $B <_F A$ is
$A \equiv_F B$.

Equivalence allows for a little improvement in the second phase. The base is
Lemma~\ref{rcn-less}: if $a \in A \backslash RCN(A,F)$ then $A \backslash \{a\}
<_F A$. Since $B <_F A$ does not hold during the second phase, the sets $A
\backslash \{a\}$ with $a \in A \backslash RCN(A,F)$ do not need to be checked,
only the ones with $a \in A \cap RCN(A,F)$ do.

If $F$ is not inequivalent, $B$ may not be equivalent to $A$. Correctness is
still guaranteed because of the final check $F \equiv SHMIN(F)$, while
completeness is not anyway.

\end{itemize}

The complete algorithm with all these efficiency improvements in place follows.

\

\begin{verbatim}
def shmin(f):
    s = set()
    d = set()

    for c in f:
        h = head(c)
        b = body(c)

        # only one clause for each head

        if h in d:
            continue
        d |= {h}

        # minimize according to <F

        a = None
        r,u = rcnucl(b, f)
        while a != b and b - r:
            a = b
            for e in b - r:
                nb = (b | r) - {e,h}
                nr,nu = rcnucl(nb, u)
                if h in nr:
                    b = nb
                    r = nr
		    u = nu
                    break

        # minimize according to set containment

        a = None
        while a != b and b & r:
            a = b
            for e in b & r:
                nb = b - {e}
                nr,nu = rcnucl(nb, u)
                if h in nr:
                    b = nb
                    r = nr
                    u = nu
                    break

        s |= {frozenset([h]) | frozenset(['-' + l for l in b])}
    return s
\end{verbatim}

\

The aim of $SHMIN(F)$ is to accelerate forgetting: if it is equivalent to $F$
forgetting can be done on it in place of $F$.

Any other single-head formula could be used in the same way, but $SHMIN(F)$ is
a good choice because it is guaranteed to be equivalent to $F$ in at least one
case: when $F$ is inequivalent and single-head equivalent. It is not in
general, as shown by the formula
{} $F =
{}   \{a \rightarrow b, b \rightarrow a, b \rightarrow c, c \rightarrow b\}$
in the {\tt twoequiv.py} test file of the {\tt singlehead.py} program.
A single-head formula equivalent to $F$ is for example
{} $\{a \rightarrow b, b \rightarrow c, c \rightarrow a\}$.
In order to produce it, $SHMIN(F)$ would have to replace $b \rightarrow a$ with
$c \rightarrow a$, which it does not because $\{b\} \equiv_F \{c\}$. Replacing
a set with an equivalent one would deprive the algorithm of its termination
guarantee since equivalent sets form loops. Checking for sets already analyzed
is unfeasible because they may be exponentially many.

When $F$ is single-head equivalent but not inequivalent, {\tt shmin(f)} may not
find the single-head version of $F$. Yet, it may. Formula~\ref{minf-f} is an
example:
{} $F =
{}    \{a \rightarrow b, b \rightarrow a, b \rightarrow c, c \rightarrow a\}$;
this formula is in the {\tt incomplete.py} test file of the {\tt
singlehead.py} program.
The {\tt shmin(f)} function sometimes generates
{} $\{a \rightarrow b, b \rightarrow c, c \rightarrow a\}$,
which is equivalent to $F$, and sometimes
{} $\{a \rightarrow b, b \rightarrow a, b \rightarrow c\}$,
which is not. It depends on the order of the clauses in the main loop. All
clauses of $F$ have a minimal body since $F$ makes every literal equivalent to
each other. As a result, {\tt shmin(f)} returns $a \rightarrow b$, $b
\rightarrow c$ and whichever between $b \rightarrow a$ and $c \rightarrow a$
comes first in the main loop. The latter makes the output equivalent to $F$,
the former does not. While depending on the order of analysis of the clauses is
an undesirable algorithm behavior, this example also shows a positive feature
of {\tt shmin(f)}: it may find a single-head equivalent formula even if
$MIN(F)$ is not single-head.

When $SHMIN(F)$ is not equivalent to $F$, is its calculation wasted time? Maybe
not. Not completely, at least. Since $SHMIN(F)$ outputs a clause $B \rightarrow
x$ only if $F \models B \rightarrow x$, these clauses are all entailed by $F$.
Globally, $F \models SHMIN(F)$. Non-equivalence may only be due to $SHMIN(F)
\not\models A \rightarrow x$ for some $A \rightarrow x \in F$. Such clauses can
be added to $SHMIN(F)$, or the algorithm be run again to determine a
minimal-body clause for each. The resulting formula is still better than the
original because it turns a single-head equivalent subformula into a
single-head subformula. For example, if $SHMIN(F)$ implies all clauses of $F$
but $a \rightarrow b$, then $F \backslash \{a \rightarrow b\}$ is single-head
equivalent and $SHMIN(F)$ is a single-head formula equivalent to it. The
replacing algorithm for forgetting performs well on formulae like $SHMIN(F)
\cup \{a \rightarrow b\}$ that contain only two same heads. The addition of $a
\rightarrow b$ only doubles the recursive calls at most. Running time is still
polynomial.

\subsection{Disproving single-head equivalence}

The algorithm for $SHMIN(F)$ tries to produce a single-head formula that is
semantically close to $F$. When computing forgetting, all of $SHMIN(F)$ is
required, possibly with the addition of other clauses if it is not equivalent
to $F$. If the aim is instead just to check whether a single-head equivalent
formula exists, generating all of $SHMIN(F)$ is wasteful. For example, if $F$
contains $a \rightarrow x$ and $b \rightarrow x$ but implies neither $a
\rightarrow b$ nor $b \rightarrow a$, it is not single-head equivalent.
Computing all of $SHMIN(F)$ is unnecessary.

Unfortunately, such a property only concerns an individual head, and looking at
one head at time may not be sufficient. A counterexample is
{} $\{a \rightarrow b, b \rightarrow c, c \rightarrow b\}$,
in the testing file {\tt local.py} of the {\tt singlehead.py} program.
It is single-head but for $b$, and is equivalent to
{} $\{a \rightarrow c, b \rightarrow c, c \rightarrow b\}$,
which is single-head but for $c$. For each of its variables, an equivalent
formula that is single-head on that variable exists. Yet, no equivalent formula
is single-head on all variables. Every property concerning the heads separately
fails at recognizing that it is not single-head equivalent.

Nonetheless, a sufficient condition may be useful anyway, one that allows to
sometimes cut $SHMIN(F)$ short because no single-head formula equivalent to $F$
exists.

A minimal modification of the algorithm is to compare all sets $BCN(A,F)
\backslash \{a,x\}$ that meet the conditions for replacing $A$. If two of them
are incomparable, they may lead to different minimal bodies. Unfortunately,
this is not always the case. A counterexample is
{} $F = \{abd \rightarrow x, ab \rightarrow d, d \rightarrow x\}$.
Removing either $a$ or $b$ from $A = \{a,b,d\}$ produces a set that can replace
it: $\{a,d\}$ entails $x$ but not $b$; $\{b,d\}$ entails $x$ but not $a$. These
sets are incomparable, yet the formula is equivalent to the single-head formula
$\{ab \rightarrow d, d \rightarrow x\}$,
as shown by the {\tt alternatives.py} test file of the {\tt singlehead.py}
program.

Even sets of variables Algorithm~\ref{shmin-algorithm} no longer replaces may
be incomparable even if the formula is single-head.

The counterexample is based on the formula in the proof of
Lemma~\ref{less-minus-one-not-necessary}. There, the formula $F$ was
architected to make a set $A$ minimal by having each of its elements entailed
by the others and the consequences of $A$. Here, two copies of the same
structure make two minimal sets incomparable:

\[
F = \{
ab \rightarrow d,
ad \rightarrow b,
bd \rightarrow a,
a'b' \rightarrow d',
a'd' \rightarrow b',
b'd' \rightarrow a',
dd' \rightarrow x
\}
\]

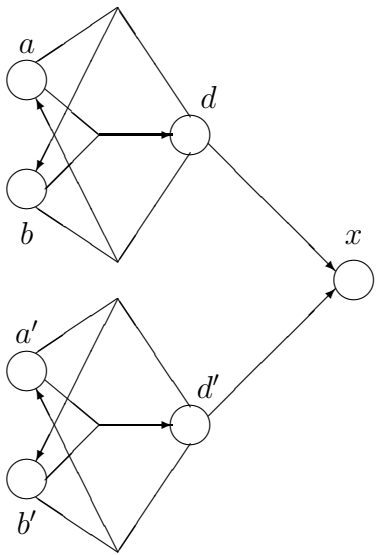
\setlength{\unitlength}{5000sp}%
\begingroup\makeatletter\ifx\SetFigFont\undefined%
\gdef\SetFigFont#1#2#3#4#5{%
  \reset@font\fontsize{#1}{#2pt}%
  \fontfamily{#3}\fontseries{#4}\fontshape{#5}%
  \selectfont}%
\fi\endgroup%
\begin{picture}(1838,2724)(5562,-5923)
{\color[rgb]{0,0,0}\thinlines
\put(5671,-3571){\circle{202}}
}%
{\color[rgb]{0,0,0}\put(5671,-4111){\circle{202}}
}%
{\color[rgb]{0,0,0}\put(6481,-3841){\circle{202}}
}%
{\color[rgb]{0,0,0}\put(5671,-5011){\circle{202}}
}%
{\color[rgb]{0,0,0}\put(5671,-5551){\circle{202}}
}%
{\color[rgb]{0,0,0}\put(6481,-5281){\circle{202}}
}%
{\color[rgb]{0,0,0}\put(7291,-4561){\circle{202}}
}%
{\color[rgb]{0,0,0}\put(5761,-3616){\line( 6,-5){270}}
\put(6031,-3841){\line(-1,-1){270}}
}%
{\color[rgb]{0,0,0}\put(6031,-3841){\vector( 1, 0){360}}
}%
{\color[rgb]{0,0,0}\put(6481,-3751){\line(-2, 3){360}}
\put(6121,-3211){\line(-3,-2){405}}
}%
{\color[rgb]{0,0,0}\put(6481,-3931){\line(-2,-3){360}}
\put(6121,-4471){\line(-3, 2){405}}
}%
{\color[rgb]{0,0,0}\put(6121,-4471){\vector(-1, 2){405}}
}%
{\color[rgb]{0,0,0}\put(6121,-3211){\vector(-1,-2){405}}
}%
{\color[rgb]{0,0,0}\put(5761,-5056){\line( 6,-5){270}}
\put(6031,-5281){\line(-1,-1){270}}
}%
{\color[rgb]{0,0,0}\put(6031,-5281){\vector( 1, 0){360}}
}%
{\color[rgb]{0,0,0}\put(6481,-5191){\line(-2, 3){360}}
\put(6121,-4651){\line(-3,-2){405}}
}%
{\color[rgb]{0,0,0}\put(6481,-5371){\line(-2,-3){360}}
\put(6121,-5911){\line(-3, 2){405}}
}%
{\color[rgb]{0,0,0}\put(6121,-5911){\vector(-1, 2){405}}
}%
{\color[rgb]{0,0,0}\put(6121,-4651){\vector(-1,-2){405}}
}%
{\color[rgb]{0,0,0}\put(6571,-3886){\vector( 1,-1){630}}
}%
{\color[rgb]{0,0,0}\put(6571,-5236){\vector( 1, 1){630}}
}%
\put(5671,-3436){\makebox(0,0)[b]{\smash{{\SetFigFont{12}{24.0}
{\rmdefault}{\mddefault}{\updefault}{\color[rgb]{0,0,0}$a$}%
}}}}
\put(6571,-3706){\makebox(0,0)[b]{\smash{{\SetFigFont{12}{24.0}
{\rmdefault}{\mddefault}{\updefault}{\color[rgb]{0,0,0}$d$}%
}}}}
\put(5671,-4381){\makebox(0,0)[b]{\smash{{\SetFigFont{12}{24.0}
{\rmdefault}{\mddefault}{\updefault}{\color[rgb]{0,0,0}$b$}%
}}}}
\put(7291,-4381){\makebox(0,0)[b]{\smash{{\SetFigFont{12}{24.0}
{\rmdefault}{\mddefault}{\updefault}{\color[rgb]{0,0,0}$x$}%
}}}}
\put(5671,-4876){\makebox(0,0)[b]{\smash{{\SetFigFont{12}{24.0}
{\rmdefault}{\mddefault}{\updefault}{\color[rgb]{0,0,0}$a'$}%
}}}}
\put(6571,-5146){\makebox(0,0)[b]{\smash{{\SetFigFont{12}{24.0}
{\rmdefault}{\mddefault}{\updefault}{\color[rgb]{0,0,0}$d'$}%
}}}}
\put(5671,-5821){\makebox(0,0)[b]{\smash{{\SetFigFont{12}{24.0}
{\rmdefault}{\mddefault}{\updefault}{\color[rgb]{0,0,0}$b'$}%
}}}}
\end{picture}%
\nop{
 +-----+---------+
 |     |         |
 |     V         |
 |  +- a ---+    |
 |  |       +--> d -------+
 +--|- b ---+    |        |
    |  ^         |        |
    |  |         |        |
    +--+---------+        |
                          +------> x
 +-----+---------+        |
 |     |         |        |
 |     V         |        |
 |  +- a' --+    |        |
 |  |       +--> d' ------+
 +--|- b' --+    |
    |  ^         |
    |  |         |
    +--+---------+
}

This formula is single-head.
It is used as an example in the {\tt incomparable.py} test file of the {\tt
singlehead.py} program.
Since it implies
{} $abd' \rightarrow x$
and
{} $a'b'd \rightarrow x$,
adding these two clauses preserves equivalence. Therefore, the resulting
formula is single-head equivalent.

The same line of proof of Lemma~\ref{less-minus-one-not-necessary} shows that
$A = \{a,b,d'\}$ is minimal when restricting to single-variable removal. The
starting point is $BCN(A,F) = \{a,b,d,d',x\}$. It makes $BCN(A,F) \backslash
\{a,x\}$ equal to $\{b,d,d'\}$, which entails $a$; the same holds for $b$ by
symmetry; for $d'$, it holds $BCN(A,F) \backslash \{d',x\} = \{a,b,d\}$, which
does not entail $x$. Either way, removing a single variable from $BCN(A,F)
\backslash \{x\}$ does not produce a set that replaces $A$.

The same applies to $B = \{d,a',b'\}$ by symmetry.

These two sets $A$ and $B$ are incomparable: $A$ does not imply $a' \in B$ and
$B$ does not imply $a \in A$.

Conclusion: the single-head equivalent formula
{} $F \cup \{abd' \rightarrow x, a'b'd \rightarrow x\}$
makes two sets $A$ and $B$ minimal but incomparable when removing only a single
variable. Yet, the formula is single-head equivalent. Minimal-yet-incomparable
sets do not disprove single-head equivalence.

Still better, two minimal-yet-incomparable sets do not disprove single-head
equivalence. A similar example with three, four and more sets is easy to make
by replicating the core of the counterexample, the clauses around $a$, $b$ and
$d$. Arbitrarily many sets may be minimal but incomparable. Still, many is not
all. Lemma~\ref{equivalent-preconditions} proves that not all sets implying the
same variable can be incomparable in a single-head formula.

This is not obvious, since the lemma proves something seemingly unrelated: if a
formula $F$ is equivalent to the single-head formula $F'$ that contains $A
\rightarrow x$, then $F$ contains $B \rightarrow x$ with $A \equiv_F B$. What
does it tell about comparability? Every non-tautologic clause $C \rightarrow x$
entailed by $F$ is also entailed by $F'$, and $F' \models C \rightarrow x$
implies $F' \models C \rightarrow A$ by Lemma~\ref{set-implies-set} since $A
\rightarrow x$ is the only clause of $F'$ with head $x$. Equivalence implies $F
\models C \rightarrow A$, which defines $A \leq_F C$, which implies $B \leq_F
C$ because of $A \equiv_F B$.

The body of a clause of $F$ is less than or equal to all other bodies of
non-tautologic clauses entailed by $F$. The algorithm only creates bodies of
clauses entailed by $F$. The sufficient condition that disproves single-head
equivalence is: if none of the bodies of $F$ is less than or equal to all
bodies created by the algorithm, then $F$ is not single-head equivalent. This
is the case even for bodies that are not minimal. What matters is only that the
clause with that body is entailed by $F$, and all bodies generated by the
algorithm are.

\begin{corollary}

If $F$ entails $C \rightarrow x$ but contains no clause $B \rightarrow x$
such that $B \leq_F C$, then it is not single-head equivalent.

\end{corollary}

This condition is only sufficient to disprove single-head equivalence. In the
other way around, it is yet another necessary condition. Its sufficiency is
disproved by a previous example:
{} $F = \{a \rightarrow b, b \rightarrow c, c \rightarrow b\}$,
in the {\tt inloop.py} test file of the {\tt singlehead.py} program.
It is proved not single-head equivalent in Section~\ref{insidious}. Yet, it
satisfies the condition: the entailed clauses are $a \rightarrow b$, $a
\rightarrow c$, $b \rightarrow c$ and $c \rightarrow b$ and their superclauses.
The first, third and fourth are in $F$. The second $a \rightarrow c$ satisfies
the condition because of $\{b\} \leq_F \{a\}$ and $b \rightarrow c \in F$. The
superclauses satisfy the condition because their bodies are greater than the
body of the subclause.

\subsection{Python implementation}
\label{python}

The {\tt singlehead.py} program available from
{} {\tt https://github.com/paololiberatore/singlehead}
implements the {\tt shmin()} function. It can be called directly on a formula
and tells whether it is single-head equivalent according to {\tt shmin()}.

\begin{verbatim}
singlehead.py -f 'a->b' 'abd->c' 'b=d' 'b->c'
\end{verbatim}

The clauses are passed each as a commandline option. Formulae like {\tt ab->cd}
or {\tt ab=cd} are accepted in place of clauses. Variables are single
characters, which bounds them to the ones accepted by the Python interpreter,
currently about a million.

The program first outputs the formula in definite Horn form, with subformulae
like {\tt ab->cd} or {\tt ab=cd} turned into clauses. For each clause of the
formula, it prints the bodies that replace its because of $\leq_F$ and then
because of $\subset$, followed by the final clause of {\tt shmin()}.
Eventually, it prints the generated formula on a single line, and whether it is
equivalent to the input formula.

\begin{verbatim}
## cmdline formula ##
formula: a->b abd->c b=d b->c
b->d | | | b->d
a->b | dc | d | d->b
dba->c | db | b | b->c
d->b | [head already in shmin]
b->c | [head already in shmin]
shmin: b->d d->b b->c
shmin equivalent: False
expected result: None
\end{verbatim}

As an example, the fourth line {\tt a->b | dc | d | d->b} is the result of
processing the clause {\tt a->b}. Its body {\tt a} is first replaced by {\tt
dc} because of $dc <_F a$. No further replacement is possible according to the
order, this is way the separator {\tt |} follows. The body {\tt bc} contains
{\tt d}, which implies the head {\tt b}. Therefore, {\tt bc} is replaced by
{\tt d}. No subset of it implies {\tt b}. The final clause is {\tt d->b}, the
last part of the line.

A line like this is printed for each input clause: the input clause, the bodies
that replace its according to $<_F$ and then to $\subset$, and the final
clause. Three pipe characters {\tt |} separate the two minimization phases from
each other and from the input and output clauses.

An exception is {\tt d->b | [head already in shmin]}, meaning that the input
clause {\tt d->b} is not processed at all because a clause of the same head is
already generated.

The line {\tt shmin: b->d d->b b->c} shows the output formula. The following
line tells whether it is equivalent to the input formula. A positive answer is
certain: the output formula is single-head and equivalent to the given one. A
negative answer is inconclusive, as the input formula may still be single-head
equivalent.

The last line of the output is redundant for formulae given on the command
line, but essential to automated testing. If the formula is given in a file
rather than the commandline arguments, an expected result can be provided. The
name of the file is provided as an argument, possibly preceded by the {\tt -t}
option:

\begin{verbatim}
singlehead.py -t tests/conditiontwo.py
\end{verbatim}

As an example, the formula disproving the sufficiency of
Condition~\ref{second-condition} to single-head equivalence is in the file {\tt
conditiontwo.py}. Such a file contains one or more calls to the {\tt analyze()}
function. Its first argument is a string describing the formula, the second is
the expected result of the test, the following are the clauses of the formula.

\begin{verbatim}
analyze(
    'second condition is insufficient',
    False,
    'ab->x', 'bx->c', 'ac->d', 'd->x')
\end{verbatim}

Since the program fails to produce a single-head equivalent formula, the test
is deemed passed because of the second argument {\tt False}. The string {\tt
TEST PASSED} ends the program output. Other possible outcomes are {\tt TEST
FAILED} and {\tt TEST INCONCLUSIVE}. The latter is the mark of incompleteness
of the program: the formula is declared single-head equivalent, but the program
fails to prove it.

\section{Conclusions}
\label{conclusions}

If a definite Horn formula does not have two clauses with the same head,
logical forgetting~\cite{delg-17} is easy to compute and produces a
polynomially sized output~\cite{libe-20-a}. The single-head restriction is easy
to tell: no two clauses have the same head. A formula that is not single-head
may still be equivalent to one that is. Forgetting can be performed on that,
since it is the same on equivalent formulae. The single-head equivalence
concept is where the troubles begin.

Checking the clauses of a formula for duplicated heads is easy as it only
requires a simple scan of the formula. Checking for an equivalent formula with
this property is difficult because of the many equivalent formulae. A parallel
with a classical problem is in order: checking the size of a formula is
straightforward; checking whether a formula is equivalent to one of a given size
is not~\cite{coud-94,coud-sasa-02,uman-etal-06}. It took twenty years just for
being precisely classified in the polynomial hierarchy~\cite{stoc-76,uman-01}.

The first direction of attack is to turn the definition based on the many
equivalent formulae into a condition based on the semantics of the formula
alone. Equivalent formulae are either all single-head equivalent or they are
all not. Given a set of models, the theory predicts that all formulae satisfied
precisely by them are the same on single-head equivalence. Another parallel
explains the aim: a set of propositional interpretations are the models of some
Horn formula if and only if the intersection of every two of them is also one
of them~\cite{mcki-43}; the intersection is defined as the model setting a
variable to true only if and only if both intersecand models do. This condition
is easy to express, at least in theory: every two models, their intersection.
It tells whether a set of models is Horn-expressible or not without checking
the many equivalent formulae that express the set of models. Similarly, a set
of models is either single-head expressible or not. The definition of
single-head expressible is that some single-head formula has precisely this set
of models. What would be useful is a necessary and sufficient condition to
single-head expressible, simple like ``two models, their intersection''.

Such a condition seems not too hard to find. If a formula is single-head
equivalent, it is equivalent to a single-head formula. That formula contains at
most a clause with each variable in the head. The problem is to tell its body.
Still better, the problem is to tell whether a single body suffices, and all
others reduce to them. Such reducibility proved slippery ground: sometimes, a
clause seems redundant because it reduces to another, but the reduction
requires the clause itself. An entire section of this article shows necessary
conditions to single-head equivalence that all fell on sufficiency. A series of
examples and counterexamples show that single-head equivalence is not obvious
to express when looking at a set of models. A particularly significant example
is Formula~\ref{cyclic-with-ab}:
{} $F = \{
{} 	ab \rightarrow x,
{} 	bx \rightarrow c,
{} 	ac \rightarrow d,
{} 	d \rightarrow x
{} $\}.
In spite of being just a mere, single Boolean formula, it is one of the main
outcomes of this article. It exemplifies the complication of clauses reducing
to others: $ab \rightarrow x$ appears to be reducible to $d \rightarrow x$
thanks to $F \models ab \rightarrow d$, but this entailment requires $ab
\rightarrow x$ itself. It also shows useful elsewhere in the article, but its
main feature is to negate the semantical simplicity of single-head equivalence.
Every necessary and sufficient condition to single-head equivalence and every
algorithm for single-head equivalence must be tested on this formula. Doing so
tells whether the main pitfall of single-head equivalence is ducked or not.

The two necessary conditions in Section~\ref{insidious} fell on it. They tell
the formula single-head equivalent. It is not. They may be necessary, but they
are not sufficient.

This Section~\ref{insidious} looks like a string of failures: a first condition
is found that looks obviously the same as single-head equivalence, but is not;
a three-clauses counterexample fails it; another condition patches it, but its
being syntactic stops it; equivalence between sets of variables shows to be the
problem, but does not forbid single-head equivalence; a second condition seems
to account for equivalence between sets of variables;
Formula~\ref{cyclic-with-ab} crushes this delusion. All of this may look like
someone banging their head on the wall over and over again, blind to the large
door nearby. But maybe the door is not so close, and maybe not that large. The
sequence of failed conditions, examples and counterexamples make it look not.

The counterexamples all share a feature: their clauses form cycles. Not by
chance: forbidding cycles makes all problems disappear. Necessary and
sufficient conditions become easy, an efficient algorithm for checking
single-head equivalence becomes possible.

Cyclicity in logics is a well-established
concept~\cite{angl-87,angl-87,gold-sloa-95,lee-lifs-03}. Not surprisingly, the
related problem of formula minimization is easy on acyclic propositional
formulae~\cite{hamm-koga-95}. Single-head equivalence is a form of formula
minimization where formulae are measured not by the total number of clauses or
literal occurrences but by the maximal number of same heads, and the question
is whether a formula can be minimized to measure one. Like formula
minimization, checking single-head equivalence is easy on acyclic formulae.

Still better, it is easy on semantically acyclic formulae. Because a formula
may be syntactically cyclic and may be semantically cyclic. The distinction is
the same as that of cyclic functions and cyclic formulae by Hammer and
Kogan~\cite{hamm-koga-95}. Semantical matches semantical, and single-head
equivalence is a semantical concept. It is the semantical version of being
single-head: it depends on the models of the formula rather than its
clauses---its syntax. A formula is single-head equivalent if it is equivalent
to a formula that is single-head; a formula is semantically acyclic if it is
equivalent to a formula whose clauses do not form cycles. The parallel is
evident.

Acyclic formulae do not suffer from the subtleties of single-head equivalence.
The very first sufficient condition that failed to be necessary in general
succeeds on semantically acyclic formulae.

Syntactic acyclicity has its part too. It implies a property that does not hold
in general, not even on all semantically acyclic formulae: if a formula is
single-head equivalent, it is equivalent to a single-head subset of its.
Irredundancy highlights the value of this result: a syntactically acyclic
irredundant formula is single-head equivalent if and only if it is single-head.
Single-head equivalence can be checked by making the formula irredundant, for
example removing one by one its redundant clauses, and checking whether the
result is single-head or not. The same property does not hold on semantically
acyclic formulae. Not a surprise: subsets are syntactic, as well as
irredundancy. The semantic cognate of irredundancy is minimization:
semantically acyclic formulae can be checked for single-head equivalence by
minimizing them.

Both syntactically and semantically acyclic formulae have their algorithm for
checking single-head equivalence, and for finding a single-head equivalent
formula if any: make the formula irredundant or minimal, and check it for
duplicated heads. Nothing similar works in the general definite Horn case: an
irredundant or minimal formula may have duplicate heads and still being
single-head equivalent.

What helps is ordering the bodies of the clauses according to entailment. A
body is greater than another if implies it. The bodies of a single-head formula
are minimal. Since the order is defined semantically, they are also minimal for
all equivalent formulae. Seen from the other direction, if a formula is
equivalent to a single-head formula, the bodies of that are minimal according
to the order, which is the same for both formulae. The problem shifts from
finding a single-head formula that is equivalent to the given one to finding
the minimal bodies.

Minimizing is no easy task since sets of variables are exponentially many. The
implemented algorithm exploits testing the removal of a single variable at
time. When reaching a minimum according to the ordering, the set of literals is
further minimized according to set containment.

The algorithm proves complete on inequivalent formulae, an extension of
semantically acyclic formulae. It is therefore complete also in its subcases of
semantically and syntactically acyclic formulae. It is not complete in the
general definite Horn case. Yet, it is superior to the methods for acyclic
formulae based on making the formula irredundant or minimal. First, it works on
the larger set of inequivalent formulae. Second, while it does not work in
general like them, it is still useful: it produces a formula that can be
completed by the addition of other clauses. Even when the formula is not
single-head equivalent, it may reduce the duplicated heads, thereby speeding
the forgetting algorithm.

Inequivalence means that a formula makes two sets of variables equivalent only
if it makes them equivalent to their intersection. Every syntactically or
semantically acyclic formula is inequivalent. Not the other way around:
Formula~\ref{cyclic-with-ab} is inequivalent but cyclic. Inequivalent formulae
are easy to check for single-head equivalence using the algorithm based on the
order between sets of variables. Yet, inequivalence do not inherit the
necessary and sufficient conditions for single-head equivalence:
Formula~\ref{cyclic-with-ab} again meets them both but is not single-head
equivalent.

The algorithm based on the order of the bodies always runs in polynomial time
but is incomplete in general: it may fail to find a single-head equivalent
formula even if one exists. An alternative algorithm~\cite{libe-20-c} is
complete, but sometimes requires exponential time. Instead of searching a body
for each head, it searches for the appropriate head for each body.

An algorithm is polynomial but incomplete, another is complete but not
polynomial. The question is whether the problem itself can be solved in
polynomial time at all. The complexity of single-head equivalence is an open
question. It is in \np\  since it can be solved by guessing and checking: the
guessed formula has at most as many clauses as variables because of the
single-head restriction, and checking it for equivalence is polynomial-time
because of the Horn restriction. Whether the problem is also \np-hard is still
to be established.

Some tests of the algorithm suggest a probabilistic workaround. On some
formulae, it succeeds. On some others, it fails. But on yet some others, it
sometimes fails and sometimes succeeds. For example, the {\tt singlehead.py}
program run on the {\tt incomplete.py} test file may not find a single-head
equivalent formula only to find it the second time it is run on the same
formula.

The log of execution shows why: the program minimizes the first clause for each
head and discards the others. This makes it depend on the order of analysis of
the clauses in the formula. A formula is a set of clauses, and standard
implementations of sets do not guarantee the same order of visit of the
elements in a set. The main loop of the algorithm may sometimes follow an order
and sometimes another. If it fails, it may still succeed when run a second or
third time on the same formula.

Or not. The implementations of sets do not ensure the same order every time,
but do not guarantee they are different either, and especially they do not
guarantee that all orders are eventually followed. The easiest way to get close
to that is to randomly sort the clauses of the formula. With high probability,
all possible orders of clauses are tested when running the algorithm many
times.

\nojournal

Probabilities may also come to help within the algorithm itself. When it does
not find a minimal body is because it only tests bodies obtained by removing a
single variable. None of these may be less than the current one, while another
body with two or more elements removed is. A solution is to sometimes replace
the current body with one that is not less than it but just equivalent to it.
Doing it always impairs termination since equivalent bodies form loops. Doing
it randomly does not. This way, termination is maintained because sometimes the
current body is not replaced, but a minimal body is found with positive
probability.

\endnojournal

\bibliographystyle{alpha}
\newcommand{\etalchar}[1]{$^{#1}$}

\end{document}